**FERSIWN GYMRAEG ISOD**

Corpws Cenedlaethol Cymraeg Cyfoes
National Corpus of Contemporary Welsh

# The National Corpus of Contemporary Welsh

Project Report, October 2020

**Authors:** Dawn Knight[1], Steve Morris[2], Tess Fitzpatrick[2], Paul Rayson[3], Irena Spasić and Enlli Môn Thomas[4].

# 1. Introduction

## 1.1. Purpose of this report

This report provides an overview of the CorCenCC project and the online corpus resource that was developed as a result of work on the project. The report lays out the theoretical underpinnings of the research, demonstrating how the project has built on and extended this theory. We also raise and discuss some of the key operational questions that arose during the course of the project, outlining the ways in which they were answered, the impact of these decisions on the resource that has been produced and the longer-term contribution they will make to practices in corpus-building. Finally, we discuss some of the applications and the utility of the work, outlining the impact that CorCenCC is set to have on a range of different individuals and user groups.

## 1.2. Licence

The CorCenCC corpus and associated software tools are licensed under Creative Commons CC-BY-SA v4 and thus are freely available for use by professional communities and individuals with an interest in language. Bespoke applications and instructions are provided for each tool (for links to all tools, refer to section 10 of this report). When reporting information derived by using the CorCenCC corpus data and/or tools, CorCenCC should be appropriately acknowledged (see 1.3).

- To access the corpus visit: www.corcencc.org/explore
- To access the GitHub site: https://github.com/CorCenCC
    - GitHub is a cloud-based service that enables developers to store, share and manage their code and datasets.

## 1.3. Referencing CorCenCC

Appropriate credit needs to be given when using the CorCenCC corpus data and/or tools. To reference the CorCenCC corpus, please use the following:

[1] Cardiff University
[2] Swansea University
[3] Lancaster University
[4] Bangor University

- **CorCenCC corpus**: Knight, D., Morris, S., Fitzpatrick, T., Rayson, P., Spasić, I., Thomas, E-M., Lovell, A., Morris, J., Evas, J., Stonelake, M., Arman, L., Davies, J., Ezeani, I., Neale, S., Needs, J., Piao, S., Rees, M., Watkins, G., Williams, L., Muralidaran, V., Tovey-Walsh, B., Anthony, L., Cobb, T., Deuchar, M., Donnelly, K., McCarthy, M. and Scannell, K. (2020). *CorCenCC: Corpws Cenedlaethol Cymraeg Cyfoes - The National Corpus of Contemporary Welsh* [Digital Resource]. Available at: www.corcencc.org/explore

Other project publications can be found in section 10 of this report and on the 'Outputs' tab of the CorCenCC website: www.corcencc.org/outputs

### 1.4. Acknowledgements

The research on which this report, and the accompanying online corpus resource, are based was funded by the UK Economic and Social Research Council (ESRC) and Arts and Humanities Research Council (AHRC) as the *Corpws Cenedlaethol Cymraeg Cyfoes (The National Corpus of Contemporary Welsh): A community driven approach to linguistic corpus construction* project (Grant Number ES/M011348/1). Information about project team members can be found at www.corcencc.org/contacts. Without their input, expertise, enthusiasm and collegiality, the CorCenCC project would not have been possible.

We would also like to acknowledge Cardiff University and Swansea University for their contribution of PhD scholarships, enabling us to include postgraduate researchers in the project team. We particularly extend personal thanks to those colleagues at all our respective universities who have generously given their time and support to us at critical stages of the project.

The fulfilment of the CorCenCC project is also thanks to our project stakeholders (see 2.2), and especially those in the project advisory group: not only have they been generous in facilitating the collection of, or directly contributing, data, they have also been encouraging in their support for the aims of the project and their engagement with the planning process, as well as their commitment to the sustainability and continuation of CorCenCC.

## 2. Vision and objectives

### 2.1. Project overview

CorCenCC is an inter-disciplinary and multi-institutional project that has created a large-scale, open-source corpus of contemporary Welsh. A corpus, in this context, is a collection of examples of spoken, written and/or e-language examples from real life contexts, that allows users to identify and explore language as it is actually used, rather than relying on intuition or prescriptive accounts of how it 'should' be used. Corpora let us investigate how we use language across different genres and communicative mediums (i.e. spoken, written or digital), and how it varies according to the speaker/writer and the communicative purpose. This evidence-based approach is used by academic researchers, lexicographers, teachers, language learners, assessors, resource developers, policy makers, publishers, translators and others, and is essential to the development of technologies such as predictive text production, word processing tools, machine translation, speech recognition and web search tools.

Prior to the construction of CorCenCC, there were a number of Welsh language corpora in existence, including the 460k word spoken Siarad corpus (Deuchar et al., 2018), the 24m word e-language-based Crúbadán Welsh Corpus (Scannell, 2007) and the crowdsourced 40-hour Paldaruo Speech Corpus comprised of read-aloud texts (Cooper et al., 2019). Consideration was given as to whether these could be integrated or aligned with the objectives of CorCenCC. Deuchar and Scannell were engaged as consultants for the project, while Canolfan Bedwyr (Bangor) was represented on the project advisory group. However, given that the previous corpora were compiled to realise different, distinct and bespoke aims and visions, it was considered necessary to create a new, complete dataset.

CorCenCC is the first corpus of the Welsh language that covers all three aspects of contemporary Welsh: spoken, written and electronically mediated (e-language). It offers a snapshot of the Welsh language across a range of contexts of use, e.g. private conversations, group socialising, business and other work situations, in education, in the various published media, and in public spaces. It includes examples of news headlines, personal and professional emails and correspondence, academic writing, formal and informal speech, blog posts and text messaging (the specific composition of the corpus is discussed in section 3.3). Language data was sampled from a range of different speakers and users of Welsh, from all regions of Wales, of all ages and genders, with a wide range of occupations, and with a variety of linguistic backgrounds (e.g. how they came to speak Welsh), to reflect the diversity of text types and of Welsh speakers found in contemporary Wales. In this way, the CorCenCC corpus provides the means for empowering users of Welsh to better understand and observe the language across diverse settings, and creates a solid evidence base for the teaching of contemporary Welsh to those who aspire to use it. Over time, the corpus has the potential to make a significant contribution to the transformation of Welsh as the language of public, commercial, education and governmental discourse.

To that end, CorCenCC is designed to enable, for example, community users to investigate dialect variation or idiosyncrasies of their own language use; professional users to profile texts for readability or develop digital language tools; Welsh language learners to draw on real life models of Welsh; and researchers to investigate patterns of language use and change. The corpus is also anticipated to reveal new insights into the vocabulary and language patterns of Welsh and to serve as a major resource for teaching the Welsh language to both those who have it as their first language and new speakers of it. This multifaceted impact potential has been made possible by CorCenCC's significant contribution at the methodological level, in extending the scope, relevance and design infrastructure of language corpora. Specifically, the project has involved the development of important new tools and processes, including a unique user-driven corpus design in which language data was collected and validated through crowdsourcing, and an in-built pedagogic toolkit (Y Tiwtiadur) developed in consultation with representatives of all anticipated academic and community user groups (for a detailed discussion of CorCenCC's user-driven design, see Knight et al., 2020 – details in section 10.2. below).

## 2.2. Project team

The CorCenCC project involved 4 academic institutions (Cardiff, Swansea, Lancaster and Bangor Universities), 1 Principal Investigator (PI – Dawn Knight), 2 Co-Investigators (CIs – Tess Fitzpatrick and Steve Morris) who made up, with the PI, the CorCenCC Management Team (CMT), a total of 7 other CIs (Irena Spasić, Paul Rayson, Enlli Môn Thomas, Alex Lovell, Jonathan Morris, Jeremy Evas, Mark Stonelake), 10 Research Assistants/Associates (RAs), and 180+ transcribers working over the course of the project.

In addition, there were 6 consultants, 2 PhD students, 4 undergraduate summer placement students, 4 professional service support staff and 2 project volunteers. The project also benefitted from contributions and support from representatives of a range of stakeholders including the Welsh Government, National Assembly for Wales, BBC, S4C, WJEC, Welsh for Adults, Gwasg y Lolfa, SaySomethinginWelsh and University of Wales Dictionary of the Welsh Language, via a Project Advisory Group (PAG). Nia Parry (TV presenter, producer and researcher; Welsh tutor, *Welsh in a week* (S4C)), Nigel Owens (international rugby referee; TV presenter), Cerys Matthews (Musician author; radio and TV presenter) and Damian Walford Davies (poet; professor of English and Welsh Literature; former Chair of Literature Wales) are the official ambassadors of the CorCenCC project. A full list of all individuals involved in the project can be found at www.corcencc.org/contacts - and many are referred to throughout this report, as relevant.

The project was facilitated by a strong cross-institution team, which supported staff recruitment, financial management, information technology (including equipment, software, project server maintenance and websites), media and communications outreach (including press release coordination, and radio and TV appearances), legal guidance on forms, contracts and licences, and Welsh language translators and interpreters (providing written translations of reports, key project documents and other outputs, and live simultaneous interpreting during the Whole Project Team (WPT) meetings and public dissemination events). Over 210 (bi)weekly reports were written across the course of the project, detailing work completed, issues and key risks, work to be undertaken, new ideas, thoughts and opportunities; ten Whole Project Meetings were held, and well over 100 additional meetings took place. Seven internal mailing lists were created to enable communication between team members located at different sites, along with a central, whole project Gantt chart, to collaboratively document and track deliverables and key project milestones. To maintain communication with the general public and other stakeholders, 24 editions of a project newsletter were published, circulated to individuals, and uploaded to the main website. The newsletters updated readers on data collection, reported on presentations and keynote presentations delivered (of which there were 54 in total), and introduced them to individual members of the team via our regular 'meet the team' slot. The function of this was to sustain interest and a sense of investment in the project (in line with the user-driven design). The project websites (www.corcencc.org | www.corcencc.cymru), Facebook and Twitter feeds also facilitated public engagement, amassing over 140,000 website hits and gaining 1029 followers on Twitter, 374 on Facebook (to August 2020). Although mentioned last here, the most important members of the extended CorCenCC team are the 2000+ individual contributors to the corpus.

### 2.3. Work Packages (WPs)

Work on the CorCenCC project was distributed across six coordinated work packages (WPs), each with specific tasks, aims and objectives. Led by Knight, WP0 attended to the on-going design, scoping and training activities, and involved all members of the project team. The other WPs were:

- WP1: Collect, transcribe and anonymise the data
- WP2: Develop the part-of-speech tag-set/tagger
- WP3: Develop a semantic tagger for Welsh and semantically tag all data
- WP4: Scope, design and construct Y Tiwtiadur
- WP5: Construct the infrastructure to host CorCenCC and build the corpus

While work was distributed across these work-packages, colleagues had a mutual understanding of the shared vision for the project and worked collaboratively to achieve it, with a considerable measure of interdependence between WPs that required discussion and coordination. For example, WP3 built on the research undertaken in WP1 for corpus collection, and employed WP2's part of speech (POS) tagger as a first step in the semantic analysis of the Welsh language data. WP3's output then fed into WP4 for the online pedagogic toolkit (Y Tiwtiadur), which used the multiple levels of corpus annotation to improve the engagement with and affordances of the toolkit for teachers and learners. Additionally, WP3's semantically tagged corpus fed directly into the corpus infrastructure developed in WP5. In the following sections we provide a detailed description of the WPs, outline their main aims and objectives, and reflect on their key achievements, contributions and potential applications. These descriptions were written by the respective WP leads.

## 3. Work Package 1: Collect, transcribe and anonymise the data

### 3.1. WP1: Description

The main work of WP1 concerned the sourcing, collecting and processing of the data to be included in CorCenCC. Core elements of this procedure were i) the creation of the project's sampling frame; ii) establishing transcription conventions; iii) ensuring a uniform approach to ethical compliance in the collection of the data. WP1 was co-led by Morris and Knight, who were joined by a team of Welsh speaking researchers, including CIs Evas, J. Morris, and Lovell, and RAs Needs, Rees, Arman, Watkins and Williams (at differing points throughout the project). Deuchar and McCarthy, leaders in the field of corpus linguistics, provided on-going consultative advice for this phase throughout the project and additional support was provided by a number of project volunteers and interns.

### 3.2. WP1: Objectives

The aims and objectives for WP1 were

- to design a sampling frame for the corpus
- to source and collect appropriate data
- to design and apply transcription protocols to spoken data

*Sampling frame*
The project proposal's Case for Support included an outline guide to our objectives regarding language modes (spoken, written, e-language), genres and topics, with approximately how many words of data would be collected under each heading. One of the first tasks in WP1 was to further refine and develop this outline guide. A sampling frame was created to underpin the data collection for the project, to ensure that we captured a range of different speakers across different discourse contexts and geographical locations. The sampling frame was designed to reflect current demographics of Welsh speakers using up-to-data census information (ONS, 2011). An innovative aspect of the CorCenCC sampling frame is the detailed consideration of domains in which the Welsh language is used. In a context where the great majority of Welsh speakers are bilingual and there is an uneven geographical spread in terms of density of speakers, age of speakers and language domains, the sampling frame needed to reflect the contemporary sociolinguistic situation of the language as accurately as possible.

*Sourcing the data*
The targets for, and sources of, the spoken, written and e-language data to be collected for CorCenCC were driven by our sampling frame and were shaped by initial investigation of where contemporary Welsh language is spoken and where written and e-language material is concentrated. In a bilingual context, certain domains may be under-represented (e.g. national daily newspapers). It was therefore necessary to ensure that the data was a true reflection of what is available and accessible to users of the language, rather than replicating frameworks designed for creating corpora in languages where the majority of the speakers are monolinguals.

*Transcription*
There were two preparatory steps: (i) the creation of transcription conventions for Welsh and (ii) the recruitment of transcribers (see also 3.3 below). There were particular challenges in step (i) for a number of reasons:

- In the written language, authors often denote variation in spoken varieties, so that the same basic linguistic meaning is written in many different ways. An example would be the first person singular present tense of 'to be'. In English, this might be realised as '*I am*' or '*I'm*'. Formal written Welsh would give '*Yr wyf (i)*' or '*Rwyf i*'. However, spoken Welsh produces some of the following possibilities: '*Rydw i / Dw i / Rwy / Wy / Fi*'. Writers would use these forms to represent speakers from different areas and they can be observed in literature and in other written media. The transcription conventions therefore needed to be able to reflect this reality while indexing it to the same meaning, so that searches could find the different realisations.
- Given that CorCenCC includes electronic language as well as written and spoken data, and that the conventions for representing language in the electronic context are not fully established in any language, it was necessary to be similarly accommodating in these cases.
- With regard to transcribing spoken material, the general principle adopted was to align what was heard with the closest written realisation from a set that began with the existing range of written forms but was augmented as necessary to ensure all spoken forms were appropriately captured. This principle was refined and developed through several iterations of the CorCenCC transcription conventions.

- Having adopted this general transcription principle, there was a need to ensure consistency across the CorCenCC transcriber pool. Any tendency towards prescriptivism or defaulting to formal written Welsh had to be robustly resisted.

Transcribers were recruited through campaigns particularly targeted at members of the translation profession (through their association's newsletter), students at university and those who had been involved in transcribing for other projects. Every transcriber had to pass a preliminary test piece (adhering to the CorCenCC transcription conventions) and quality was maintained through randomised checking of 25 per cent of all transcribed work with remedial corrections made where issues were identified.

## 3.3. WP1: Achievements

*Sampling frame*

Table 1 presents the initial sampling frame that was proposed in the project proposal's Case for Support. These initial distributions were based on the desirability of raising the representation of extemporised language (spoken and e-language) relative to prepared language (written).

***Table 1.*** Initial sampling frame for the CorCenCC corpus (taken from the Case for Support).

| Type | Example sources (approximate) | Words | Total |
|---|---|---|---|
| Spoken | Welsh learner discourse | 600,000 | **4m** |
| | Conversations with friends; with family; televised interviews and TV chat shows (BBC); workplace Welsh | 400,000 each | |
| | BBC radio shows; service encounters | 400,000 each | |
| | Phone calls; primary, secondary, tertiary and adult classroom interaction; political speeches; formal and informal interaction at the National Eisteddfod | 250,000 each | |
| Written | Welsh learner writing | 600,000 | **4m** |
| | Books; papurau bro (i.e. community newspapers); political documents; stories | 400,000 each | |
| | Letters and diaries; academic essays; academic textbooks; magazines; adverts; flyers/information leaflets; formal letters | 290,000 each | |
| | Signs | 60,000 | |
| E-Language | Discussion boards; emails; blogs | 500,000 each | **2m** |
| | Websites; tweets | 300,000 | |
| | Text messages | 200,000 | |
| | | **10,000,000** | |

Following the commencement of the project, more detailed sampling frames were developed, for the spoken (Table 2), written (Table 3) and e-language (Table 4) components of the corpus. This iterative refinement of the sampling frame was informed by the thematic groupings and discourse categorisations of existing major corpora including BNC 1994, the Spoken BNC 2014, CANCODE and CANELC (see Aston and Burnard, 1997, McEnery et al., 2017, Carter and McCarthy, 2004 and Knight et al., 2013). The frameworks of these existing corpora provided a useful point of departure for examining to what extent the same groupings and categorisations are usable in the minoritised language context. More fine-grained information about each of the sub-genres, and justifications for these proposed distributions, is available in Knight et al., 2021 (see section 10.2.).

***Table 2.*** Broad revised sampling frame for the Spoken element of the CorCenCC corpus.

| Contexts | % of sub-corpus | Word count |
|---|---|---|
| Cyhoeddus/Sefydliadol / *Public/Institutional* | 10% | 400,000 |
| Cyfryngau / *Media* | 15% | 600,000 |
| Trafodol / *Transactional* | 10% | 400,000 |
| Proffesiynol / *Professional* | 10% | 400,000 |
| Pedagogaidd / *Pedagogical* | 10% | 400,000 |
| Cymdeithasu / *Socialising* | 22.5% | 900,000 |
| Preifat / *Private* | 22.5% | 900,000 |
| | **100%** | **4,000,000** |

***Table 3.*** Broad revised sampling frame for the Written element of the CorCenCC corpus.

| Sources | % of sub-corpus | Word count |
|---|---|---|
| Llyfrau / *Books* | 41.75% | 1,670,000 |
| Cylchgronau, Papurau Newydd, Cyfnodolion<br>*Magazines, Newspapers, Journals* | 19.25% | 770,000 |
| Deunydd amrywiol / *Miscellaneous material* | 39% | 1,560,000 |
| | **100%** | **4,000,000** |

***Table 4.*** Broad revised sampling frame for the e-Language element of the CorCenCC corpus

| Sources | % of sub-corpus | Word count |
|---|---|---|
| Blog | 30% | 600,000 |
| Gwefan / *Website* | 30% | 600,000 |
| Ebost / *Email* | 20% | 400,000 |
| Negeseuon Testun Electronig Byr / *Short Electronic Text Messages* | 20% | 400,000 |
| | **100%** | **2,000,000** |

While the sampling frame acted as an approximate guide for data collection - an 'ideal' as it were - it is rare that a final, completed corpus, mirrors the composition of the sampling frame (for further discussions of this see Hawtin, 2018). A variety of factors influenced the final composition of the corpus, including access to specific individuals and/or data types, permissions and more practical issues concerning the amount of time it takes to process specific types of data, and the extent to which this is predictable. Once all of these factors had played out, the composition of the corpus was as seen in Table 5.

The corpus is over 11,000,000 words in size, but the composition has changed so that the spoken element contains just over 2,800,000 words. While smaller than originally planned, this sub-corpus, *in itself*, is the biggest naturally occurring spoken Welsh corpus in existence.

For a detailed breakdown of the corpus (including specific contexts, genres and topics, and demographic metadata categories and their definitions) see Knight et al., 2021 (see section 10.2.).

Note that the online corpus query tools provide an overall count of 14,338,149 **tokens** in the corpus, and makes calculations based on this value. Tokens are the smallest unit contained within a corpus, which includes words (i.e. items starting with a letter of the alphabet) and nonwords (i.e. items starting with a character that is not a letter of the alphabet). Corpora, therefore, always contain more tokens than words. The values discussed in this chapter are based on words only as this arguably provides a more accurate account of the **units of meaning** contained within the corpus.

***Table 5.*** Final composition of CorCenCC.

| **SPOKEN** | | | |
|---|---|---|---|
| **spoken_context** | **No. of texts** | **No. of words** | **Total** |
| broadcast | 564 | 750,078 | 1,332 texts<br><br>2,865,095 words |
| educational | 136 | 296,709 | |
| private | 93 | 245,719 | |
| professional | 80 | 477,983 | |
| public or institutional | 137 | 433,361 | |
| social | 131 | 456,487 | |
| transactional | 191 | 204,758 | |
| **WRITTEN** | | | |
| **written_genre** | **No. of texts** | **No. of words** | **Total** |
| academic_journal | 10 | 304,447 | 707 texts<br><br>3,940,082 words |
| book | 137 | 1,928,582 | |
| essays_coursework_and_exams | 31 | 26,047 | |
| leaflet_document_announcement | 341 | 806,030 | |
| letter | 53 | 12,873 | |
| magazine | 80 | 329,203 | |
| miscellaneous | 5 | 8,251 | |
| newsletter | 33 | 78,803 | |
| papur_bro | 13 | 117,334 | |
| thesis | 4 | 328,512 | |
| **E-LANGUAGE** | | | |
| **elanguage_genre** | **No. of texts** | **No. of words** | **Total** |
| blog | 48 | 2,345,909 | 9,397 texts<br><br>4,402,003 words |
| email | 781 | 141,554 | |
| SMS | 8,487 | 93,541 | |
| website | 81 | 1,820,999 | |
| | **11,436** | **11,207,180** | |

*Sourcing data*

When recruiting contributors of spoken data, the aim was to ensure representation from all areas of Wales. Spoken data was sourced via two main approaches: (i) recruitment of participants to be recorded and (ii) recruitment of participants to contribute spoken data via the CorCenCC app (see also 3.3.). The scope of (i) included not only research assistants going into the field to record speakers but also participants recording themselves in various interactions. This was facilitated through a network of local 'champions' (active language animateurs in targeted areas) or the *Mentrau Iaith* (each local authority in Wales has an associated *Menter Iaith*, i.e. community-based organisation dedicated to raising the profile of the Welsh language local language initiatives). Recruitment for (ii) was achieved by publicising the app (for example through social media, television appearances and publicity materials) to endeavour to reach a different cohort of participants who would be recording individually and in more private domains. Large Welsh language events such as the National Eisteddfod and Tafwyl provided opportunities for the team to reach a large cross-section of participants as well as raise general awareness of the project.

Recruiting participants to contribute using the CorCenCC phone app proved challenging. The app was made available on IoS, Android and via a web-interface (to accommodate those who did not have access to a mobile phone), and campaigns in the media e.g. appearance on television programmes such as S4C's *Prynhawn Da*, on both Welsh and English medium radio and through local engagement events generated much initial enthusiasm. However, this did not translate into a great deal of uptake of the app. Feedback from partners at the *Mentrau Iaith* suggested that people might be disproportionately concerned about being identifiable (despite frequent assurances of anonymisation), due to the relatively small size of the language community.

Promotional material (aimed at encouraging participation but also as an effective way of raising awareness of the project) included pens, coasters, leaflets and postcard size information sheets. An 'unofficial' mascot - based on a cat called Cor-pws - was designed to facilitate the participation of those under 18 and proved popular with contributors of all ages. Facebook and Twitter accounts for CorCenCC were set up in the first months of the project to further enhance the recruitment and participation of contributors.

In terms of written data, the good relationship forged at the beginning of the project with Welsh language publishers such as *Gwasg y Lolfa* led to the incorporation into the corpus of many up to date novels and books. A unique source of written data in the Welsh language is the locally based *Papurau Bro* (i.e. local community Welsh-language newspapers). It was decided to work with our local *Mentrau Iaith* contacts to collect these. Fairly rapid data capture, for example, sampling from the Welsh language academic journal *Gwerddon* through the *Coleg Cymraeg Cenedlaethol* and adult L2 pedagogical resources / examination papers through the *Welsh Joint Education Committee* resulted from our engagement with other project stakeholders in the planning process for the project.

Regarding e-language data, we were unable to collect data from Twitter or Facebook accounts because of issues of ownership restrictions, but website owners and blog authors cooperated generously and targets were exceeded. SMS messages proved more difficult to capture (much in the same way and for the same reasons as data collection through the app) but contributions through a dedicated WhatsApp number proved easier to elicit. Similarly, the collection of personal emails proved challenging while the collection of e-based workplace correspondence was easier to carry out. Contracts with BBC Cymru/Wales and S4C enabled the inclusion of samples of contemporary television and radio material, including podcasts. Notably, strong working relationships were developed with these institutions that led to them also supplying workplace emails, newsletters etc. for the written material.

Ethical approval for all aspects of data collection was obtained from all four of the universities involved in the project. Permission forms signed by participants included their agreement for the collection of important metadata (e.g. age, gender, geographical location) necessary for the corpus. For a detailed discussion of the ethical considerations/challenges encountered when constructing CorCenCC, see Knight et al., 2021b.

*Transcription*

Although transcription conventions for Welsh have been created for other projects (e.g. Deuchar et al., 2014), it was decided to create a bespoke set of conventions for the transcription of the CorCenCC data. These enabled us to fully reflect the whole spectrum of dialect/register variation captured in our speech data (making them more useful to academic researchers) as well as more accurately representing the speech of participants itself. Specifically, without an accurate representation of differences, it would not be possible to capture variations in Welsh, nor the distance between many spoken varieties and standard written Welsh. Transcribers were specifically instructed <u>not</u> to correct spoken patterns which might be considered non-standard or which included code-switching (i.e. switching between different languages during a communicative episode). Y Tiwtiadur (the pedagogic toolkit - see WP4) demanded that we should be able to identify *iaith anweddus* (inappropriate language) so that this could be systematically excluded from corpus applications used with children, for example, so transcribers were instructed to mark possible instances for review.

The recruitment of transcribers was an on-going challenge. As mentioned above, all prospective transcribers were given an initial test in which they had to transcribe a short piece according to the CorCenCC transcription conventions. A detailed overview, and rationale for, the decisions involved in the development of the transcription conventions for CorCenCC is available in Knight et al., 2021b.

## 3.4. WP1: Key contributions

The main contribution of WP1 is the eleven million words of data that form the core of the corpus. In addition, the following range of resources, created as part of the data generation process, help achieve one of the stated aims of the CorCenCC project: to increase capacity and expand the interface between the Welsh language (and by extension other minoritised languages around the world) and the discipline of applied linguistics (including, in particular, corpus linguistics, sociolinguistics, and language planning and policy):

- A sampling frame for the creation of a general corpus of a minority language;
- A definition of 'inappropriate language' suitable for the context of a minority language where speakers are all bilinguals;
- A bespoke set of transcription conventions which can be applied to contemporary spoken Welsh;
- A team of Welsh-speaking research assistants who have been trained in the principles of corpus creation, who have had the opportunity to work with international experts across the world, and who are able to apply their skills to future projects.

The work from the WP1 team has been disseminated at a range of international and national conferences, and the publications (e.g. Knight et al., 2021a, Knight et al., 2021b) give more detail of the theoretical/methodological contributions of CorCenCC (see section 10 for references).

### 3.5. WP1: Applications and impact

The process of planning and implementing the core areas of work within WP1 offers a template for those researching other minoritised or minority languages. Feedback at international conferences confirmed awareness on the part of researchers elsewhere of the transferability of the CorCenCC methodology and processes to other language projects (e.g., corpus planning for Irish and Maltese, in recent conferences).

The corpus compiled under the auspices of WP1 will allow academic research into contemporary trends in the Welsh language. Examples of questions that might be addressed, given our data collection and transcription protocols, include:

- Are mutation patterns changing in contemporary Welsh and if so, which domains and genre are in the vanguard and rear-guard?
- How do the forms used in Welsh e-language compare to those in (various varieties of) spoken and written language?
- What does the corpus tell us about the current state of the 'traditional' Welsh geographical dialects and of new emerging varieties, including their social distribution?
- What words and phrases not previously formally recorded as part of Welsh are found in the corpus, and can any trends be identified?
- How prevalent is code-switching in spoken and e-language data and what appears to trigger it?

CorCenCC will reveal much about the current vibrancy and vitality of the Welsh language. We must be prepared for elements of these revelations to be welcomed and for other elements to be resisted by the wider Welsh-speaking community. Ultimately, should there be a community desire for it, CorCenCC can pave the way for an evidence-based discussion on what constitutes - or might constitute – 'standard Welsh' in the twenty-first century.

## 4. Work Package 2: Develop a part-of-speech tagset and use it to tag the WP1 data

### 4.1. WP2: Description

The purpose of WP2 was to operationalise a suitable means of identifying and labelling ('tagging') the parts of speech (e.g. noun, verb, and subtypes of such categories) featuring in the language collected in WP1, so that the corpus could be searched and analysed in the various ways that future users would require. The tools required were a part-of-speech tagger and a tagset. An existing resource, the Bangor Autoglosser (Donnelly and Deuchar, 2011), was used as the starting point, due to its demonstrable reliability and suitability for tagging Welsh. Under the guidance of WP2 leads Knight and project consultant Donnelly (a computational linguist who has worked closely on developing Welsh corpora), Neale (one of the project RAs) applied the Bangor Autoglosser to the emerging WP1 dataset, and identified where the tools needed adaptation and refinement—largely in relation to tagging spoken and e-language sources, and the wider range of regional and genre variation. The modifications were applied in the latter stages of the project by Tovey-Walsh (a project RA), resulting in CorCenCC's own tagger and tagset, CyTag (see 4.3 below).

## 4.2. WP2: Objectives

The aims and objectives for WP2 were

- to construct and train the Welsh part-of-speech tagger
- to develop an appropriate tagset
- to tag all data

## 4.3. WP2: Achievements

CorCenCC's bespoke part-of-speech (POS) tagging software, CyTag, was developed in the first eighteen months of the project, and was publicly released in March 2018. Evaluations and applications of CyTag to date indicate that it is complete, robust, and working well. CyTag leverages open-source materials to help in the process of deciding on parts-of-speech. It works primarily by using the information found in Donnelly's Eurfa – the largest open- source, freely available dictionary for Welsh (Donnelly, 2013a) – to produce a list of possible tags for each word in a Welsh text. This is supported by specific lists of place names, first names and surnames extracted from Wikipedia data.

Once a list of words has been produced, a set of bespoke rules can then be applied to prune the list of possible tags for each word, based on the tags or features of its neighbouring words, until we arrive at the correct one. For example, the form 'yn' can mean 'in' but it also has two roles as a grammatical particle. In one role, it converts an adjective to an adverb (e.g. *yn dda* 'well', from *da*, 'good'). In the other, it is associated with the verb *bod* ('to be', one form of which is *mae* 'is') to introduce a verb, noun or adjective complement (e.g. *mae'r llyfr yn dda* 'the book is good'). As can be seen from these examples, *yn dda* can mean both 'good' and 'well', but in both cases *yn* would be classified in the same way. The tagger needed to distinguish *between* yn as a preposition and *yn* as a particle, and does so in the following way. In the sentence *mae Cymru yn wlad Geltaidd* ('Wales is a Celtic country') Cytag correctly tags 'yn' as a particle introducing a complement, because it is preceded by *mae,* and because 'wlad' is a soft mutation of 'gwlad' ('country'); we have a rule to select the complement particle tag for 'yn' if the following word is a soft- mutated noun. (When 'yn' means 'in' it is followed by a nasal mutation). The results of the evaluation process showed that CyTag was achieving an accuracy level of more than 95%, which is comparable with the best of the part-of-speech taggers for other languages. CyTag currently contains a text segmenter, a sentence splitter, a tokeniser, and the POS tagger itself. The website for CyTag is available at: http://cytag.corcencc.org.

The CyTag part-of-speech (POS) tagset contains 145 fine-grained, i.e. 'rich' tags, which are mapped into 13 categories compliant with the Expert Advisory Group on Language Engineering Standards (EAGLES, 1996). These tags include major syntactic categories (e.g. noun, article, preposition, verb and so on) as well as two categories representing 'unique' particles to Welsh and 'other' forms such as abbreviations, acronyms, symbols, digits etc. The full set of 145 tags covers Welsh morphology based on gender (masculine or feminine), number (singular or plural), person (first person, third person, etc.) and tense (past, present, future, etc.). The tags themselves are encoded in Welsh.

The full tagset can be accessed at: https://cytag.corcencc.org/tagset?lang=en. See Neale et al., 2018 for a thorough technical overview of CyTag and an in-depth evaluation of its accuracy.

One important benefit of the CorCenCC approach to developing tagging software is its use of minimal, easily adaptable rules applied to existing knowledge and resources. This method is transferable to languages for which preannotated training data is scarce, making it valuable for capturing features of minority languages.

## 4.4. WP2: Key Contributions

The main contribution of the WP2 work was the creation of freely available software tools and linguistic resources which significantly extend the existing resource for Welsh language analysis and text mining. Specifically, we produced the following:

- The CorCenCC POS tagset (https://cytag.corcencc.org/tagset?lang=en), with:
    - 145 ‘rich’ POS tags
    - 13 EAGLES-compliant ‘basic’ categories

  This tagset can either be adopted as a standard (i.e. set of conventions) and/or further enriched and/or adapted by future users when tagging Welsh language datasets.
- A gold-standard evaluation corpus. A gold-standard corpus is one that has been manually annotated and checked by multiple individuals. This effectively provides a model that can train and evaluate the automated (computerised) approach. This gold-standard evaluation corpus has also been released for other researchers to use in the development of their own tools. It comprises
    - 611 sentences
    - 14,876 tokens
- The CyTag website: (https://cytag.corcencc.org)
    - This website contains bilingual (Welsh and English) information about the tagger, including a working demo (see Figure 1a and 1b for screenshots in Welsh and English respectively). Users can access the tagger via this site and use it to tag their own data.
- CyTag on Github (https://github.com/CorCenCC/CyTag)
    - The open-source tagger has also been made available to all (i.e. available as free software, under the terms of version 3 of the GNU General Public License) via the GitHub website. Here, users can again download and use the tagger. They can also make improvements to the tagger and share updated versions with other, future, users.

The current version of CyTag includes the following improvements which were made by Tovey-Walsh:

- a revised version of the tokenizer

- an update to the way in which mutations are handled, improving its ability to capture mutation in shorter words and proper nouns
- inclusion of words with a leading lower-case character in the lexicon (e.g. ‘cymraeg’)
- inclusion of high-frequency English words, enabling them to be identified by the tagger (as non-Welsh words), reducing the number of words being tagged as ‘unknown’
- inclusion of Welsh determiners and indefinite pronouns in the tagset (e.g. ‘neb’ (=nobody), ‘pob’ (=every)) and tagging module, and added these rules to CyTag’s constraint grammar

***Figure 1a.*** A screenshot of the CyTag online demo interface in Welsh

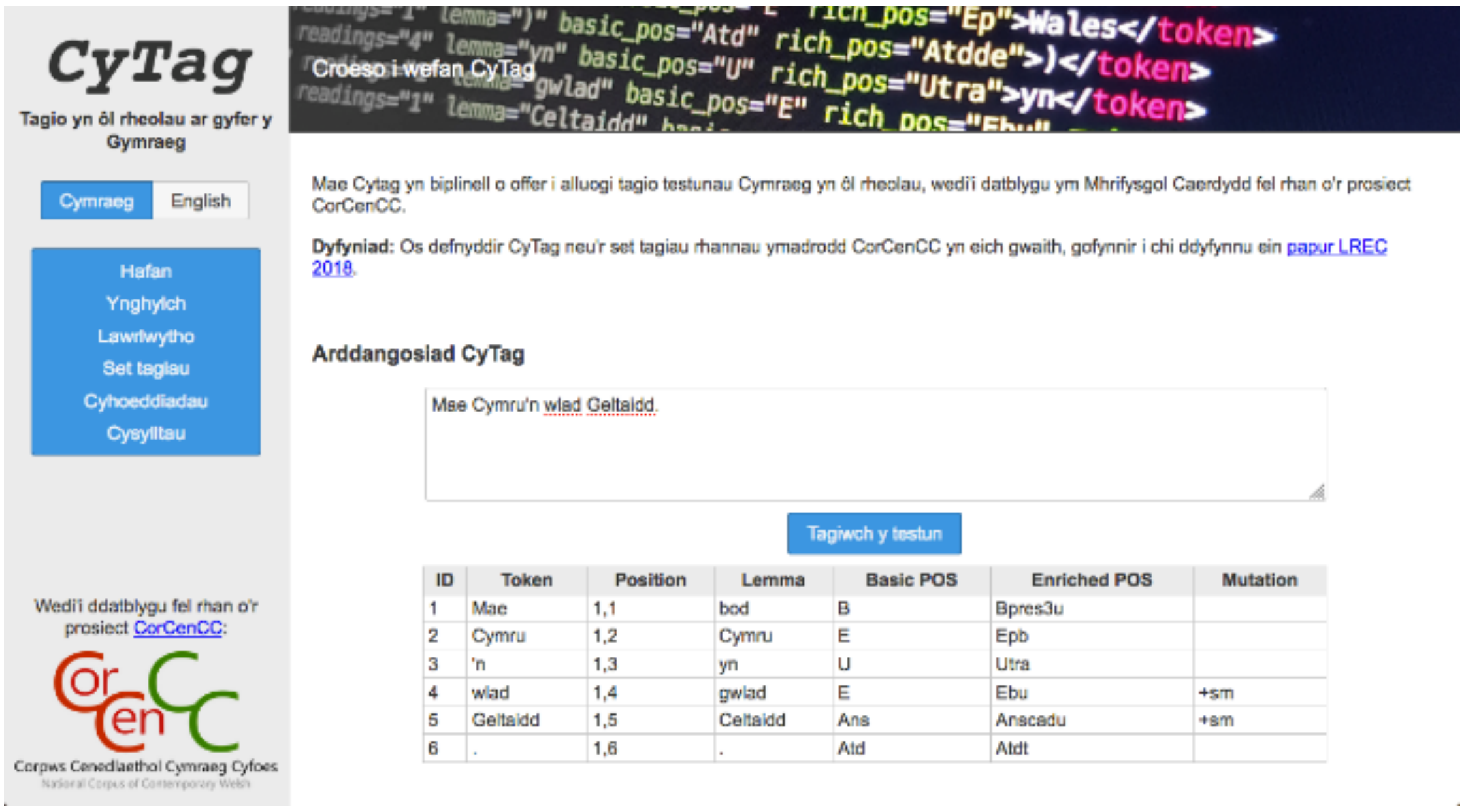

***Figure 1b.*** A screenshot of the CyTag online demo interface in English

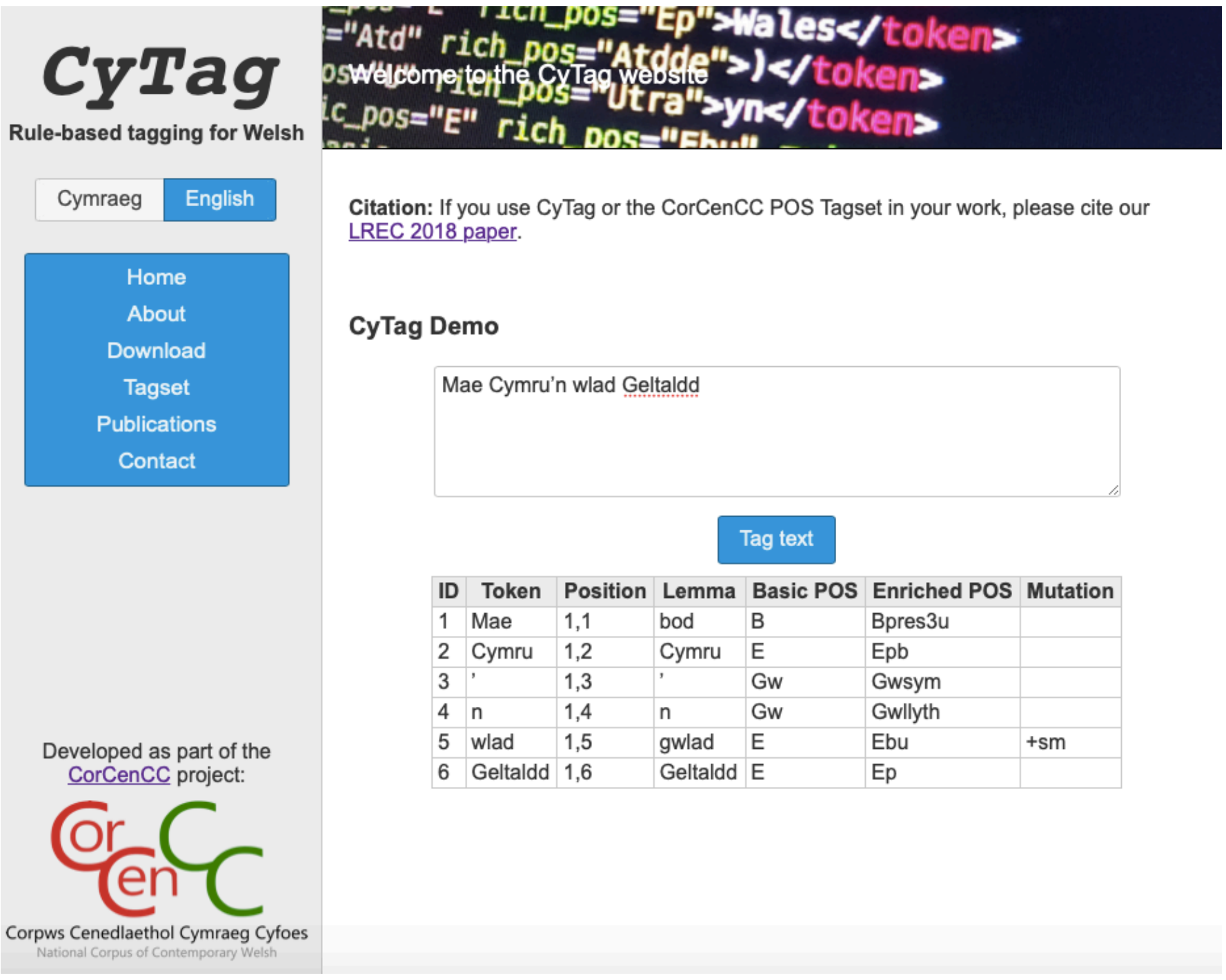

### 4.5. WP2: Applications and impact

Cytag is available in its own right as a tagger. This means it can be used by anyone who knows how to use a tagger, to tag their own data. Cytag has been pre-applied to the entire CorCenCC corpus, so that users do not need to know how to use a tagger to derive important information about the language. Such users might include researchers, language teachers, technology developers and lexicographers. CyTag can also be improved by other users in the future.

## 5. Work Package 3: Scope and develop a semantic tagger for Welsh and use it to semantically tag the WP1 data

### 5.1. WP3: Description

Work package 3 (WP3) was led by Rayson, with Piao acting as Research Associate until August 2018 (when he moved up to a fulltime lecturing position) at which point Ezeani joined the team. WP3's team drew on Welsh language expertise from across the whole CorCenCC project community where required. The major element of the research in WP3 was to design and develop the Welsh semantic annotation software system that would feed into associated linguistic resources.

### 5.2. WP3: Objectives

There were four aims and objectives for WP3:

- to design a novel semantic tagset for Welsh
- to develop the automatic annotation system
- to road-test crowdsourcing methods for semantic tagging
- to semantically tag all data

The research conducted in WP3 built on 30 years of work on automatic semantic analysis in corpus and computational linguistics in the UCREL research centre at Lancaster University. Of particular value for CorCenCC was the extensive work done on other languages (Finnish, Russian, Chinese, Dutch, Italian, Portuguese, Spanish, Malay).

As part of the CorCenCC project, we needed to re-evaluate the existing UCREL Semantic Analysis System (USAS) tagset in order to accommodate the special characteristics of Welsh, and the practical requirements of the pedagogic toolkit development (WP4) and the corpus end users (WP5). We also aimed to develop novel algorithms and methods to assign contextually appropriate semantic fields to Welsh lexical units, both single words and multiword expressions. Also, to complement the crowdsourcing method for corpus data collection in WP1, we aimed to pilot crowdsourcing methods for the assignment of semantic fields so as to extend the underlying semantic dictionaries and enable them to reflect the interpretations of Welsh speakers.

### 5.3. WP3: Achievements

The first task was to create a USAS tagset for Welsh language. This was achieved by drawing on methods developed in a previous AHRC-funded project, SAMUELS (AH/L010062/1), and updating the Java framework for Welsh. The novel semantic analysis in Welsh involved the following steps.

A mapping across semantic fields from the existing multilingual framework was undertaken, so as to review its suitability for the Welsh language. The potential meanings assigned to tags were, therefore, first derived automatically by converting English dictionaries through bilingual dictionaries and small parallel corpora, and then checked by Welsh speakers from the CorCenCC team, and modified as required. The novel semantic tagset for Welsh was released in April 2016.

Our research on crowdsourcing entailed the creation of a method and experiments involving Amazon Mechanical Turk (AMT) users to investigate whether untrained non-professional Welsh speakers could create a reasonably high-quality semantic lexicon entry by assigning one or more suitable pre-specified semantic fields to a word or phrase from the corpus. This research was published in Rayson and Piao, 2017 (see section 10.2).

A part-of-speech tagger was needed for us to work on WP3, and in order to make progress with this before WP2 was complete, we temporarily used a pre-existing part-of-speech tagger, a component of the Welsh Natural Language Toolkit (WNLT) as part of the new semantic tagger for Welsh created in Java (CySemTag). Later, when the output from WP2 was available, we adopted CyTag. Initially a (SOAP) API website was made available on the UCREL GitHub account, enabling users to access the application. A new (REST) API was later made available at http://ucrel-api.lancaster.ac.uk/, to again increase accessibility to this application.

A Java framework for single word and multiword expression tagging was created by Piao, and using a variety of methods the linguistic resources were created. The final resources contain 143,287 single word entries and a collection of sample multiword entries, plus 329,800 inflectional forms from various corpora. Both web-based (using a SOAP API) and desktop GUI versions of the Welsh semantic tagger were created. In co-operation with the wider CorCenCC team, a gold standard corpus was manually checked for evaluation and tagger improvement purposes (details of the gold standard corpus are provided in 4.4).

To complete our research in this work package, Ezeani (RA) undertook a multi-task learning experiment to investigate whether state-of-the-art vector-based word embedding models for low-resource languages (in our case Welsh) could be used with neural network models for POS and semantic tagging. Our results showed that such an approach to tagging compared very well with the existing taggers (see Rayson and Piao, 2017 – section 10.2.) All our taggers and linguistic resources have been made available as open access with permissive licences.

## 5.4. WP3: Key contributions

A key contribution of the WP3 research is the freely available software tools and linguistic resources which augment the resource bank for Welsh language analysis and text mining. The CySemTag Java code is released on GitHub and has been incorporated into the Wmatrix (Rayson et al., 2004) corpus annotation and analysis system. This system is very widely used in corpus linguistics, and means that future researchers can use it for Welsh corpora. Overall the work in WP3 has extended the scope of research in corpus and computational linguistics in at least two ways. Firstly, it has demonstrated a method for effectively extending semantic analysis techniques to the specific challenges of the Welsh language. Secondly, it has shown that crowdsourcing methods can be used to contribute to the development of such resources.

### 5.5. WP3: Applications and impact

The semantic tagger developed in WP3 has been applied to the full CorCenCC corpus, generating a rich interpretation of the data that will be of value to those interested in how contemporary Welsh is used to make meaning across genres, styles and mediums, and how language processes can be automated for new technologies. Furthermore, the underlying principles established in creating the CySemTag provide the basis for future extensions so that other low-resource languages can be analysed. Similarly, future researchers will be able to extend the multi-task learning experiments to further languages to investigate whether those languages, notwithstanding their differing grammatical frameworks, can also benefit. Incorporating the CySemTag into Wmatrix (Piao et al., 2018) extends its reach across the international research community, enabling the automatic content analysis of Welsh language corpora that might be collected by others in the future.

## 6. Work Package 4: Scope, design and construct an online pedagogic toolkit

### 6.1. WP4: Description

WP4 was led by Thomas and Fitzpatrick, working with project RAs Needs and J. Davies, and project lead Knight, with consultative advice provided by Stonelake (CI), Anthony (project consultant - designer and developer of AntConc) and Cobb (project consultant – developer of Compleat Lexical Tutor) and E. Davies (WJEC).

One field where corpora can be particularly informative is language learning/teaching. As teachers and learners become more adept with the use of technology, and as corpora continue to develop in size and application/functionality, corpus-informed learning is gaining a lot of ground both in classroom-based settings and in personal study. Corpora can be used to highlight the most common words, phrases and patterns in a language. They can show which words tend to go together, and which ones occur in which types of text (e.g. formal written texts, spoken conversations, professional e-mails, or personal text messages). Corpus users can search for specific words and see them in example sentences. Corpora therefore provide a rich source of language for the learner that demonstrates how their target language is actually used in practice, in various domains.

An innovative aspect of the CorCenCC project, led by the WP4 team, has been the development of a series of bespoke tools - Y Tiwtiadur - to be used within and outside Welsh language classes ranging from primary school to adult education. Together, these tools help demonstrate how Welsh is really used in four distinct corpus-based exercises that draw on data collected in WP1 and the taggers/tagsets developed in WP2 and WP3. The tools are:

- a Gap Filling (Cloze) tool allowing teachers (or learners, in self-study contexts) to delete words from any text in the corpus, at specified intervals to encourage or assess comprehension abilities and prediction strategies

- This tool allows users to create a gap fill task using texts from the CorCenCC corpus. The "Text type" option enables users to select particular genres of text (e.g. 'blog' or 'book-fiction' genres). The "Gap frequency" setting allows users to set the gap to appear as often as is desired, depending on how difficult the task needs to be (the recommended setting is every 7th - 9th word). Using the "Text length" option, users can choose to see a random sample of a text up to 100, 200, 300, 400, or 500 words long. On clicking "Start", a new panel shows the gap fill task with the words that have been removed from the text appearing in a separate panel. To complete the task, users are required to choose words from the list and type them into the appropriate gaps in the text. When "Check" is clicked, the correctly placed words are highlighted in green, and the incorrectly placed words are highlighted in red. The availability of texts from the corpus enables teachers and learners to undertake this activity many times, with new opportunities for learning on each occasion.

- a Word Profiler tool that enables the grading of texts by word frequency
  - This tool profiles a text selected or created by the user according to word frequency. Users are required to copy and paste a text into the "Input Text" area or type a text directly into the area. Clicking "Start" creates the profile, where each word is categorised according to its frequency level. In a separate panel, an explanation of the results is provided. The "Level"/"Frequency band" columns relate to the number of times a word appears in the 10-million-word CorCenCC corpus. Words in the "K1" (Top 1000) band are the 1000 most commonly used words in Welsh, according to CorCenCC. Typically, the more words the text has in the lower frequency bands (e.g. those in 3001-4000 (K4), 4001-5000 (K5) and >5001 (K6)), the more challenging it will be for the learner to comprehend. Learner-generated texts can also be profiled; typically, learners acquire more high-frequency words initially, and develop mastery of lower-frequency bands as proficiency develops (see e.g. Nation 2001). In the default setting, the tool will highlight words in levels K1 to K6+. Users can change the tool to highlight words that are not in these levels by clicking on the "Highlight non-level words" option. Words in the 5001+ band may include misspelled words and words from other languages, as well as vocabulary that is used infrequently in the corpus or is not captured in the corpus.
- a Word Identification tool testing learners' ability to guess a word in context
  - This tool displays multiple corpus extracts (concordance lines) that all contain a particular word. The word is blanked out, and the task is to identify the word that fits into all the gaps. There are options to select the "Frequency band" of the word (K1, K2 and K3), the specific "Word type" (e.g. noun, verb) and for the maximum number of sentences to be displayed. To generate the extracts, users click on "Start". In this tool, the 'correct' responses given are those contained in CorCenCC; in some instances, a different response may also be plausible within the language more generally.
- a Word Task Creator tool that facilitates intensive work on a specified vocabulary item

- This tool generates multiple corpus extracts (concordance lines) from CorCenCC that all contain a target word specified by the user. Users type their target word into the "Word" entry box. They then select how many extracts they want to generate (maximum 20) using the "Maximum lines" option. If users want to specify the part-of-speech of the target word (e.g. noun, verb), they do so via the "Part of speech" option. The task is generated by clicking on "Start". This tool enables two types of learning activity. One is observation regarding the words surrounding the specified one. This assists with acquisition of grammatical structures and collocation patterns. The other is a more refined version of the Word Identification tool above, whereby a teacher can specify the actual word to be guessed, rather than only a word generated by the software from a set. To facilitate this second deployment of the tool, the target word is blanked out in the results table. Clicking on "Show" reveals the target word.

These tools enable learners to work with the language in multiple ways. For example, they can explore concordance patterns across various constructions (e.g. verb + preposition, adjective + preposition, conjunction (if, as) + following tense), and guess the missing word by inspecting and analysing other words used within its immediate co-textual and contextual environment. They can identify gaps in their vocabulary knowledge, and prioritise which words to learn next. Alongside the general query tools permissible within the CorCenCC corpus, *Y Tiwtiadur* offers a unique example of Data Driven Learning (Johns, 1991) in the form of inductive, direct-use, corpus-based pedagogy (Leńko-Szymańska and Boulton, 2015) that can help supplement Welsh language learning across the lifespan.

## 6.2. WP4: Objectives

The work conducted within WP4 responded to three objectives:

- to design and production of an online Welsh-based pedagogic toolkit (described above) that works directly with the corpus data to support language teaching and learning.
- to produce frequency-based pedagogical word lists.

A key innovation of the CorCenCC project is that it integrates a corpus with an online pedagogic toolkit. Y Tiwtiadur works directly with the corpus data to support language learning and teaching by providing extensive opportunities to examine authentic Welsh language extracts. Whereas pedagogical corpus tools have hitherto been secondary adjuncts to pre-existing corpora, from the outset the design of CorCenCC included an educational interface to support Welsh language learning and teaching. Y Tiwtiadur was inspired by the online Compleat Lexical Tutor (Lextutor - https://www.lextutor.ca/ - Cobb, 2000) - one of the most high profile and frequently accessed (c.15000 users p/day) online data-driven language learning toolkits. The four tasks in Y Tiwtiadur are based on some of the most popular exercises featured in Lextutor.

A secondary innovation is that the toolkit was based on the concept of data-driven Learning (DDL - Johns, 1991), whereby 'learners inspect the evidence and look for patterns in the data from which they can form generalisations' (Thompson, 2005: 10). The four pedagogical exercises function to enable and encourage learners (first and second language) to observe and extrapolate from Welsh language patterns in order to facilitate learning. This is crucial to learner autonomy (see Aston, 2001; Little, 2007). Rather than the teacher telling the student how the language works, the students are supported in working it out for themselves, utilising the concept of inductive (or 'discovery') learning, as advocated clearly within the constructivist approach to learning. This approach contrasts with deductive (or 'tutor-directed') learning, such as when learners are provided with structural rules. Integrating learning resources into the corpus, and constructing the corpus according to learner needs, enables learners to access relevant corpus data. The four pedagogical exercises built into CorCenCC facilitate focused attention to form, which is recognised as a key element of effective language learning.

Given that CorCenCC is the first corpus of its kind to draw on a wide variety of linguistic genres, linguistic forms, and linguistic representations of Welsh, a third innovation built into Y Tiwtiadur is the ability for learners and teachers/tutors to filter both the type and quantity of corpus data in generating their resulting outputs. In order that the learning facility is applicable to all ages, one key aspect of this third innovation has been the ability to filter out texts that are likely to include inappropriate content (such as expletives). Applicable to learners of all ages is the ability to distinguish between examples of language from different data types when looking specifically at uses of complex structures that may be used variably across speakers (e.g. Welsh mutation or some noun plural forms in texts of varying levels of formality). In this way, teachers and learners can manage the potential for confusion when more than one form is in use.

## 6.3. WP4: Achievements

The development of Y Tiwtiadur was user-informed from the outset. Tutors/teachers and learners, representing a wide range of Welsh language proficiency ability levels across the different language education sectors, were consulted at different time points to allow Y Tiwtiadur to be iteratively designed, tested and improved. The aims and objectives of WP4 were pursued across three main phases: (i) a consultation phase, (ii) a product development phase, and (iii) a showcasing phase.

*(i) Consultation Phase*

During the consultation phase, a questionnaire was piloted with a small number of practitioners in the field of Welsh language teaching to explore which resources learners and teachers already used and what they would ideally like to see developed as part of Y Tiwtiadur. Based on their feedback, a more detailed and targeted questionnaire was developed for a larger audience. This was distributed to Welsh teachers and tutors at two national conferences, and later shared in an online version. In all, 44 questionnaires were returned by Welsh teachers, trainers, lecturers and tutors from a wide variety of contexts – from those who teach at primary schools (Welsh-medium and English-medium) to those who teach adults – and 10 focus groups were conducted, accessing the views of 55 teachers/tutors and 14 Welsh for Adult learners.

In addition to the questionnaire, we met teachers and tutors face-to-face, to raise awareness of the corpus and Y Tiwtiadur, and to continue to collect views that would help shape the development of the toolkit. The questionnaire responses along with follow-up focus group meetings helped us to identify the priorities for Y Tiwtiadur. The discussions we had were extremely useful, and the interchange of ideas that happened over the course of the focus groups and via the feedback obtained through the questionnaires enhanced our thinking about how to steer the work as we moved forward.

*(ii) Product Development and (iii) Showcasing Phase*

Following the consultation, the WP4 and WP5 teams collaborated to develop prototypes of the tool. This work was then developed further by J. Davies (supervised by Teahan) in collaboration with Anthony. There was an opportunity to demonstrate the work completed so far on Y Tiwtiadur at the annual conference of the National Centre for Learning Welsh in 2019, and tutors attending the workshops gave supportive and constructive feedback, including many helpful ideas with regard to usability. Their feedback informed the remainder of the development work. Positive insights included indications of how Y Tiwtiadur could be implemented with learners. Three main themes emerged:

*1. Supporting the understanding of mutation at sentence vs. lexical level*

A challenging feature of Welsh (and the other Celtic languages) is mutation – a morphophonological process whereby a phonological change is triggered in a closed set of word-initial consonants when certain words appear in particular syntactic contexts (Ball and Müller, 1992; Thomas and Mayr, 2010). For example, the initial 'c'/k/ sound in ***cath*** /kɑθ/ 'cat' undergoes a process of lenition whereby /k/ is softened to /ɡ/ - /ɡɑθ/. This change occurs after the definite article *y* "the", and after the numeral *dau* (masculine)/*dwy* (feminine) "two", for example. This phonological change is reflected in the written form, whereby *cath* is written as *gath*. In some cases, a word-initial consonant can be transformed in three ways, depending on the context (e.g. *cath* 'cat' /kɑθ/ (underlying form), *gath* /ɡɑθ/ (Soft Mutation), *nghath* /ŋ̊ɑθ/ (Nasal Mutation) and *chath* /χɑθ/ (Aspirate Mutation). These changeable forms can be problematic both in reading and writing, affecting vocabulary and literacy development/learning because they make it difficult to recognise words, and because the rules for the mutations are sometimes complicated. Corpus-driven examples can draw attention to form, and help learners identify where, when and how words change across contexts and the extent of any variation in the realisation of 'target' forms.

*2. Supplementing gaps in dictionary support*

Relating directly to mutation, as well as other changeable forms, many learners often fail to find words in paper and online dictionaries, either because they are looking for the mutated form (e.g. ***gath***) and not the underlying one (i.e. *cath*) or because they mis-spelled or mis-represented the word in text. The corpus will allow learners to discover how a form of interest to them is typically used in various types of genre and medium.

*3. Supporting learners in evaluating their writing*

One key feature of Y Tiwtiadur is the option for learners and/or teachers/tutors to use the software to code a text of their own choosing for difficulty in terms of the frequency of forms used. This can be useful in determining the suitability of a text for a given learner or a group of learners. In addition, learners can profile their own writing, so as to evaluate the extent of their own vocabulary knowledge/use.

In addition to these insights into how the tools might be implemented, stakeholders provided valuable reflections on the concerns they had about using such a resource in the classroom. Five key issues were raised, as highlighted below:

*1. Modelling 'incorrect' language*

A recurrent concern expressed among teachers and tutors from all educational contexts was that corpus data, unless 'corrected', might serve to model the very forms and expressions that teachers are trying to eliminate from their learners' attempts. Since a corpus expressly does not discriminate between what is considered correct and incorrect at a prescriptive level (other than by frequency of occurrence), teachers were advised to familiarise themselves in advance with the texts they would use in sessions, and identify anything that they wished to advise the learners about. This approach acknowledged that the relationship between descriptive and prescriptive approaches to language teaching is complex. It recognised that teachers do have a duty to inform learners about the forms that others will view as ‘incorrect’ (since such awareness is an aspect of knowledge about the language). At the same time, the manual approach would encourage teachers to question their own beliefs about what is and is not ‘acceptable’ as a target form, given the attested usage.

*2. Offensive language*

Many schoolteachers (from the Primary sector in particular) raised concerns about the possibility of accidentally accessing inappropriate content, focusing primarily on content that involved inappropriate or offensive words (such as expletives). Whilst it was beyond the scope of the current project to code for levels and types of offensiveness (which, in some instances, would not always be carried at the single word level), the corpus was tagged, at the text-level, for swear words and content of a particularly sensitive nature. While the statistical calculations that the pedagogic tools are based on reference the entire 10-million-word CorCenCC corpus, the four exercises of the pedagogic toolkit filter out these texts.

*3. Complexity of Interface*

One clear aim of CorCenCC was to develop a user-friendly corpus that could easily be transferred into education. For that to work, the program had to be fit-for-purpose and usable. This sentiment was echoed by some of the tutors and teachers we met. These views helped ensure that the four pedagogical exercises had a simple and clear interface that was intuitive to the user. At the same time, this interface mirrored the interface used by CorCenCC in order to ensure cohesion across the two and to instil confidence in teachers to progress from the pedagogic tools to a broader use of the corpus if desired.

*4. Accessibility*

Whilst the development of alternative platforms for running the corpus was beyond the scope of the present study, it was clear that in schools, in particular, where there is often limited access to computers, having an app that could be downloaded onto phones would be useful. Schools already make use of existing Welsh language apps such as Duolingo and find that platform useful, so teachers enquired as to whether a similar platform could be developed for the current project. In sum, it was clear that a valuable future adaptation of Y Tiwtiadur will be developing an app for classroom use. As this was not one of the aims of the present project, and it would take considerable research and resource to complete, it has been noted as an important consideration for follow-on work.

*5. Daunting number of examples in output*

CorCenCC is a searchable corpus of over 11 million words. This means that some searches will result in huge outputs. For those who are not experienced with corpora and the types of outputs generated, the sheer volume of outputs produced can be overwhelming, and it could deter learners and teachers/tutors. For that reason, one of the innovative aspects of Y Tiwtiadur is that it allows teachers, tutors and learners the capacity to manage the amount of output from queries (as advocated in the Data Driven Learning approach). In addition to the ability to select topics or semantic categories when exploring text, the tool provides teachers/tutors and learners with the ability to limit text length when blanking out words in the Gap Fill exercise, provides a maximum output of 20 instances for the Word Identification and Word-in-Context exercises, etc. Together, these features increase the level of autonomy for the learner and teacher/tutor so that they are able to make the toolkit work for them.

## 6.4. WP4: Key contributions

Through the development of a pedagogical interface, led by the concept of data-driven learning and assessment, WP4 has contributed (i) a new pedagogical resource that is (ii) drawn from an online corpus of contemporary Welsh, of a kind that (iii) has never existed for the teaching of Welsh before, and that (iv) can serve as a model for similar work with other minority languages. It has made an invaluable contribution to language teaching and learning and, being open source, is available to support users in their inductive learning, irrespective of age, ability level and geographical location. The resource offers a new and unique opportunity for schools in Wales to embrace the concept of Data Driven Learning and to engage with and develop their own corpus-led pedagogies. In addition to this, teachers and learners, once introduced to the corpus via Y Tiwtiadur, might also feel confident enough to explore the corpus in other ways via the main CorCenCC query tools.

Various calls have been made in recent years for a corpus that can inform the delivery of Welsh (NFER, 2008: 48; Welsh Government 2013: 27, 71; Mac Giolla Chríost et al., 2012). CorCenCC, as a contemporary corpus of Welsh with an integrated pedagogic toolkit (Y Tiwtiadur), fulfils that need, informing curriculum writing, language assessment and language learning resources in the way that similar corpora do effectively for English (e.g. the Cambridge English Corpus (CEC) informs Cambridge English Language teaching resources; the British National Corpus (BNC) informs Pearson Longman's resources). An equivalent full and independent set of Welsh language teaching resources based on CorCenCC is a potential future direction of this work.

In line with the expectations laid out in the new *Curriculum for Wales: 2022*, such a resource will help develop learners' language awareness abilities and enrich their Welsh language skills in a naturalistic way, impacting ultimately on the Welsh Government's *Cymraeg 2050: Miliwn o Siaradwyr* (Welsh Government, 2017) agenda, which aspires to a million Welsh speakers by 2050.

### 6.5. WP4: Applications and impact

The use of corpus data to support language learning in schools is a rapidly developing practice, but its effective implementation is under-developed and its effectiveness is under-researched. Within the New Curriculum for Wales 2022, children will be required to learn about the concepts of language, analyse linguistic nuances, and understand how these differ across languages. Corpus data is perfectly aligned with the promotion of such metalinguistic skills and knowledge, particularly within a bilingual context such as Wales. An important next step is to disseminate this free, on-line resource to schools in Wales. This will involve working closely with WJEC, curriculum designers and Welsh Government in identifying best practice for the use of CorCenCC in teaching and learning contexts across Wales, and modelling its implementation in other minority language contexts elsewhere. This work could lead to funded research projects evaluating the effectiveness of various applications of CorCenCC and Y Tiwtiadur with different types of learners with a view towards expanding its functionality where appropriate. For example, exercises could be adapted for a phone app or other technological platforms, increasing take-up and educational impact.

## 7. Work Package 5: Construct the infrastructure to host CorCenCC

### 7.1. WP5: Description

WP5 was concerned with the technical aspects of the construction of CorCenCC, building tools to support every stage of the process of corpus construction, from data collection (with a focus on the crowdsourcing app), through collation (via the data management tools), to querying and analysis (the web-based interface).

Spasić led WP5, working with Knight, Rayson, Piao and project RAs Neale and Muralidaran. Additional technical and consultative expertise was provided by Anthony (corpus linguist, educational technology specialist and creator of Antconc), Scannell (a computational scientist specialising in NLP, machine translation and minority languages) and Donnelly (a computational linguist who worked closely on developing previous Welsh corpora).

### 7.2. WP5: Objectives

WP5 aimed to develop a computational infrastructure to support the systematic collection and storage of this large quantity of text and analytic data together with a user-friendly interface to enable interaction with this data online. An important element was the design and construction of a repository system that would allow the adding of new data to the corpus over time, so that the maintenance of the corpus would be supported by its own users, and contributions to the corpus would be a social venture. A suite of corpus-analytic tools was developed on top of the repository to support functionalities that are typically integrated into contemporary corpora, such as KWIC (Key Word in Context) concordancers and collocation tools, search and sort tools, word frequency lists, key word analysers and statistical testing facilities.

The data collection workflow is illustrated in Figure 2. The distinction between the three main language types (spoken, written and electronic (e)) is emphasised, as they required different processing before the data could be integrated into the corpus. As indicated in Figure 2, all relevant participant information and descriptive metadata was recorded at the time of data collection. Permissions to share the data in an online public resource were essential to the development of CorCenCC. These permissions were obtained from the relevant legal entities (e.g. the copyright owner; the speaker themselves) before the data was collected and locally stored. The raw data together with the corresponding permissions and metadata were deposited into a local file storage system. Subsequently, different data formats were standardised into plain text. Plain text can be processed automatically by natural language processing (NLP) tools, should another layer of linguistic metadata need to be added later; this helps future-proof the corpus, by enabling additional information to be added.

***Figure 2***. Data collection workflow in CorCenCC

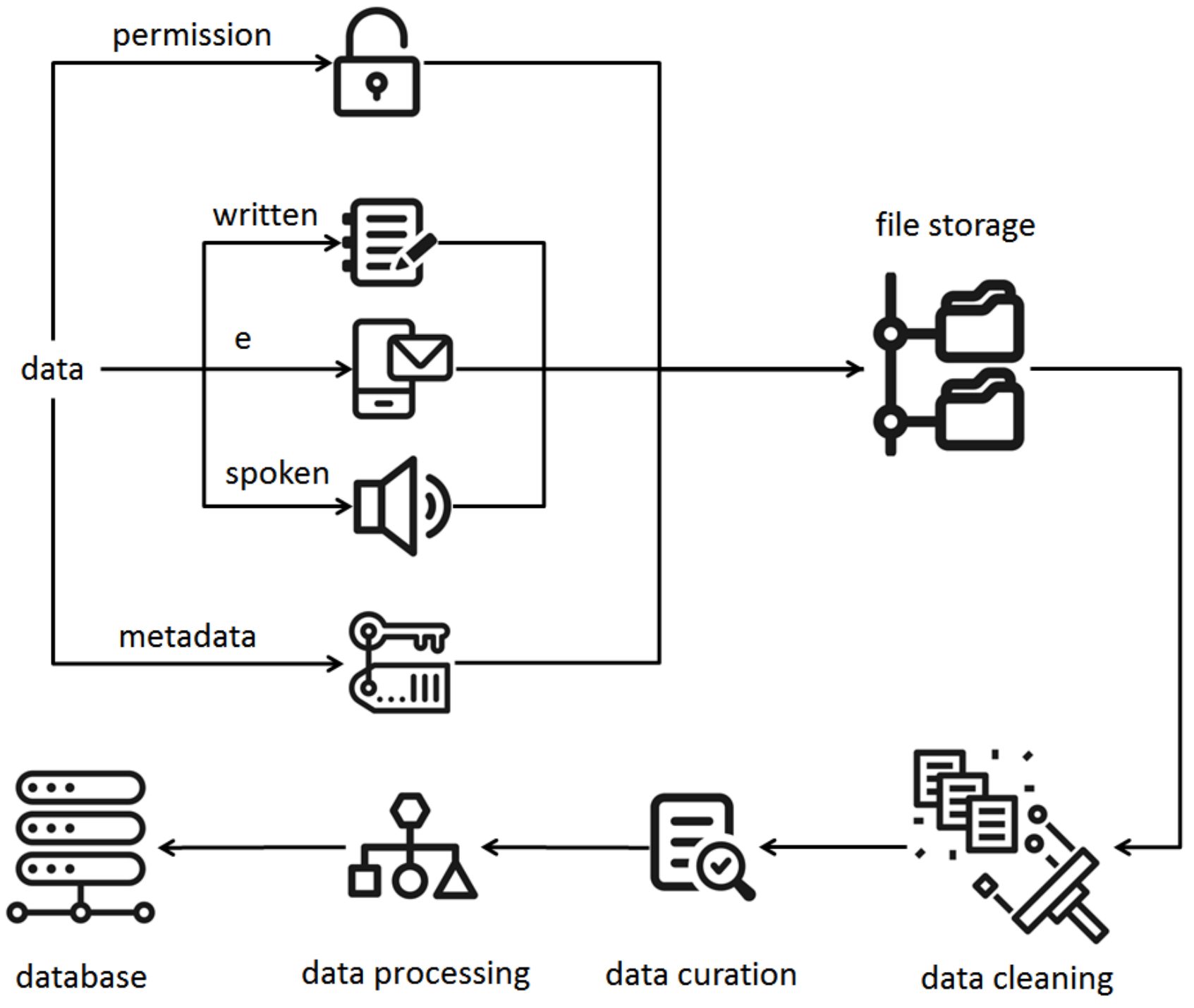

## 7.3. WP5: Achievements

One of the key innovations of the CorCenCC project was redefining the design and construction of linguistic corpora, aligning methods more succinctly with the Web 2.0 age. To this end, steps were taken to construct and evaluate a system that enables 'live' user-generated spoken data collection via crowdsourcing. Crowdsourcing is a way to gather resources (in this case linguistic examples) from the general public, by making requests for volunteers to participate. We facilitated crowdsourcing by means of an app that can be run on any Internet-enabled device; it ran in two forms: as a phone application and an interactive website. To maximise the potential user base, the mobile version of the app was implemented on both iOS (i.e. Apple) and Android platforms. The application made contributing to the corpus a very personal experience, giving users ownership and control of their own recordings.

All raw data (written, spoken and electronic) was stored systematically within a predefined folder structure, which corresponded to the sampling frame. From there, the data underwent the relevant cleaning and curation processes. To support collaborative multi-user access by the team members (from researchers to transcribers) across different sites (within and across the multiple institutions involved in the project), an online data management tool was developed on top of the file storage system. It provided a graphical user interface (GUI) that facilitated the uploading of raw data, indexing of the corresponding metadata and recording of subsequent data transformations, thus allowing the progress of all aspects of the corpus construction process to be monitored closely by all the researchers.

Once the texts had been converted to plain text format, they were marked up with layers of sociolinguistic metadata (e.g. source, genre, geographical origin) that would be used to query the data, and automatically tagged. As described in Section 4, the CorCenCC WP2 team developed CyTag (Neale et al., 2018) to achieve the tagging. CyTag is a suite of surface-level NLP tools for Welsh, based on the concept of constraint grammar (Karlsson, 1990, Karlsson et al., 1995). It supports text segmentation including sentence splitting and tokenisation as well as part-of-speech (POS) tagging and lemmatisation. It provides a bespoke solution for the basic linguistic pre-processing of Welsh including a tagset that is rich enough to capture idiosyncrasies of the language, notably in its spoken versions. To facilitate the semantic analysis of Welsh language data on a large scale, all pre-processed data was further marked up according to semantic categories using the CySemTagger developed in WP3 (see Section 5).

The corpus was stored and managed in a relational database where data could be accessed securely and concurrently by multiple users. To share the data online, we implemented a web-based interface to the database. The main reason for creating a bespoke interface rather than re-using an existing solution such as CQPweb (Hardie, 2012) was the requirement to tailor its functionality to the specific metadata of the CorCenCC corpus and its prospective users. To gather information about the user requirements, we used social media to survey current users of corpora. A total of 62 individuals responded, and their input identified the key functionality requirements.

An important consideration for the development of the corpus infrastructure was checking that it was fit for purpose. A group of corpus linguists evaluated the usability and functionality of the web-based interface. This process of evaluation involved a combination of questionnaires and talk-aloud exercises. Overall, the participants found the system useful in terms of meeting their information needs within the scope of their professional activities. The functionality was easy to understand without having to resort to help screen assistance. All participants agreed that they were likely to adopt the system and recommend it to other linguists.

### 7.4. WP5: Key contributions

The construction of a new corpus infrastructure is a major undertaking. The majority of corpus researchers use existing software to analyse the texts that they gather. Here, though, no such software existed, and it had to be built before the texts we had collected could be suitably indexed for entry into the corpus. Thus, we were simultaneously addressing several major challenges important for progressing corpus linguistic research.

Secondly, we necessarily designed from scratch much of the underlying computational infrastructure for tagging and analysing the language, given that Welsh has many features including grammatical distinctions, that do not transfer easily from languages with significant existing corpus resource (notably English), and substantial regional and register variation arising from the particular social history of the language. The key pillars of the infrastructure include a framework that supports metadata collection, an innovative mobile application designed to collect spoken data (utilising a crowdsourcing approach), a backend database that stores curated data and a web-based interface that allows users to query the data online. By using Welsh language tags, we have ensured that the corpus is not, and cannot be perceived as, an external (English) tool superimposed onto Welsh, but rather belongs to Wales and the Welsh language. Users will be encouraged by this means to buy wholeheartedly into the language as not only a source of information but also the medium through which it can be studied. At the same time, the availability of an additional English language interface will ensure an access point for the many whose interest in Welsh currently outstrips their facility with it, including the many thousands of learners of Welsh.

Thirdly, we have created tools that are freely available for others to adapt when creating their own corpora. We are particularly committed to supporting the building of corpora for other minority languages, and our user-driven model directly informs such projects by providing a template for corpus development in any other language.

Fourthly, by assembling an international team of experts, we have been able to deploy the latest technological innovations, and develop our own new ideas, lighting the pathway for future work in the Web 2.0 age. The crowdsourcing app is one of the first of its kind to be used for building a balanced corpus of natural language data by complementing more traditional methods of data collection, and successfully addressing a significant and persistent problem for the collection of high quality consented spoken data. Furthermore, we have demonstrated that it is possible to find the necessary manpower for transcribing such data and doing the essential manual tagging, even for a language with a relatively small cohort of fluent speakers.

### 7.5. WP5: Applications and impact

Though the computational infrastructure was developed for Welsh language collection, its design can be re-used to support corpus development in other minority or major language contexts, broadening the potential utility and impact of this work.

## 8. Summary of potential applications and impact

As described earlier in this document, every part of the project (as characterised by the work packages) has valuable applications that offer societal, economic and/or academic benefits. At a societal level, the corpus provides the opportunity to understand Welsh as a living language in use. In economic terms, the corpus offers scope to develop valuable new resources for Welsh learners and users, including potential for a corpus-based dictionary and a range of data-informed technological tools that might include language learning apps, predictive text production, word processing tools, machine translation, speech recognition and web search tools. Indirectly, the support of the corpus for these social and economic outcomes will promote the recognition of Welsh as a significant element of the UK and world linguistic landscape. The potential academic applications are broad and varied:

- Corpus linguistics: The innovative design of CorCenCC will inform and guide future corpus development and use in any language, via the provision of open access resources and the protocols for crowdsourcing approaches to data collection and analysis and for integrating research and teaching/learning functionalities.
- Language acquisition and bilingualism: The embedded data-driven learning facilities offer a unique opportunity for the investigation of autonomous learner behaviour and blended learning. CorCenCC incorporates a suite of frequency-based tools for research into patterns of acquisition and text and learner profiling.
- Sociolinguistics, dialectology and morphosyntactic analyses: CorCenCC extends the scope of available conversational data from the Siarad corpus (www.bangortalk.org.uk) to include speakers from a wider range of geographical regions. It will further inform us on the extent of language contact between English and Welsh. It will facilitate the exploration of sociolinguistic variation in patterns of lexical, grammatical, semantic, pragmatic and accent-based properties of language use within and across regions, by Welsh-user profile, thereby deepening the understanding of the social dynamics of Welsh as a living and reviving language.
- Language planning: CorCenCC provides base-line data with which to investigate language use in the context of policies relating to the use of Welsh in public administration, commerce and education.
- Lexicography: CorCenCC will be of direct relevance to the work of the team updating the University of Wales Dictionary of the Welsh Language (project partners), by providing primary evidence of lexical usage from attributable sources that can be included in future revisions of the dictionary.
- Computer and translation technology: As a tagged corpus, CorCenCC can be used for developing hybrid machine translation systems by feeding into a statistical machine translation engine, and facilitating the abstraction of grammatical rules from the corpus. Other areas of natural language processing (e.g. spelling correction, word prediction, assistive software for Welsh) will benefit from the more accurate and wide coverage language models that will be generated from the analysis of data from CorCenCC.
- Other research areas: CorCenCC offers new opportunities to academics engaged in stylistics and literary studies, media studies, psycholinguistics, pragmatics, business studies, health and medicine and psychology to extend their research into the Welsh language context.

CorCenCC is a freely available resource under an open licence which, when combined with the user-driven design and construction, will maximise its potential societal impact, informing the work and activities of current and future users of Welsh in a number of critical areas. Potential non-academic practitioner and professional domains include:

- Second language teaching and learning: As discussed in Section 6, reports on the teaching of Welsh for Adults (Mac Giolla Chríost et al., 2012; Welsh Government, 2013) have drawn attention to the need for a corpus of contemporary Welsh as a way to improve the efficacy of Welsh learning initiatives. CorCenCC functions to meet this need. By informing curriculum writing, language assessment and language learning resources as similar corpora do effectively in English (e.g. CEC and BNC), CorCenCC will facilitate data-driven learning, enhancing the effectiveness of teaching Welsh as a second language (compulsory in all schools in Wales up to the end of Key Stage 4). The anticipated impacts in the medium term are improved effectiveness in language learning; more nuanced awareness by teachers and learners of the inherent and natural variation in Welsh by region, genre and speaker-type; increased confidence in Welsh language speakers about the validity of their own usage patterns; and greater awareness about and pride in the Welsh language as a living and developing expression of Welshness.

- The Welsh Government and National Assembly of Wales (Language Policy): CorCenCC facilitates the realisation of action points in the Welsh Language Commissioner's strategy relating to digital content and applications, translation, terminology, language planning and research. These reflect the priorities of the Welsh Government (2014; 2017). As such, CorCenCC stands to impact the furthering of the integration of Welsh into everyday life as a language of governance, commerce and social interaction.
- The translation industry in Wales: CorCenCC outputs fit with the mid-term development of Microsoft Translate software. Preliminary research (Screen, 2014) shows that example-based machine translation alone can improve the productivity of human translators by up to 55%. By contributing to an eventual hybrid machine translation system, CorCenCC could further improve translation efficiency.
- The media in Wales: CorCenCC offers media companies the means to evaluate the linguistic nature of their output by measuring the difficulty of material (e.g. calculating the proportion of low frequency vocabulary) and establishing whose Welsh is being represented and underrepresented. By this means, CorCenCC could materially affect the accessibility and attractiveness of Welsh language programmes and media outputs to their target audiences, with consequent increased viewing/engagement figures, as well as supporting social equality.
- Welsh language publishers and lexicographers: CorCenCC provides the means to target content at audiences of different reading abilities and enhance the language tools available to authors for constructing graded readers. It will enable the commissioning of dictionaries of modern Welsh based on actual language use, thereby closing the recognised, and often problematic, gap between what Welsh users need and what they can find in reference sources. In turn this will build Welsh speakers' confidence in their linguistic abilities, making them more willing to use Welsh in a wider range of contexts.
- Language technology companies: a core requirement for companies using web-based and online social media data is a large, high-quality training corpus, and CorCenCC provides this. The CorCenCC dataset will, therefore, enable the development of a range of Welsh language technology resources that currently do not exist for the language.
- In the public domain: through its user-driven design, representatives of the likely future users of CorCenCC have been directly involved in the construction and design of the corpus which has ensured that it is user-friendly, accessible and appropriate to their needs. This approach functions to build on existing interest in Welsh language and heritage, and to foster community 'ownership' of the corpus. The potential long-term impact is a tangible change in perceptions of and attitudes to Welsh within and beyond Wales.

## 9. Associated projects and further funding

While CorCenCC itself was generously funded by the ESRC and AHRC, a range of satellite and associated research projects and other activities were funded from other sources. These are detailed below:

| Date | Funder | Amount | Description [with PI] |
|---|---|---|---|
| Jan 2017 | Cardiff University | £56,000 | College of Arts, Humanities and Social Sciences (AHSS) funding received to support a three-year PhD scholarship for Vigneshwaran Muralidaran with a study entitled 'Using insights from construction grammar for usage-based parsing' [Knight and Spasić]. |
| Feb 2017 | British Council | £2,000 | Funding to support the public launch of the CorCenCC project the Pierhead Building, Cardiff. [Knight] |
| Feb 2017 | Swansea University | £1,000 | RIAH - Research Institute for Arts and Humanities, Swansea University Funding received to support the CorCenCC project launch. [Fitzpatrick] |
| Feb 2017 | Cardiff University | £1,500 | School Research and Innovation Fund support received for the CorCenCC project launch. [Knight] |
| Oct 2017 | Welsh Government | £24,992 | Competitive commission from Welsh Government to provide a rapid evidence assessment of effective second language teaching approaches and methods. For more information see: https://tinyurl.com/ybtdsvfy [Fitzpatrick] |
| Jan 2018 | Cymraeg 2050 2017-2018 Grant Scheme GC2050/17-18/20: | £19,964 | Funding to construct a computational WordNet for Welsh. WordNet Cymru is lexical database in which words are grouped into sets of synonyms (synsets), which are then organised into a network of lexico-semantic relationships. To access the WordNet Cymru website, visit: http://corcencc.org/wncy/ [Spasić] |
| Jan 2018 | Welsh Joint Education Committee (WJEC) | £1,968 | Research grant (including intramural programme). Research grant to complete work on producing a B1 core vocabulary for Welsh for Adults (Canolradd level). For more information see: http://cronfa.swan.ac.uk/Record/cronfa48953 [Morris] |
| Mar 2018 | Swansea University | £1,200 | SPIN (Swansea paid internship) placement for data collection, transcription and interviewing of teachers/tutors 2017-18. Studentship for capacity building purposes. [Morris] |
| April 2018 | Swansea University | £57,121 | College of Arts and Humanities (COAH) funding to support a three-year PhD scholarship for Bethan Tovey-Walsh with a study entitled 'Purism and populism: The contested roles of code-switching |
| | | | and borrowing in minority language evolution'. Fees and maintenance paid [Morris and Fitzpatrick]. |
| July 2018 | Cardiff University | £2,100 | CUROP (Cardiff University Research Opportunity) internal funding for a project entitled: 'Corpws Cenedlaethol Cymraeg Cyfoes: National Corpus of Contemporary Welsh - a focus on spoken data'. Studentship for capacity building purposes. [Knight] |
| July 2018 | Cardiff University | £2,100 | CUROP (Cardiff University Research Opportunity) internal funding for a project entitled: 'Corpws Cenedlaethol Cymraeg Cyfoes: National Corpus of Contemporary Welsh - semantic tagging and data annotation'. Studentship for capacity building purposes. [Knight] |
| Oct 2018 | ESRC DTP Collaborative Studentship, Swansea University | £81,253 | Welsh and Applied Linguistics: ESRC Wales Doctoral Training Partnership PhD Studentship entitled 'Strategic bilingualism: identifying optimal context for Welsh as a second language in the curriculum'. [Morris] |
| Jan 2019 | Welsh Government | £20,000 | Funding to support the development of a Welsh language Stemmer. [Spasić] |
| Aug 2019 | Welsh Government | £90,000 | Project entitled: 'Welsh language processing infrastructure: Welsh word embeddings'. Word embeddings are a type of word representation where words or phrases with a similar meaning are mapped to vectors of real numbers. The project focused on word embeddings for Welsh (primarily on creating a lexicon and Welsh word and term embeddings) and contributes to the Welsh Language Technology Action Plan's aim to 'promote Welsh language technology and coding resources to teachers and children and others'. [Spasić]. |
| May 2020 | Welsh Government | £90,000 | Project entitled: 'Learning English-Welsh bilingual embeddings and applications in text categorisation'. This project aims to extend the results of the previous word embeddings project by creating cross-lingual representations of words in a joint embedding space for Welsh and English. [Knight] |
| | | **£451,198** | |

# 10. Summary of project outputs

## 10.1. Software tools

| Name | Details | Link |
|---|---|---|
| CorCenCC's crowdsourcing app | Designed to allow Welsh speakers to record conversations between themselves and others across a range of contexts and to upload | http://www.corcencc.org/app/<br><br>http://app.corcencc.org |
| | them, complete with ethically compliant consent from participants, for inclusion in the final corpus. Crowdsourced corpus data is a relatively new direction that complements more traditional language data collection methods, and is ideally suited to the positive community spirit that exists among speakers and learners of the Welsh language. | **Cite:** Knight, D., Loizides, F., Neale, S., Anthony, L. and Spasić, I. (2020). Developing computational infrastructure for the CorCenCC corpus – the National Corpus of Contemporary Welsh. *Language Resources and Evaluation (LREV).* |
| CyTag – Welsh Part of Speech Tagger | CyTag is an innovative Welsh tagger (complete with bespoke tagset) designed and constructed for the project. It is used in conjunction with the semantic tagger to tag all lexical items in the corpus. | http://cytag.corcencc.org<br><br>**Cite:** Neale, S., Donnelly, K., Watkins, G. and Knight, D. (2018). Leveraging Lexical Resources and Constraint Grammar for Rule-Based Part-of-Speech Tagging in Welsh. Poster presented at the *LREC (Language Resources Evaluation) 2018 Conference,* May 2018, Miyazaki, Japan. |
| CySemTag Welsh Semantic Tagger Version 1 | The Welsh Sematic Tagger applies corpus annotation automatically to Welsh language data. | http://ucrel.lancs.ac.uk/usas/<br><br>**Cite:** Piao, S., Rayson, P., Knight, D. and Watkins, G. (2018). Towards a Welsh Semantic Annotation System. *Proceedings of the LREC (Language Resources Evaluation) 2018 Conference,* May 2018, Miyazaki, Japan.<br><br>Piao, S., Rayson, P., Knight, D., Watkins, G. and Donnelly, K. (2017). Towards a Welsh Semantic Tagger: Creating Lexicons for A Resource Poor Language. *Proceedings of The Corpus Linguistics 2017 Conference,* July 2017, University of Birmingham, Birmingham, UK. |
| CorCenCC's infrastructure and query tools | The CorCenCC query tools include the following functionalities:<br>▪ Simple query<br>▪ Complex query<br>▪ Frequency list generation<br>▪ Collocation analysis<br>▪ N-gram analysis<br>▪ Concordancing<br>▪ Keyword analysis | For tools and user guide, see: www.corcencc.org/explore<br><br>**Cite:** Knight, D., Loizides, F., Neale, S., Anthony, L. and Spasić, I. (2020). Developing computational infrastructure for the CorCenCC corpus – the National Corpus of Contemporary Welsh. *Language Resources and Evaluation (LREV).* |
| Y Tiwtiadur | CorCenCC's pedagogic toolkit which is integrated with the main query tools. This includes the following teaching and learning tools:<br>▪ Gap-filling<br>▪ Vocab profiler | For tools and user guide, see: www.corcencc.org/explore<br><br>**Cite:** Davies, J., Thomas, E-M., Fitzpatrick, T., Needs, J., Anthony, L., Cobb, T. and Knight, D. (2020). *Y Tiwtiadur.* [Digital Resource]. Available at: www.corcencc.org/Y-Tiwtiadur |
| | ▪ Word identification<br>▪ Sentence-gap creator | |

10.2. Publications (by reverse date order, with project team members in boldface):

1. **Knight, D., Morris, S., Arman, L., Needs, J.** and **Rees, M.** (2021a, in prep.). *Blueprints for minoritised language corpus design: a focus on CorCenCC.* London: Palgrave.
2. **Knight, D., Morris, S.** and **Fitzpatrick, T.** (2021b, in prep.). *Corpus Design and Construction in Minoritised Language Contexts: The National Corpus of Contemporary Welsh.* London: Palgrave.
3. **Knight, D.,** Loizides, F., **Neale, S., Anthony, L.** and **Spasić, I.** (2020). Developing computational infrastructure for the CorCenCC corpus – the National Corpus of Contemporary Welsh. *Language Resources and Evaluation (LREV).*
4. Corcoran, P., Palmer, G., **Arman, L., Knight, D.** and **Spasić, I.** (2020, accepted). Word Embeddings in Welsh. *Journal of Information Science*.
5. **Muralidaran, V., Knight, D.** and **Spasić, I**. (2020, accepted). A systematic review of unsupervised approaches to usage-based grammar induction. *Natural Language Engineering.*
6. **Spasić, I.**, Owen, D., **Knight, D.** and Arteniou, A. (2019). Data-driven terminology alignment in parallel corpora. *Proceedings of the Celtic Language Technology Workshop 2019,* Dublin, Ireland.
7. **Piao, S., Rayson, P., Knight, D.** and **Watkins, G**. (2018). Towards a Welsh Semantic Annotation System. *Proceedings of the LREC (Language Resources Evaluation) 2018 Conference,* May 2018, Miyazaki, Japan.
8. **Neale, S., Donnelly, K., Watkins, G.** and **Knight, D**. (2018). Leveraging Lexical Resources and Constraint Grammar for Rule-Based Part-of-Speech Tagging in Welsh. Poster presented at the *LREC (Language Resources Evaluation) 2018 Conference,* May 2018, Miyazaki, Japan.
9. **Rayson, P.** (2018). Increasing Interoperability for Embedding Corpus Annotation Pipelines in Wmatrix and other corpus retrieval tools. Proceedings of the Challenges in the Management of Large Corpora workshop at the *LREC (Language Resources Evaluation) 2018 Conference,* May 2018, Miyazaki, Japan.
10. **Rayson, P**. and **Piao, S**. (2017). Creating and Validating Multilingual Semantic Representations for Six Languages: Expert versus Non-Expert Crowds. Proceedings of the 1st Workshop on Sense, Concept and Entity Representations and their Applications held at the *European Chapter of the Association for Computational Linguistics 2017* (EACL) conference, April, Valencia.
11. **Piao, S., Rayson, P.,** Archer, D., Bianchi, F., Dayrell, C., El-Haj, M., Jiménez, R-M., **Knight, D.,** Křen, M., Löfberg, L., Nawab, R. M. A., Shafi, J., Teh, P-L., and Mudraya, O. (2016). Lexical Coverage Evaluation of Large-scale Multilingual Semantic Lexicons for Twelve Languages. *Proceedings of the LREC (Language Resources Evaluation) 2016 Conference,* May 2016, Portorož, Slovenia.

### 10.3. Keynotes and invited talks

Research from the CorCenCC project has been presented at 17 keynotes and invited talks, and disseminated via 37 other conference papers delivered in 11 countries around the world. Details of these talks can be found on the main CorCenCC website (see: www.corcencc.org/outputs).

## References

Aston, G. (2001) *Learning with Corpora*, Athelstan, Open Library.

Aston, G. and Burnard, L. (1997) *The BNC Handbook: Exploring the British National Corpus with SARA,* Edinburgh: Edinburgh University Press.

Ball, M. and Müller, N. (1992) *Mutation in Welsh,* Clevedon: Multilingual Matters.

Brabham, D. C. (2008) 'Crowdsourcing as a model for problem solving: An introduction and cases', *Convergence* 14**:** 75-90.

Carter, R. and McCarthy, M. (2004) 'Talking, creating: Interactional language, creativity, and context', *Applied Linguistics* 25**:** 62-88.

Cobb, T. (2000). *The compleat lexical tutor* [Online], available: http://www.lextutor.ca/ [Accessed 07/07/20].

Collins. (2020). *Collins Corpus online* [Online], available: https://collins.co.uk/pages/elt-cobuild-reference-the-collins-corpus [Accessed 07/07/20].

Cooper, Jones, D. and Prys (2019) 'Crowdsourcing the Paldaruo Speech Corpus of Welsh for Speech Technology', *Information* 10**:** 247.

Cambridge University Press. (2020) *Cambridge English Corpus online* [Online], available: https://www.cambridge.org/us/cambridgeenglish/better-learning-insights/corpus [Accessed 07/07/20].

Deuchar, D., Webb-Davies, P. and Donnelly, K. (2018) *Building and Using the Siarad Corpus,* Amsterdam: John Benjamins.

Deuchar, M., Davies, P., Herring, J. R., Parafita Couto, M. and Carter, D. (2014) 'Building bilingual corpora: Welsh-English, Spanish-English and Spanish-Welsh', in Thomas, E. M. and Mennen, I. (eds) *Advances in the Study of Bilingualism.* Bristol: Multilingual Matters.

Donnelly, K. (2013a) *Eurfa v3.0 - Free (GPL) Dictionary (incorporating Konjugator and Rhymer)* [Online], available: http://eurfa.org.uk [Accessed 07/07/20].

Donnelly, K. (2013b) *Kynulliad3: a corpus of 350,000 aligned Welsh and English sentences from the Third Assembly (2007-2011) of the National Assembly for Wales* [Online], available: http://cymraeg.org.uk/kynulliad3/ [Accessed 07/07/20].

Donnelly, K. and Deuchar, M. (2011) 'The Bangor Autoglosser: A multilingual tagger for conversational text', in Proceedings of *the Fourth International Conference on Internet Technologies and Applications (ITA11),* Wrexham, Wales. pp. 17-25.

Expert Advisory Group on Language Engineering Standards. (1996) *EAGLES guidelines* [Online], available: http://www.ilc.cnr.it/EAGLES/browse.html [Accessed 07/07/20].

Estellés-Arolas, E. and González-Ladrón-De-Guevara, F. (2012) 'Towards an integrated crowdsourcing definition', *Journal of Information Science* 38**:** 189-200.

Evas, J. and Williams, C. H. (1998) 'Community language regeneration: realising potential', in Proceedings of *the International Conference on Community Language Planning,* Cardiff, Welsh Language Board. pp. 1-13.

Hardie, A. (2012) 'CQPweb – combining power, flexibility and usability in a corpus analysis tool', *International Journal of Corpus Linguistics* 17**:** 380-409.

Hawtin, A. (2018) *The Written British National Corpus 2014: Design, compilation and analysis,* Unpublished PhD thesis: Lancaster University.

Johns, T. (1991) 'Should you be persuaded: Two samples of data-driven learning materials', *English Language Research Journal* 4: 1-16.

Karlsson, F. (1990) 'Constraint grammar as a framework for parsing running text', in Proceedings of *the 13th International Conference on Computational Linguistics (COLING),* Helsinki, Finland. pp. 168-173.

Karlsson, F., Voutilainen, A., Heikkilä, J. and Anttila, A. (1995) *Constraint grammar: A language-independent framework for parsing unrestricted text,* Berlin/New York: Mouton de Gruyter.

Knight, D., Adolphs, S. and Carter, R. (2013) 'Formality in digital discourse: a study of hedging in CANELC', in Romero-Trillo, J. (ed) *Yearbook of corpus linguistics and pragmatics,* Netherlands: Springer. pp. 131-152.

Leńko-Szymańska, A. and Boulton, A. (2015) *Multiple Affordances of Language Corpora in Data-driven Learning,* Amsterdam: John Benjamins.

Little, D. (2007) 'Language learner autonomy: Some fundamental considerations revisited', *Innovations in Language Learning and Teaching* 1: 14-29.

McEnery, T., Love, R., & Brezina, V. (2017) 'Compiling and analysing the Spoken British National Corpus 2014', *International Journal of Corpus Linguistics 22*(3): 311-318.

Mac Giolla Chríost, D., Carlin, P., Davies, S., Fitzpatrick, T., Jones, A. P., Heath-Davies, R., Marshall, J., Morris, S., Price, A., Vanderplank, R., Walter, C. and Wray, A. (2012). *Adnoddau, dulliau ac ymagweddau dysgu ac addysgu ym maes Cymraeg i Oedolion: astudiaeth ymchwil gynhwysfawr ac adolygiad beirniadol o'r ffordd ymlaen [Welsh for Adults teaching and learning approaches, methodologies and resources: a comprehensive research study and critical review of the way forward]*, Bedwas: Welsh Government.

Nation, I.S.P. (2001) *Learning Vocabulary in Another Language*, Cambridge: Cambridge University Press.

Neale, S., Donnelly, K., Watkins, G. and Knight, D. (2018) 'Leveraging lexical resources and constraint grammar for rule-based part-of-speech tagging in Welsh' in Proceedings of *the Eleventh International Conference on Language Resources and Evaluation (LREC),* Miyazaki, Japan. pp. 3946-3954.

NFER (2008) *Ymchwil i'r Cwrs Dwys ar gyfer Cymraeg i Oedolion*, Swansea: National Foundation for Educational Research.

ONS (2011) *DC2612WA – Ability to speak Welsh by occupation* [Online], Office for National Statistics, Durham: Nomis. Available: www.nomisweb.co.uk/census/2011/dc2612wa [Accessed 07/07/20].

Piao, S., Rayson, P., Knight, D. and Watkins, G. (2018) 'Towards a Welsh semantic annotation system' in Proceedings of *the Eleventh International Conference on Language Resources and Evaluation (LREC),* Miyazaki, Japan. pp. 980-985.

Rayson, P., Archer, D., Piao, S. and McEnery, T. (2004) 'The UCREL semantic analysis system', in Proceedings of *the Workshop on Beyond Named Entity Recognition Semantic Labelling for NLP tasks at the 4th International Conference on Language Resources and Evaluation (LREC),* Lisbon, Portugal. pp. 1-6.

Scannell, K. (2007) 'The Crúbadán Project: Corpus building for under-resourced languages' in *Building and Exploring Web Corpora: Proceedings of the 3rd Web as Corpus Workshop*. pp. 1-10.

Scannell, K. (2012) *Kevin Scannell's website* [Online], available: http://borel.slu.edu/ [Accessed 07/07/20].

Sinclair, J. (2005) 'Corpus and text - basic principles', in Wynne, M. (ed) *Developing Linguistic Corpora: a Guide to Good Practice,* Oxford: Oxbow Books. pp. 1-16.

Thomas, E. M. and Mayr, R. (2010) 'Children’s acquisition of Welsh in a bilingual setting: a psycholinguistic perspective, in Morris, D. (ed) *Welsh in the 21st Century,* Cardiff: Cardiff University Press.

Thompson, P. (2005) 'Spoken Language Corpora' in Wynne, M. (ed) *Developing Linguistic Corpora: a Guide to Good Practice,* Oxford: Oxbow Books. pp. 59-70.

Welsh Government (2013) *Codi golygon: adolygiad o Gymraeg i Oedolion. Adroddiad ac argymhellion [Raising our sights: review of Welsh for Adults. Report and recommendations]*, Bedwas: Welsh Government.

Welsh Government (2014) *A Living Language: a language for living - Moving forward*, Cardiff: Welsh Government

Welsh Government (2017) *Cymraeg 2050: A million Welsh speakers Action Plan 2019-20*, Cardiff: Welsh Government.

Wilson, A. (2002) *The Language Engineering Resources for the Indigenous Minority Languages of the British Isles and Ireland Project* [Online], available: https://www.lancaster.ac.uk/fass/projects/biml/default.htm [Accessed 07/07/20].

Corpws Cenedlaethol Cymraeg Cyfoes
National Corpus of Contemporary Welsh

# Y Corpws Cenedlaethol Cymraeg Cyfoes

Adroddiad y Prosiect, Hydref 2020

**Awduron:** Dawn Knight[5], Steve Morris[6], Tess Fitzpatrick[6], Paul Rayson[7], Irena Spasić[5] ac Enlli Môn Thomas[8].

## 1. Cyflwyniad

### 1.1. Diben yr adroddiad hwn

Bydd yr adroddiad hwn yn rhoi trosolwg o brosiect Corpws Cenedlaethol Cymraeg Cyfoes (CorCenCC) a'r adnodd corpws ar-lein a ddatblygwyd o ganlyniad i waith y prosiect. Bydd yr adroddiad yn amlinellu sylfeini damcaniaethol yr ymchwil, gan ddangos sut mae'r prosiect wedi adeiladu ar y theori hon a'i hymestyn. Byddwn hefyd yn codi ac yn trafod rhai o'r cwestiynau gweithredol allweddol a gododd wrth i'r prosiect fynd rhagddo, gan amlinellu'r ffyrdd y cawsant eu hateb, effaith y penderfyniadau hyn ar yr adnodd sydd wedi'i lunio, a'r cyfraniad tymor hwy y byddant yn ei wneud i arferion wrth adeiladu corpws. Yn olaf, byddwn yn trafod rhai o gymwysiadau'r gwaith a'r modd o'i ddefnyddio, gan amlinellu'r effaith y mae'r CorCenCC yn debygol o'i chael ar amrediad o unigolion a grwpiau defnyddwyr gwahanol.

### 1.2. Trwydded

Trwyddedir corpws CorCenCC a'r offerynnau meddalwedd cysylltiedig dan Creative Commons CC-BY-SA f4 ac felly maent yn rhydd ar gyfer eu defnyddio gan gymunedau proffesiynol ac unigolion sydd â diddordeb mewn iaith. Darperir cymwysiadau a chyfarwyddiadau pwrpasol ar gyfer pob offeryn (cyfeiriwch at adran 10 yr adroddiad hwn am ddolenni i'r holl offerynnau). Wrth adrodd gwybodaeth a ddeilliodd o ddefnyddio data a/neu offerynnau corpws CorCenCC, dylid cydnabod CorCenCC yn briodol (gweler 1.3.).

- I gael mynediad i'r corpws, ewch i: www.corcencc.cymru/archwilio/
- I gael mynediad i wefan GitHub: https://github.com/CorCenCC
    - Mae GitHub yn wasanaeth sydd yn y cwmwl sy'n galluogi datblygwyr i storio, rhannu a rheoli eu cod a'u setiau data.

### 1.3. Cyfeirio at CorCenCC

Rhaid rhoi cydnabyddiaeth briodol wrth ddefnyddio data a/neu offerynnau corpws CorCenCC. Defnyddiwch y canlynol wrth gyfeirio at gorpws CorCenCC:

---

[5] Prifysgol Caerdydd
[6] Prifysgol Abertawe
[7] Prifysgol Caerhirfryn
[8] Prifysgol Bangor

- **Corpws CorCenCC:** Knight, D., Morris, S., Fitzpatrick, T., Rayson, P., Spasić, I., Thomas, E-M., Lovell, A., Morris, J., Evas, J., Stonelake, M., Arman, L., Davies, J., Ezeani, I., Neale, S., Needs, J., Piao, S., Rees, M., Watkins, G., Williams, L., Muralidaran, V., Tovey-Walsh, B., Anthony, L., Cobb, T., Deuchar, M., Donnelly, K., McCarthy, M. a Scannell, K. (2020). *CorCenCC: Corpws Cenedlaethol Cymraeg Cyfoes* [Adnodd digidol]. Ar gael yn: http://www.corcencc.cymru/archwilio/

Gellir dod o hyd i gyhoeddiadau eraill yn adran 10 yr adroddiad hwn ac ar dab 'Allbynnau' gwefan CorCenCC: www.corcencc.cymru/allbynnau/

### 1.4. Cydnabyddiaethau

Cafodd yr ymchwil y seiliwyd yr adroddiad hwn arni, a'r adnodd corpws ar-lein sy'n cyd-fynd â hi, eu hariannu gan Gyngor Ymchwil Economaidd a Chymdeithasol y Deyrnas Unedig (ESRC) a'r Cyngor Ymchwil i'r Celfyddydau a'r Dyniaethau (AHRC) fel *y Corpws Cenedlaethol Cymraeg Cyfoes: Dull cymunedol o adeiladu corpws ieithyddol* (Rhif Grant ES/M011348/1). Gellir dod o hyd i wybodaeth am aelodau tîm y prosiect yn www.corcencc.cymru/cysylltiadau. Ni fyddai prosiect CorCenCC wedi bod yn bosibl heb eu mewnbwn, eu harbenigedd, eu brwdfrydedd a'u colegoldeb.

Rydym hefyd am gydnabod Prifysgol Caerdydd a Phrifysgol Abertawe am eu cyfraniad i ysgoloriaethau doethuriaeth, gan ein galluogi i gynnwys ymchwilwyr ôl-raddedig yn nhîm y prosiect. Rydym yn diolch yn arbennig ac yn bersonol i'r cydweithwyr hynny yn ein holl brifysgolion perthnasol sydd wedi rhoi'n hael o'u hamser a'u cymorth i ni yn ystod camau hanfodol bwysig y prosiect.

Rhaid rhoi'r diolch am gyflawni prosiect CorCenCC hefyd i'n rhanddeiliaid prosiect (gweler 2.2.), ac yn enwedig y bobl hynny o grŵp cynghori'r prosiect: sydd nid yn unig wedi bod yn hael wrth hwyluso'r gwaith o gasglu data, neu ei gyfrannu'n uniongyrchol, ond sydd hefyd wedi bod yn galonogol yn eu cefnogaeth o safbwynt nodau'r prosiect a'u hymgysylltiad â'r broses gynllunio, yn ogystal â'u hymroddiad i gynaliadwyedd a pharhad CorCenCC.

## 2. Gweledigaeth ac amcanion

### 2.1. Trosolwg o'r prosiect

Prosiect rhyngddisgyblaethol ac amlsefydliadol yw CorCenCC sydd wedi creu corpws ffynhonnell agored o Gymraeg cyfoes ar raddfa eang. Yn y cyd-destun hwn, corpws yw casgliad o enghreifftiau o iaith lafar, ysgrifenedig a/neu e-iaith o gyd-destunau bywyd go iawn sy'n galluogi defnyddwyr i nodi ac archwilio iaith fel y mae'n cael ei defnyddio mewn gwirionedd yn hytrach na dibynnu ar reddf neu gyfarwyddiadau rhagnodol o ran sut y 'dylai' gael ei defnyddio. Mae corpysau'n ein galluogi i ymchwilio i sut yr ydym yn defnyddio iaith ar draws sawl genre a chyfrwng cyfathrebu gwahanol (h.y. ar lafar, yn ysgrifenedig neu'n ddigidol), a sut mae'n amrywio yn ôl y siaradwr/awdur a diben y cyfathrebu. Defnyddir y dull sy'n seiliedig ar dystiolaeth hwn gan ymchwilwyr academaidd, geiriadurwyr, athrawon, dysgwyr iaith, aseswyr, datblygwyr adnoddau, llunwyr polisïau, cyhoeddwyr, cyfieithwyr ac eraill, ac mae'n hanfodol i waith datblygu technolegau fel llunio testun rhagfynegol, offerynnau prosesu geiriau, cyfieithu peirianyddol, adnabod llais ac offerynnau chwilio ar y we.

Cyn adeiladu CorCenCC, roedd nifer o gorpysau'r Gymraeg yn bodoli, gan gynnwys corpws Siarad (Deuchar ac eraill, 2018), a oedd yn cynnwys 460,000 o eiriau llafar, Corpws Cymraeg Crúbadán (Scannell, 2007), a oedd yn seiliedig ar e-iaith ac yn cynnwys 24 miliwn o eiriau, a Chorpws Llafar Paldaruo (Cooper ac eraill, 2019), a gafodd ei gyfrannu'n dorfol ac a oedd yn cynnwys darnau o destun a oedd wedi'u darllen yn uchel. Rhoddwyd ystyriaeth i p'un a ellid integreiddio'r rhain yn amcanion CorCenCC, neu eu halinio â nhw. Gweithiodd Deuchar a Scannell fel ymgynghorwyr ar gyfer y prosiect, a chafodd Canolfan Bedwyr (Bangor) ei chynrychioli yng ngrŵp cynghori'r prosiect. Fodd bynnag, o gofio y cafodd y corpysau blaenorol eu casglu er mwyn gwireddu nodau a gweledigaethau gwahanol, penodol a phwrpasol, ystyriwyd bod angen creu set ddata newydd a chyflawn.

CorCenCC yw corpws cyntaf y Gymraeg sy'n cwmpasu pob un o dair elfen iaith y Gymraeg cyfoes: llafar, ysgrifenedig ac iaith sydd wedi'i chyfryngu'n electronig (e-iaith). Mae'n cynnig ciplun o'r Gymraeg ar draws amrediad o gyd-destunau lle caiff ei defnyddio, e.e. sgyrsiau preifat, cymdeithasu fel grŵp, busnes a sefyllfaoedd gwaith eraill, ym maes addysg, yng nghyfryngau amrywiol y wasg, ac mewn mannau cyhoeddus. Mae'n cynnwys enghreifftiau o benawdau newyddion, gohebiaeth a negeseuon e-bost personol a phroffesiynol, gwaith ysgrifenedig academaidd, iaith lafar ffurfiol ac anffurfiol, postiadau ar flogiau a negeseuon testun (caiff cynnwys penodol y corpws ei drafod yn adran 3.3). Cafodd data am iaith ei samplu gan amrediad o siaradwyr a defnyddwyr Cymraeg, o bob rhanbarth o Gymru, o bob oedran a rhyw, o amrediad eang o swyddi ac o amrywiaeth o gefndiroedd ieithyddol (e.e. sut y daethant i siarad Cymraeg), i adlewyrchu amrywiaeth y mathau o destun a'r siaradwyr Cymraeg a geir yn y Gymru gyfoes. Yn y modd hwn, mae corpws CorCenCC yn darparu ffordd o rymuso defnyddwyr Cymraeg i ddeall yr iaith yn well, ac arsylwi arni, ar draws lleoliadau amrywiol, ac mae'n creu sylfaen dystiolaeth gadarn ar gyfer addysgu Cymraeg cyfoes i'r bobl hynny sydd am ei defnyddio. Dros amser, gallai'r corpws gyfrannau'n sylweddol at drawsnewid y Gymraeg fel yr iaith drafod ym meysydd y cyhoedd, masnach, addysg a llywodraeth.

I'r perwyl hwnnw, nod CorCenCC yw galluogi defnyddwyr cymunedol, er enghraifft, i ymchwilio i amrywiadau neu fympwyon tafodieithol yn eu hiaith eu hunain; defnyddwyr proffesiynol i broffilio testunau i weld pa mor ddarllenadwy ydynt neu ddatblygu offerynnau iaith digidol; dysgwyr yr iaith i ddysgu o fodelau go iawn o Gymraeg; ac ymchwilwyr i ymchwilio i batrymau o ran sut caiff iaith ei defnyddio a'i newid. Rhagwelir y bydd y corpws hefyd yn datgelu mewnwelediadau newydd i batrymau geirfa ac iaith y Gymraeg ac y bydd yn adnodd pwysig ar gyfer addysgu'r Gymraeg i siaradwyr y mae'n iaith gyntaf ganddynt a siaradwyr Cymraeg newydd. Mae'r effaith amlochrog bosibl hon wedi bod yn bosibl oherwydd cyfraniad sylweddol CorCenCC ar y lefel fethodolegol, wrth ymestyn cwmpas, perthnasedd a seilwaith dylunio corpysau ieithyddol. Yn benodol, mae'r prosiect wedi golygu datblygu offerynnau a phrosesau newydd pwysig, gan gynnwys gwaith cynllunio corpws unigryw a yrrir gan ddefnyddwyr lle cafodd data am iaith ei gasglu a'i ddilysu'n dorfol, a phecyn cymorth addysgeg (Y Tiwtiadur), yr oedd wedi'i ymgorffori, a ddatblygwyd drwy ymgynghori â chynrychiolwyr o bob grŵp o ddefnyddwyr academaidd a chymunedol a ragwelwyd (gweler Knight ac eraill, 2020a – ceir manylion yn adran 10.2. isod – am drafodaeth fanwl o gynllun CorCenCC a yrrir gan ddefnyddwyr).

## 2.2. Tîm y prosiect

Roedd y canlynol ynghlwm wrth brosiect CorCenCC: pedwar sefydliad academaidd (sef Prifysgol Caerdydd, Prifysgol Abertawe, Prifysgol Caerhirfryn a Phrifysgol Bangor), un prif ymchwilydd (Dawn Knight), dau gyd-ymchwilydd (Tess Fitzpatrick a Steve Morris) a ffurfiodd, ynghyd â'r prif ymchwilydd, Dîm Rheoli CorCenCC, a chyfanswm o saith cyd-ymchwilydd eraill (Irena Spasić, Paul Rayson, Enlli Môn Thomas, Alex Lovell, Jonathan Morris, Jeremy Evas a Mark Stonelake), deg cynorthwyydd/cysylltai ymchwil, a 180+ o drawsgrifwyr a fu'n gweithio yn ystod y prosiect.

Yn ogystal, roedd chwe ymgynghorydd, dau fyfyriwr doethuriaeth, pedwar myfyriwr israddedig ar leoliad haf, pedwar aelod o staff cymorth gwasanaethau proffesiynol, a dau wirfoddolwr prosiect. Bu'r prosiect ar ei ennill hefyd gan gyfraniadau a chymorth gan gynrychiolwyr o amrediad o randdeiliaid, gan gynnwys Llywodraeth Cymru, Cynulliad Cenedlaethol Cymru, y BBC, S4C, CBAC, Cymraeg i Oedolion, Gwasg y Lolfa, SaySomethinginWelsh a Geiriadur Prifysgol Cymru, drwy Grŵp Cynghori Prosiect. Nia Parry (cyflwynydd, cynhyrchydd ac ymchwilydd teledu; tiwtor Cymraeg, *Welsh in a week* (S4C)), Nigel Owens (dyfarnwr rygbi rhyngwladol; cyflwynydd teledu), Cerys Matthews (cerddor; awdur; cyflwynydd radio a theledu), a Damien Walford Davies (bardd; Athro Llenyddiaeth Cymraeg a Saesneg; cyn-Gadeirydd Llenyddiaeth Cymru) yw llysgenhadon swyddogol prosiect CorCenCC. Gellir gweld rhestr lawn o'r holl unigolion a fu ynghlwm wrth y prosiect yn www.corcencc.cymru/cysylltiadau/ – a chyfeirir at lawer ohonynt drwy gydol yr adroddiad hwn fel y bo'n berthnasol.

Cafodd y prosiect ei hwyluso gan dîm traws-sefydliadol cadarn, a gefnogodd waith recriwtio staff, rheoli ariannol, technoleg gwybodaeth (gan gynnwys cyfarpar, meddalwedd, cynnal a chadw'r gweinyddion a gwefannau), allgymorth y cyfryngau a chyfathrebu (gan gynnwys cydlynu datganiadau i'r wasg, a siarad ar y radio a'r teledu), arweiniad cyfreithiol o safbwynt ffurflenni, contractau a thrwyddedau, a chyfieithwyr a chyfieithwyr ar y pryd Cymraeg (a ddarparodd gyfieithiadau ysgrifenedig ar gyfer adroddiadau, dogfennau prosiect allweddol ac allbynnau eraill, a chyfieithu ar y pryd yn ystod holl gyfarfodydd y Tîm Prosiect Cyfan a digwyddiadau rhannu gwybodaeth â'r cyhoedd). Cafodd dros 210 o adroddiadau eu hysgrifennu, ddwywaith yr wythnos, yn ystod y prosiect, a oedd yn nodi manylion y gwaith a gyflawnwyd, problemau a risgiau allweddol, y gwaith i'w wneud, syniadau newydd, meddyliau a chyfleoedd; cynhaliwyd deg cyfarfod Tîm Prosiect Cyfan, a digwyddodd ymhell dros 100 o gyfarfodydd ychwanegol. Cafodd saith rhestr bostio fewnol eu llunio er mwyn hwyluso'r cyfathrebu rhwng aelodau'r tîm a oedd wedi'u lleoli ar safleoedd gwahanol, ynghyd â siart Gantt ganolig ar gyfer y prosiect cyfan, er mwyn cofnodi ac olrhain y pethau yr oedd angen eu cyflawni, a cherrig milltir prosiect allweddol, ar y cyd. Er mwyn cynnal y cyfathrebu â'r cyhoedd a rhanddeiliaid eraill, cafodd 24 rhifyn o newyddlen brosiect eu cyhoeddi a'u dosbarthu i unigolion, a'u lanlwytho i'r brif wefan. Roedd y newyddlenni'n darparu darllenwyr â'r wybodaeth ddiweddaraf am waith casglu data, yn adrodd ar gyflwyniadau a phrif gyflwyniadau a roddwyd (sef cyfanswm o 54 ohonynt), ac yn eu cyflwyno i aelodau unigol y tîm drwy ein pwt 'dyma'r tîm' rheolaidd. Diben hwn oedd cynnal diddordeb ac ymdeimlad o fuddsoddi yn y prosiect (yn unol â'r cynllun a yrrir gan ddefnyddwyr). Gwnaeth gwefannau'r prosiect (sef www.corcencc.cymru | www.corcencc.org), a ffrydiau Facebook a Twitter hefyd hwyluso â'r broses o ennyn diddordeb y cyhoedd, gan arwain at dros 140,000 o ymweliadau â'r gwefannau a 1,029 o ddilynwyr ar Twitter a 374 ar Facebook (hyd at fis Awst 2020). Er y cânt eu crybwyll yn olaf yma, aelodau pwysicaf y tîm CorCenCC estynedig yw'r 2,000+ o unigolion a gyfrannodd at y corpws.

### 2.3. Pecynnau gwaith (WP)

Cafodd y gwaith ar brosiect CorCenCC ei ddosbarthu ar draws chwe phecyn gwaith (WP) cydgysylltiedig, yr oedd gan bob un ohonynt dasgau, nodau ac amcanion penodol. Dan arweiniad Knight, roedd WP0 yn cynnwys gweithgareddau dylunio, rhychwantu a hyfforddi parhaus, ac yn cynnwys holl aelodau tîm y prosiect. Roedd cynnwys y pecynnau gwaith eraill fel a ganlyn:

- WP1: Casglu, trawsgrifio ac anonymeiddio'r data
- WP2: Datblygu'r set dagiau / tagiwr rhannau ymadrodd
- WP3: Datblygu tagiwr semantig ar gyfer y Gymraeg a thagio'r holl ddata yn semantig
- WP4: Cwmpasu, dylunio ac adeiladu Y Tiwtiadur
- WP5: Adeiladu'r seilwaith i letya CorCenCC a chreu'r corpws

Er y cafodd y gwaith ei ddosbarthu ar draws y pecynnau gwaith hyn, roedd cydweithwyr yn meddu ar gyd-ddealltwriaeth o'r weledigaeth a rennir ar gyfer y prosiect, gan weithio ar y cyd er mwyn ei chyflawni, ac roedd lefel sylweddol o gyd-ddibyniaeth rhwng y pecynnau gwaith yr oedd yn gofyn am waith trafod a chydlynu. Er enghraifft, roedd WP3 yn adeiladu ar y gwaith ymchwil a wnaethpwyd yn WP1 o safbwynt casglu data ar gyfer y corpws, a defnyddiodd dagiwr rhannau ymadrodd WP2 fel y cam cyntaf ym mhroses dadansoddi'r data am y Gymraeg yn semantig. Yna, llywiodd allbwn WP3 waith WP4 ar gyfer y pecyn cymorth pedagogaidd ar-lein (Y Tiwtiadur), a ddefnyddiodd lefelau lluosog o anodiadau'r corpws i wella ymgysylltiad athrawon a dysgwyr â'r pecyn cymorth, a'r hyn y gellid ei ddefnyddio ar ei gyfer. Yn ogystal, llywiodd y corpws yr oedd wedi'i dagio'n semantig yn WP3 seilwaith y corpws a ddatblygwyd yn WP5. Yn yr adrannau canlynol, byddwn yn rhoi disgrifiad manwl o'r pecynnau gwaith, yn amlinellu eu prif nodau ac amcanion, ac yn myfyrio ar eu prif gyflawniadau, eu prif gyfraniadau a ffyrdd posibl o'u defnyddio. Ysgrifennwyd y disgrifiadau hyn gan arweinwyr perthnasol y pecynnau gwaith.

## 3. Pecyn Gwaith 1: Casglu, trawsgrifio ac anonymeiddio'r data

### 3.1. WP1: Disgrifiad

Prif waith WP1 oedd cyrchu, casglu a phrosesu'r data a oedd i'w gynnwys yn CorCenCC. Elfennau craidd y weithdrefn hon oedd i) creu fframwaith samplu'r prosiect; ii) sefydlu confensiynau trawsgrifio; iii) sicrhau dull cyson o safbwynt cydymffurfiaeth foesegol o ran casglu'r data. Cafodd WP1 ei arwain ar y cyd gan Morris a Knight, ac ymunodd tîm o ymchwilwyr a oedd yn siarad Cymraeg â nhw, gan gynnwys y cyd-ymchwilwyr Evas, J.

Morris, a Lovell, a'r cynorthwywyr/cysyllteion ymchwil Needs, Rees, Arman, Watkins a Williams (ar bwyntiau gwahanol yn ystod y prosiect). Darparodd Deuchar a McCarthy, sy'n arwain ym maes ieithyddiaeth corpws, gyngor ymgynghorol parhaus ar gyfer y cam hwn drwy gydol y prosiect a chafodd cymorth ychwanegol ei ddarparu gan sawl gwirfoddolwr ac intern prosiect.

## 3.2. WP1: Amcanion

Roedd nodau ac amcanion WP1 fel a ganlyn:

- dylunio fframwaith samplu ar gyfer y corpws
- cyrchu a chasglu data priodol
- cynllunio a chymhwyso protocolau trawsgrifio i ddata llafar

*Y fframwaith samplu*

Roedd yr Achos am Gefnogaeth yng nghynnig y prosiect yn cynnwys canllaw amlinellol i'n hamcanion mewn perthynas ag elfennau (llafar, ysgrifenedig ac e-iaith), genres a thestunau ieithyddol, ac amcangyfrif o faint o eiriau y byddai'n cael eu casglu o dan bob pennawd. Un o dasgau cyntaf WP1 oedd mireinio a datblygu'r canllaw ar-lein hwn ymhellach. Cafodd fframwaith samplu ei llunio er mwyn tanategu'r gwaith casglu data ar gyfer y prosiect, er mwyn sicrhau y byddem yn casglu amrediad o siaradwyr gwahanol ar draws cyd-destunau trafod a lleoliadau daearyddol gwahanol. Cynlluniwyd y fframwaith samplu i adlewyrchu demograffeg bresennol siaradwyr Cymraeg gan ddefnyddio'r wybodaeth cyfrifiad ddiweddaraf (ONS, 2011). Un o'r agweddau arloesol ar fframwaith samplu CorCenCC yw'r ystyriaeth fanwl y mae'n ei rhoi i'r meysydd lle defnyddir y Gymraeg. Mewn cyd-destun lle bo'r mwyafrif helaeth o siaradwyr Cymraeg yn ddwyieithog a bod gwasgariad daearyddol anghyson yn nhermau dwysedd siaradwyr, oedran siaradwyr a meysydd ieithyddol, roedd angen i'r fframwaith samplu adlewyrchu sefyllfa sosioieithyddol gyfredol yr iaith yn y modd mwyaf cywir posibl.

*Cyrchu'r data*

Cafodd y targedau ar gyfer y data llafar, ysgrifenedig ac e-iaith a oedd i'w gasglu ar gyfer CorCenCC, a'i ffynonellau, eu gyrru gan ein fframwaith samplu a chawsant eu llunio gan ymchwiliad cychwynnol i ble y siaredir y Gymraeg a lle ceir y defnydd mwyaf ohoni yn nhermau deunydd ysgrifenedig ac e-iaith. Mewn cyd-destun dwyieithog, gallai meysydd penodol fod wedi'u tangynrychioli (e.e. papurau newydd dyddiol cenedlaethol). Felly, roedd angen sicrhau bod y data'n adlewyrchiad gwirioneddol o'r hyn sydd ar gael i ddefnyddwyr yr iaith ac y mae modd ei gyrchu ganddynt, yn hytrach nag atgynhyrchu fframweithiau sydd â'r nod o greu corpysau mewn ieithoedd lle bo'r mwyafrif o siaradwyr yn unieithog.

*Trawsgrifio*

Roedd dau gam paratoadol: (i) creu confensiynau trawsgrifio ar gyfer y Gymraeg a (ii) recriwtio trawsgrifwyr (gweler 3.3. isod hefyd). Roedd heriau penodol ynghlwm wrth cam (i) am sawl rheswm:

- Yn yr iaith ysgrifenedig, mae awduron yn aml yn dynodi amrywiaeth ym mathau gwahanol o iaith lafar, fel bod yr un ystyr ieithyddol yn cael ei ysgrifennu mewn llawer

o ffyrdd gwahanol. Enghraifft o hyn byddai amser presennol y person cyntaf unigol ar gyfer 'bod'. Yn Saesneg, gellid cyfleu hyn ar ffurf '*I am*' neu '*I'm*'. Byddai Cymraeg ffurfiol ysgrifenedig yn rhoi '*Yr wyf (i)*' neu '*Rwyf i*'. Fodd bynnag, ceir y posibiliadau canlynol ar gyfer Cymraeg llafar: '*Rydw i / Dw i / Rwy / Wy / Fi*'. Byddai awduron yn ysgrifennu'r ffurfiau hyn i gynrychioli siaradwyr o ardaloedd gwahanol a gellir arsylwi arnynt mewn llenyddiaeth a chyfryngau ysgrifenedig eraill. Roedd angen i'r confensiynau trawsgrifio felly allu adlewyrchu'r realaeth hon wrth ei mynegeio at yr un ystyr, fel y gallai chwiliadau ddod o hyd i'r sylweddiadau gwahanol.

- O gofio bod CorCenCC yn cynnwys iaith electronig yn ogystal â data ysgrifenedig a llafar, a'r ffaith nad yw'r confensiynau ar gyfer cynrychioli iaith yn y cyd-destun electronig wedi'u sefydlu'n llawn mewn unrhyw iaith, roedd angen darparu ar gyfer yr achosion hyn mewn modd tebyg.
- Mewn perthynas â thrawsgrifio deunydd llafar, yr egwyddor gyffredinol a fabwysiadwyd oedd alinio'r hyn y cafodd ei glywed â'r sylweddiad ysgrifenedig agosaf o set a oedd yn dechrau â'r amrediad presennol o ffurfiau ysgrifenedig ond yr ategwyd ati yn ôl yr angen er mwyn sicrhau bod pob ffurf lafar wedi'i chipio'n briodol. Cafodd yr egwyddor hon ei mireinio a'i datblygu drwy sawl iteriad o gonfensiynau trawsgrifio CorCenCC.
- Ar ôl mabwysiadu'r egwyddor drawsgrifio gyffredinol hon, roedd angen sicrhau cysondeb ar draws tîm trawsgrifio CorCenCC. Roedd yn rhaid ymwrthod yn gadarn ag unrhyw duedd tuag at ragysgrifiadaeth neu nodi ffurf ysgrifenedig ffurfiol y Gymraeg.

Cafodd trawsgrifwyr eu recriwtio drwy ymgyrchoedd a oedd wedi'u targedu'n benodol at aelodau o'r proffesiwn cyfieithu (drwy newyddlen eu cymdeithas), myfyrwyr prifysgol, a'r bobl hynny a oedd wedi bod ynghlwm wrth y gwaith o drawsgrifio ar gyfer prosiectau eraill. Roedd yn rhaid i bob trawsgrifydd lwyddo mewn darn prawf rhagarweiniol (gan lynu wrth gonfensiynau trawsgrifio CorCenCC), a chafodd ansawdd ei sicrhau drwy wirio, ar hap, 25 y cant o'r holl waith a drawsgrifiwyd, gan wneud cywiriadau lle nodwyd problemau.

## 3.3. WP1: Cyflawniadau

*Y fframwaith samplu*

Mae Tabl 1 yn cynrychioli'r fframwaith samplu cychwynnol a gynigiwyd yn yr Achos am Gefnogaeth yng nghynnig y prosiect. Roedd y dosbarthiadau cychwynnol hyn yn seiliedig ar y dymuniad o ganolbwyntio ar gynrychioli'r iaith a siaredir yn ddifyfyr (ar lafar ac e-iaith) o'i chymharu ag iaith sydd wedi'i pharatoi (ysgrifenedig). Ar ôl cychwyn y prosiect, cafodd fframiau samplu mwy manwl eu datblygu ar gyfer cydrannau llafar (Tabl 2), ysgrifenedig (Tabl 3) ac e-iaith (Tabl 4) y corpws. Cafodd proses mireinio iteraidd y fframwaith samplu ei llywio gan grwpiau thematig a chategoreiddio trafod corpysau pwysig a oedd eisoes yn bodoli, gan gynnwys Corpws Cenedlaethol Prydain (BNC) 1994, Corpws Llafar Cenedlaethol Prydain 2014, CANCODE a CANELC (gweler Aston a Burnard, 1997, McEnery ac eraill, 2017, Carter and McCarthy, 2004 a Knight et al., 2013). Darparodd fframweithiau'r corpysau hyn a oedd eisoes yn bodoli fan cychwyn defnyddiol ar gyfer archwilio i ba raddau y mae'n bosib defnyddio'r un grwpiau a chategorïau yng nghyd-destun ieithoedd sydd wedi'u lleiafrifo. Ceir gwybodaeth sydd â rhagor o fanylder am bob un o'r is-genres, a'r cyfiawnhad dros y dosbarthiadau arfaethedig hyn, yn Knight ac eraill, 2021a (gweler adran 10.2.).

***Tabl 1.*** Fframwaith samplu cychwynnol ar gyfer CorCenCC (sydd wedi'i dynnu o'r Achos am Gefnogaeth).

| **Math** | **Ffynonellau enghreifftiol (amcangyfrif)** | **Geiriau** | **Cyfanswm** |
|---|---|---|---|
| Llafar | Iaith drafod dysgwyr Cymraeg | 600,000 | 4 miliwn |
| | Sgyrsiau â ffrindiau; â'r teulu; cyfweliadau a ddarlledwyd ar y teledu a sioeau siarad teledu (BBC); Cymraeg yn y gweithle | 400,000 yr un | |
| | Sioeau radio'r BBC; cyfarfyddiadau gwasanaethau | 400,000 yr un | |
| | Galwadau ffôn; rhyngweithiadau mewn dosbarthiadau ysgolion cynradd ac uwchradd a cholegau trydyddol ac mewn dosbarthiadau i oedolion; areithiau gwleidyddol; rhyngweithiadau ffurfiol ac anffurfiol yn yr Eisteddfod Genedlaethol | 250,000 yr un | |
| Ysgrifenedig | Gwaith ysgrifenedig gan ddysgwyr Cymraeg | 600,000 | 4 miliwn |
| | Llyfrau; papurau bro; dogfennau gwleidyddol; storïau | 400,000 yr un | |
| | Llythyrau a dyddiaduron; traethodau academaidd; gwerslyfrau academaidd; cylchgronau; hysbysebion, taflenni gwybodaeth/cyhoeddusrwydd; llythyrau ffurfiol | 290,000 yr un | |
| | Arwyddion | 60,000 | |
| E-iaith | Byrddau trafod; negeseuon e-bost; blogiau | 500,000 yr un | 2 filiwn |
| | Gwefannau; trydariadau | 300,000 | |
| | Negeseuon testun | 200,000 | |
| | | **10,000,000** | |

***Tabl 2.*** Fframwaith samplu diwygiedig bras ar gyfer elfen lafar CorCenCC.

| **Cyd-destunau** | **% yr is-gorpws** | **Nifer y geiriau** |
|---|---|---|
| Cyhoeddus/sefydliadol | 10% | 400,000 |
| Cyfryngau | 15% | 600,000 |
| Trafodol | 10% | 400,000 |
| Proffesiynol | 10% | 400,000 |
| Addysgegol | 10% | 400,000 |
| Cymdeithasu | 22.5% | 900,000 |
| Preifat | 22.5% | 900,000 |
| | **100%** | **4,000,000** |

***Tabl 3.*** Fframwaith samplu diwygiedig bras ar gyfer elfen ysgrifenedig CorCenCC.

| **Ffynonellau** | **% yr is-gorpws** | **Nifer y geiriau** |
|---|---|---|
| Llyfrau | 41.75% | 1,670,000 |
| Cylchgronau, papurau newydd, cyfnodolion | 19.25% | 770,000 |
| Deunydd amrywiol | 39% | 1,560,000 |
| | 100% | 4,000,000 |

*Tabl 4.* Fframwaith samplu diwygiedig bras ar gyfer elfen e-iaith CorCenCC.

| **Ffynonellau** | **% yr is-gorpws** | **Nifer y geiriau** |
|---|---|---|
| Blog | 30% | 600,000 |
| Gwefan | 30% | 600,000 |
| Negeseuon e-bost | 20% | 400,000 |
| Negeseuon testun electronig byr | 20% | 400,000 |
| | **100%** | **2,000,000** |

Er bod y fframwaith samplu'n gweithredu fel offeryn amcangyfrif ar gyfer casglu data – 'delfryd' fel petai – prin iawn y mae corpws gorffenedig yn efelychu cyfansoddiad y fframwaith samplu (gweler Hawtin, 2018 am drafodaethau pellach ar hyn). Cafodd amrywiaeth o ffactorau ddylanwad ar gyfansoddiad terfynol y corpws, gan gynnwys pa mor hygyrch yr oedd unigolion penodol a/neu fathau o ddata, caniatadau, a materion mwy ymarferol yr oedd yn ymwneud â faint o amser mae'n ei gymryd i brosesu mathau penodol o ddata, a'r graddau y gellir rhagweld hyn. Unwaith roedd y ffactorau hyn i gyd wedi chwarae eu rhan, roedd cyfansoddiad y corpws fel a ganlyn (Tabl 5):

*Tabl 5.* Cyfansoddiad terfynol CorCenCC.

| **LLAFAR** | | | |
|---|---|---|---|
| **llafar_cyd-destun** | **Nifer y testunau** | **Nifer y geiriau** | **Cyfanswm** |
| darllediad | 564 | 750,078 | 1,332 testun<br><br>2,865,095 geiriau |
| pedagogaidd | 136 | 296,709 | |
| preifat | 93 | 245,719 | |
| proffesiynol | 80 | 477,983 | |
| cyhoeddus neu sefydliadol | 137 | 433,361 | |
| cymdeithasol | 131 | 456,487 | |
| trafodol | 191 | 204,758 | |
| **YSGRIFENEDIG** | | | |
| **ysgrifenedig_genre** | **Nifer y testunau** | **Nifer y geiriau** | **Cyfanswm** |
| cyfnodolyn | 10 | 304,447 | 707 testun<br><br>3,940,082 geiriau |
| llyfr | 137 | 1,928,582 | |
| traethodau_gwaith cwrs_ac_arholiadau | 31 | 26,047 | |
| taflen_dogfen_cyhoeddiad | 341 | 806,030 | |
| llythyr | 53 | 12,873 | |
| cylchgrawn | 80 | 329,203 | |
| amrywiol | 5 | 8,251 | |
| cylchlythyr | 33 | 78,803 | |
| papur_bro | 13 | 117,334 | |
| traethawd hir | 4 | 328,512 | |
| **IAITH ELECTRONIG** | | | |
| **eiaith_genre** | **Nifer y testunau** | **Nifer y geiriau** | **Cyfanswm** |
| blog | 48 | 2,345,909 | 9,397 testun<br><br>4,402,003 geiriau |
| e-bost | 781 | 141,554 | |
| neges destun | 8,487 | 93,541 | |
| gwefan | 81 | 1,820,999 | |
| | **11,436** | **11,207,180** | |

Mae'r corpws yn cynnwys dros 11,000,000 o eiriau, ond mae'r cyfansoddiad wedi newid fel bod yr elfen lafar yn cynnwys ychydig dros 2,800,000 o eiriau. Er bod yr is-gorpws hwn ychydig yn llai na'r hyn a gynlluniwyd yn wreiddiol, *ynddo'i hun*, dyma'r corpws mwyaf o safbwynt y Gymraeg llafar a siaredir yn naturiol. Am ddadansoddiad manwl o'r corpws (gan gynnwys cyd-destunau, genres a phynciau penodol ynghyd â chategorïau metadata demograffig a'u diffiniadau) gweler Knight ac eraill 2021a (adran 10.2.).

Dylid nodi fod yr offer ymholi corpws ar-lein yn rhoi cyfanswm o 14,338,149 o **docynnau** yn y corpws ac yn gwneud cyfrifiadau ar sail y gwerth hwnnw. Tocynnau yw'r uned leiaf a gynhwysir mewn corpws, sy'n cynnwys geiriau (h.y. eitemau sy'n dechrau gyda llythyren o'r wyddor) a ffugeiriau (h.y. eitemau sy'n dechrau gyda nod nad yw'n llythyren o'r wyddor). Felly, mae corpora bob amser yn cynnwys mwy o docynnau na geiriau. Seilir y gwerthoedd a drafodir yn y bennod hon ar eiriau yn unig gan fod modd dadlau bod hyn yn rhoi cyfrif mwy cywir o'r **unedau ystyr** a gynhwysir yn y corpws.

*Cyrchu data*

Wrth recriwtio cyfranwyr data llafar, y nod oedd sicrhau y byddai pob ardal yng Nghymru'n cael ei chynrychioli. Cyrchwyd data llafar drwy dau brif ddull: (i) recriwtio'r cyfranogwyr y byddai'n cael eu recordio a (ii) recriwtio'r cyfranogwyr a fyddai'n cyfrannu data llafar drwy ap CorCenCC (gweler 3.3. hefyd). Roedd cwmpas (i) nid yn unig yn golygu cynorthwywyr ymchwil a recordiodd siaradwyr yn y maes, ond hefyd cyfranogwyr yn recordio'u hunain yn ystod rhyngweithiadau amrywiol. Cafodd hyn ei hwyluso drwy rwydwaith o 'ymgyrchwyr' lleol (*animateuriaid* ieithyddol gweithredol ym meysydd a dargedwyd) neu'r Mentrau Iaith (mae Menter Iaith yn gysylltiedig â phob awdurdod lleol yng Nghymru, h.y. sefydliad sy'n seiliedig yn y gymuned sy'n ymrwymedig i godi proffil y Gymraeg drwy fentrau iaith lleol). Cyflawnwyd y broses recriwtio ar gyfer (ii) drwy roi cyhoeddusrwydd i'r ap (er enghraifft, drwy'r cyfryngau cymdeithasol, siarad ar y teledu a deunydd cyhoeddusrwydd) er mwyn gwneud popeth posibl i gyrraedd carfan wahanol o gyfranogwyr a fyddai'n recordio ar eu pen eu hunain ac mewn meysydd mwy preifat. Rhoddodd digwyddiadau mawr yng Nghymru, fel yr Eisteddfod Genedlaethol a Thafwyl, gyfleoedd i'r tîm gyrraedd trawstoriad o gyfranogwyr yn ogystal â chodi ymwybyddiaeth o'r prosiect yn gyffredinol.

Roedd yn her recriwtio pobl i gyfrannu gan ddefnyddio ap ffôn CorCenCC. Roedd yr ap ar gael ar iOS ac Android a thrwy ryngwyneb ar y we (i ddarparu ar gyfer y bobl hynny nad oedd ffôn symudol yn hygyrch iddynt), a chynhyrchodd ymgyrchoedd yn y cyfryngau lawer o frwdfrydedd chychwynnol, e.e. ei drafod ar raglenni teledu fel *Prynhawn Da* ar S4C, ac ar y radio drwy gyfrwng y Gymraeg a'r Saesneg, a thrwy ddigwyddiadau ymgysylltu lleol. Fodd bynnag, ni wnaeth hyn olygu bod llawer o bobl wedi defnyddio'r ap. Roedd adborth gan bartneriaid yn y Mentrau Iaith yn awgrymu y gallai pobl fod yn orbryderus y byddai modd eu hadnabod (er gwaethaf ymdrechion i roi sicrwydd y byddent yn cael eu hanonymeiddio), oherwydd bod y gymuned ieithyddol yn gymharol fach.

Roedd deunydd hyrwyddo (yr oedd wedi'i anelu at annog cyfranogiad, ond a oedd hefyd yn ffordd effeithiol o godi ymwybyddiaeth o'r prosiect) yn cynnwys ysgrifbinnau, matiau diod, taflenni a thaflenni gwybodaeth o faint cerdyn post. Cafodd masgot 'answyddogol' – yn seiliedig ar gath o'r enw Cor-pws – ei ddylunio er mwyn hwyluso cyfranogiad gan bobl o dan 18 oed, a buodd yn boblogaidd gan gyfranwyr o bob oedran.

Sefydlwyd cyfrifon Facebook a Twitter ar gyfer CorCenCC yn ystod ychydig fisoedd cyntaf y prosiect er mwyn gwella'r broses o recriwtio cyfranogwyr a'r cyfraniad ganddynt.

O safbwynt data ysgrifenedig, arweiniodd y cydberthynas da a sefydlwyd â chyhoeddwyr Cymraeg fel gwasg y Lolfa ar ddechrau'r prosiect at ymgorffori llawer o nofelau a llyfrau cyfoes yn y corpws. Mae'r papurau bro lleol yn ffynhonnell unigryw o ddata ysgrifenedig yn y Gymraeg (h.y. papurau newydd Cymraeg ar gyfer y gymuned leol). Penderfynwyd gweithio â'n cysylltiadau Menter Iaith lleol er mwyn casglu'r rhain. O ganlyniad i'n hymgysylltiad â rhanddeiliaid eraill y prosiect yn ystod cam cynllunio'r prosiect, cafodd data ei gasglu'n eithaf cyflym, er enghraifft gwaith samplu o gyfnodolyn academaidd Cymraeg *Gwerddon* drwy'r Coleg Cymraeg Cenedlaethol, ac adnoddau addysgegol / papurau arholiad ail iaith i oedolion drwy Gyd-bwyllgor Addysg Cymru.

O safbwynt data e-iaith, nid oeddem yn gallu casglu data o gyfrifon Twitter na Facebook oherwydd problemau'n ymwneud â chyfyngiadau perchenogaeth, ond cydweithiodd perchenogion gwefannau ac awduron blogiau'n hael, a churwyd y targedau gan gryn tipyn. Bu'n anoddach cipio negeseuon testun electronig byr (mewn modd tebyg i gasglu data drwy'r ap, ac am yr un rhesymau), ond bu'n haws casglu cyfraniadau drwy rif WhatsApp neilltuedig. Yn yr un modd, roedd casglu negeseuon e-bost personol yn her, ond roedd hi'n haws casglu e-ohebiaeth o weithleoedd. Roedd contractau â BBC Cymru/Wales ac S4C wedi golygu y gallai samplau o ddeunydd teledu a radio cyfoes, gan gynnwys podlediadau, gael eu cynnwys. Mae'n werth nodi bod y cydberthnasau gwaith cadarn a sefydlwyd â'r sefydliadau hyn wedi golygu eu bod hefyd wedi cyflenwi negeseuon e-bost o'r gweithle, newyddlenni ac ati ar gyfer y deunydd ysgrifenedig.

Cafwyd cymeradwyaeth foesegol ar gyfer pob agwedd ar gasglu data gan bob un o'r pedair prifysgol a fu ynghlwm wrth y prosiect. Roedd y ffurflenni caniatâd a gafodd eu llofnodi gan gyfranogwyr yn cynnwys eu cydsyniad ar gyfer casglu metadata pwysig (e.e. oedran, rhyw, lleoliad daearyddol) a oedd yn angenrheidiol ar gyfer y corpws. Gweler Knight ac eraill, 2021a, am drafodaeth fanwl ar yr ystyriaethau moesegol / heriau a wynebwyd wrth adeiladu CorCenCC.

*Trawsgrifio*

Er bod confensiynau trawsgrifio wedi cael eu creu ar gyfer y Gymraeg ar gyfer prosiectau eraill (e.e. Deuchar et al., 2014), penderfynwyd creu set o gonfensiynau bwrpasol ar gyfer trawsgrifio data CorCenCC. Gwnaeth y rhain ein galluogi i adlewyrchu'n llawn y sbectrwm cyfan o amrywiaeth mewn tafodieithoedd/cyweiriau a gipiwyd yn ein data llafar (gan ei wneud yn fwy defnyddiol i ymchwilwyr academaidd) yn ogystal a chynrychioli iaith lafar y cyfranogwyr eu hunain mewn modd mwy cywir. Yn benodol, oni chynrychiolwyd y gwahaniaethau'n gywir, ni fyddai'n bosibl cipio amrywiaethau yn y Gymraeg, na'r pellter a geir rhwng llawer o amrywiadau llafar a'r Gymraeg ysgrifenedig safonol. Rhoddwyd cyfarwyddyd i drawsgrifwyr <u>beidio</u> â chywiro patrymau llafar y gellid eu hystyried yn ansafonol neu a oedd yn cynnwys newid cod (h.y. newid rhwng ieithoedd gwahanol yn ystod achos o gyfathrebu). Roedd Y Tiwtiadur (y pecyn cymorth pedagogaidd – gweler WP4) yn datgan y dylem allu nodi iaith anweddus fel y gellid ei hepgor o gymwysiadau corpws a ddefnyddir â phlant mewn modd systematig, er enghraifft, felly rhoddwyd cyfarwyddyd i drawsgrifwyr nodi achosion posibl ar gyfer eu hadolygu.

Roedd recriwtio trawsgrifwyr yn her barhaus. Fel y trafodwyd uchod, safodd pob darpar drawsgrifydd brawf cychwynnol lle’r oedd angen iddo drawsgrifio darn byr yn unol â chonfensiynau trawsgrifio CorCenCC. Ceir trosolwg manwl o’r penderfyniadau a wnaethpwyd wrth ddatblygu’r confensiynau trawsgrifio ar gyfer CorCenCC, a’r rhesymeg drostynt, yn Knight ac eraill, 2021a.

### 3.4. WP1: Cyfraniadau allweddol

Prif gyfraniad WP1 yw’r data o 11 miliwn o eiriau sy’n ffurfio craidd y corpws. Yn ogystal, mae’r amrediad canlynol o adnoddau, y cafodd eu creu fel rhan o’r broses o gynhyrchu data, yn helpu i gyflawni un o’r nodau a bennwyd ar gyfer prosiect CorCenCC: cynyddu’r gallu ac ymestyn y rhyngwyneb rhwng y Gymraeg (ac ieithoedd eraill sydd wedi’u lleiafrifo o gwmpas y byd, yn eu tro) a disgyblaeth ieithyddiaeth gymhwysol (gan gynnwys, ieithyddiaeth corpws, sosioieithyddiaeth, a chynllunio a pholisi ieithyddol, yn benodol):

- Fframwaith samplu ar gyfer creu corpws cyffredinol o iaith leiafrifol;
- Diffiniad o 'iaith anweddus' sy’n addas ar gyfer cyd-destun iaith leiafrifol lle mae’r holl siaradwyr yn ddwyieithog;
- Set bwrpasol o gonfensiynau trawsgrifio y gellir eu cymhwyso i Gymraeg lafar cyfoes;
- Tîm o gynorthwywyr ymchwil sy’n siarad Cymraeg, sydd wedi derbyn hyfforddiant ar egwyddorion creu corpws, sydd wedi cael y cyfle i weithio gydag arbenigwyr rhyngwladol ar draws y byd, ac sy’n gallu cymhwyso’u sgiliau i brosiectau yn y dyfodol.

Cafodd y gwaith a wnaed gan dîm WP1 ei rannu yn ystod amrediad o gynadleddau rhyngwladol a chenedlaethol, a cheir mwy o fanylion o safbwynt cyfraniadau damcaniaethol/methodolegol CorCenCC yn y cyhoeddiadau (e.e. Knight ac eraill, 2021a, Knight ac eraill, 2021b).

### 3.5. WP1: Cymwysiadau ac effaith

Mae’r broses o gynllunio a gweithredu’r meysydd craidd o waith o fewn WP1 yn cynnig templed i’r bobl hynny sy’n ymchwilio i ieithoedd sydd wedi’u lleiafrifo neu ieithoedd lleiafrifol eraill. Cadarnhaodd adborth a gafwyd yn ystod cynadleddau rhyngwladol fod gan ymchwilwyr mewn lleoliadau eraill ymwybyddiaeth o bosibiliadau trosglwyddo methodoleg a phrosesau CorCenCC i brosiectau ieithyddol eraill (e.e. cynllunio corpws ar gyfer yr Wyddeleg a Malteg yn ystod cynadleddau diweddar).

Bydd y corpws a gasglwyd dan nawdd WP1 yn hwyluso ymchwil academaidd i dueddiadau cyfoes yn y Gymraeg. Mae’r mathau o gwestiynau y gellid mynd i’r afael â nhw, o gofio’n protocolau casglu a thrawsgrifio data, yn cynnwys y canlynol:

- A yw patrymau treiglo’n newid yn y Gymraeg cyfoes ac, os ydynt, pa feysydd a genre sydd ar flaen y gad a pha rai sydd o’r tu ôl?
- Sut mae’r ffurfiau a ddefnyddir mewn e-iaith Gymraeg yn cymharu â’r rheiny a geir mewn iaith lafar ac ysgrifenedig (ac amrywiadau gwahanol ohonynt)?

- Beth mae'r corpws yn ei ddweud wrthym am gyflwr presennol y tafodieithoedd daearyddol Cymraeg 'traddodiadol' ac am amrywiadau newydd sy'n dod i'r amlwg, gan gynnwys eu dosbarthiad cymdeithasol?
- Pa eiriau ac ymadroddion, nad oeddent wedi'u cofnodi fel rhan o'r Gymraeg yn ffurfiol yn flaenorol, a geir yn y corpws, ac a ellir nodi unrhyw dueddiadau?
- Pa mor gyffredin yw'r achosion o newid cod mewn data llafar ac e-iaith, a beth sydd fel petai'n ei achosi?

Bydd CorCenCC yn datgelu llawer mwy am fywiogrwydd a bywioldeb presennol y Gymraeg. Bydd yn rhaid i ni fod yn barod i dderbyn y caiff elfennau o'r datgeliadau hyn eu croesawu ac y caiff elfennau eraill eu gwrthod gan y gymuned ehangach o siaradwyr Cymraeg. Yn y pen draw, pe byddai'n ddymuniad gan y gymuned, gall CorCenCC fod yn gyfrwng ar gyfer trafodaeth sy'n seiliedig ar dystiolaeth ar yr hyn sy'n 'Gymraeg safonol' – neu a allai fod – yn ystod yr unfed ganrif ar hugain.

# 4. Pecyn Gwaith 2: Datblygu set dagiau ar gyfer rhannau ymadrodd a'i defnyddio i dagio data WP1

## 4.1. WP2: Disgrifiad

Diben WP2 oedd gweithredu modd priodol o nodi a labelu ('pennu tagiau') rhannau ymadrodd (e.e. enw, berf, ac is-deipiau o gategorïau o'r fath) a oedd yn nodweddiadol o'r iaith a gasglwyd yn WP1, fel y gellid chwilio a dadansoddi'r corpws yn y ffyrdd amrywiol y byddai'n ofynnol gan ddefnyddwyr yn y dyfodol. Yr offerynnau gofynnol oedd tagiwr rhannau ymadrodd a set dagiau. Cafodd adnodd a oedd eisoes yn bodoli, sef Autoglosser Bangor (Donnelly a Deuchar, 2011), ei ddefnyddio fel y man cychwyn, oherwydd ei ddibynadwyedd a'i addasrwydd profedig ar gyfer pennu tagiau i'r Gymraeg. Dan arweiniad arweinwyr WP2, sef Knight a'r ymgynghorydd prosiect Donnelly (ieithydd cyfrifiadurol sydd wedi gweithio'n agos ar ddatblygu corpysau Cymraeg), cymhwysodd Neale (un o gynorthwywyr ymchwil y prosiect) Autoglosser Bangor i'r set o ddata WP1 a oedd yn dod i'r amlwg, a nododd lle'r oedd angen addasu a mireinio'r offerynnau – mewn perthynas, yn bennaf, â phennu tagiau i ffynonellau iaith lafar ac e-iaith, a'r amrediad ehangach o ran amrywiad rhanbarthol a genre. Cafodd yr addasiadau eu cymhwyso yn ystod camau diweddaraf y prosiect gan Tovey-Walsh (myfyrwraig PhD y prosiect), gan arwain at dagiwr a set dagiau CorCenCC ei hun, sef CyTag (gweler 4.3. isod).

## 4.2. WP2: Amcanion

Roedd nodau ac amcanion WP2 fel a ganlyn:

- adeiladu tagiwr rhannau ymadrodd y Gymraeg a'i hyfforddi
- datblygu set dagiau briodol
- pennu tagiau i'r holl ddata

## 4.3. WP2: Cyflawniadau

Cafodd meddalwedd bwrpasol CorCenCC ar gyfer pennu tagiau i rannau ymadrodd, CyTag, ei datblygu yn ystod deunaw mis cyntaf y prosiect, a chafodd ei rhyddhau yn gyhoeddus ym mis Mawrth 2018. Mae gwerthusiadau a chymwysiadau CyTag hyd yn hyn yn nodi ei bod yn gyflawn ac yn gadarn a'i fod yn gweithio'n dda. Mae CyTag yn liferu deunyddiau ffynhonnell agored i helpu gyda'r broses o benderfynu ar rannau ymadrodd. Mae'n gweithio yn bennaf drwy ddefnyddio'r wybodaeth sydd yng ngeiriadur Donnelly, *Eurfa* – y geiriadur Cymraeg ffynhonnell agored mwyaf sydd ar gael yn rhad ac am ddim (Donnelly, 2013a) – i lunio rhestr o dagiau posibl ar gyfer pob gair mewn testun Cymraeg. Ategir hyn gan restrau penodol o enwau lleoedd, enwau cyntaf a chyfenwau sydd wedi'u dethol o ddata Wikipedia.

Unwaith y mae rhestr o eiriau wedi'i llunio, gellir cymhwyso set o reolau pwrpasol i fireinio'r rhestr o dagiau posibl a bennwyd ar gyfer pob gair – yn seiliedig ar y tagiau neu'r nodweddion a bennwyd i'r geiriau o'i gwmpas – nes cyrraedd yr un cywir. Er enghraifft, gall y ffurf 'yn' olygu 'in', ond mae ganddi hefyd ddwy rôl fel geiryn gramadegol. Mewn un rôl, mae'n trosi ansoddair yn adferf (e.e. *'*yn dda*'*, o 'da'). Yn y llall, mae'n cael ei gysylltu â'r ferf 'bod' (un ffurf ohoni yw 'mae') er mwyn cyflwyno berf, enw neu ddibeniad ansoddeiriol (e.e. 'mae'r llyfr yn dda'). Fel y gwelir o'r enghreifftiau hyn, gall 'yn dda' olygu 'good' a 'well', ond yn y ddwy achos byddai 'yn' yn cael ei ddosbarthu'r un ffordd. Roedd angen i'r tagiwr allu gwahaniaethu rhwng 'yn' fel arddodiad ac 'yn' fel geiryn, ac mae'n gwneud hynny yn y ffordd ganlynol: Yn y frawddeg 'mae Cymru yn wlad Geltaidd', mae CyTag yn pennu tag cywir i 'yn' fel geiryn sy'n cyflwyno dibeniad, oherwydd bod 'mae' o'i flaen, ac oherwydd bod 'wlad' yn dreiglad meddal o 'gwlad'; mae gennym reol i ddewis tag y geiryn dibeniadol ar gyfer 'yn' lle bo'r gair canlynol yn enw sydd wedi'i dreiglo'n feddal. (Pan fo 'yn' yn golygu 'in', caiff hyn ei ddilyn gan dreiglad trwynol.) Dangosodd ganlyniadau'r broses werthuso fod CyTag yn cyflawni lefel o gywirdeb o dros 95%, sy'n gymharol â'r enghreifftiau gorau o dagwyr rhannau ymadrodd ar gyfer ieithoedd eraill. Ar hyn o bryd, mae CyTag yn cynnwys rhannwr testun, holltwr brawddegau, tocynnwr, a'r tagiwr rhannau ymadrodd ei hun. Ceir gwefan CyTag yn: http://cytag.corcencc.org.

Mae set dagiau rhannau ymadrodd CyTag yn cynnwys 145 o dagiau manwl, h.y. tagiau 'cyfoethog', sydd wedi'u mapio'n 13 categori sy'n cydymffurfio â Safonau'r Grŵp Cynghori Arbenigol ar Beirianneg Ieithyddol (EAGLES, 1996). Mae'r tagiau hyn yn cynnwys categorïau cystrawennol pwysig (e.e. enw, bannod, arddodiad, berf ac ati) yn ogystal â dau gategori sy'n cynrychioli geirynnau sy'n 'unigryw' i'r Gymraeg a ffurfiau 'eraill' fel byrfoddau, acronymau, symbolau, digidau ac ati. Mae'r set lawn o 145 o dagiau yn cwmpasu morffoleg y Gymraeg yn seiliedig ar genedl (gwrywaidd neu fenywaidd), rhif (unigol neu luosog), person (cyntaf, trydydd ac ati) a'r amser (gorffennol, presennol, dyfodol ac ati). Mae'r tagiau eu hunain wedi'u hamgodio'n Gymraeg.

Ceir y set dagiau lawn yn: https://cytag.corcencc.org/tagset?lang=cy. Gweler Neale et al., 2018 am drosolwg technegol trylwyr o CyTag a gwerthusiad manwl o'i chywirdeb.

Un o fanteision pwysig dull CorCenCC o safbwynt datblygu meddalwedd sy'n pennu tagiau yw'r modd y mae'n defnyddio'r nifer lleiaf o reolau sy'n hawdd eu cymhwyso i wybodaeth ac adnoddau sydd eisoes yn bodoli. Mae modd trosglwyddo'r dull hwn i ieithoedd y mae data hyfforddi sydd wedi'i anodi ymlaen llaw amdanynt yn brin, gan ei wneud yn werthfawr ar gyfer cipio nodweddion ieithoedd lleiafrifol.

## 4.4. WP2: Cyfraniadau allweddol

Prif gyfraniad gwaith WP2 oedd creu offerynnau meddalwedd ac adnoddau ieithyddol sydd ar gael yn rhad ac am ddim ac sy'n ymestyn yr adnodd sydd eisoes yn bodoli ar gyfer dadansoddi, a chloddio testun, y Gymraeg. Yn benodol, gwnaethom greu'r canlynol:

- Set dagiau tagiwr rhannau ymadrodd CorCenCC, gyda:
  - 145 o dagiau rhannau ymadrodd 'cyfoethog'
  - 13 categori 'sylfaenol' sy'n cydymffurfio ag EAGLES

  Gellir mabwysiadau'r set dagiau hon fel rhestr safonol (h.y. set o gonfensiynau) a/neu ychwanegu ati a/neu ei haddasu ymhellach gan ddefnyddwyr yn y dyfodol wrth bennu tagiau i setiau data'r Gymraeg.
- Corpws gwerthuso o 'safon aur'. Mae corpws safon aur yn un sydd wedi'i anodi â llaw a'i wirio gan unigolion lluosog. Mae hyn, mewn effaith, yn darparu model sy'n gallu hyfforddi'r dull awtomatig (cyfrifiadurol) a'i werthuso. Mae'r corpws safon aur hwn hefyd wedi'i ryddhau er mwyn i ymchwilwyr eraill allu ei ddefnyddio yn y gwaith o ddatblygu eu hofferynnau eu hunain. Mae'n cynnwys y canlynol:
  - 611 o frawddegau
  - 14,876 o docynnau
- Gwefan CyTag: (https://cytag.corcencc.org)

  Mae'r wefan hon yn cynnwys gwybodaeth ddwyieithog (Cymraeg a Saesneg) am y tagiwr, gan gynnwys arddangosiad gweithiol (gweler Ffigur 1a ac 1b am sgrinluniau yn Gymraeg a Saesneg yn ôl eu trefn). Gall defnyddwyr ddefnyddio'r tagiwr drwy'r wefan hon a'i ddefnyddio i bennu tagiau i'w data eu hunain.
- CyTag ar Github (https://github.com/CorCenCC/CyTag)

  Mae'r tagiwr ffynhonnell agored hefyd ar gael i bawb (h.y. ar gael fel meddalwedd rhad ac am ddim, dan delerau fersiwn 3 Trwydded Gyhoeddus Gyffredinol GNU) drwy wefan GitHub. Yma, gall defnyddwyr eto lawrlwytho'r tagiwr a'i ddefnyddio. Gallant hefyd wella'r tagiwr a rhannu fersiynau a ddiweddarwyd â defnyddwyr eraill y dyfodol.

***Ffigur 1a.*** Sgrinlun o ryngwyneb arddangosol ar-lein CyTag yn Gymraeg

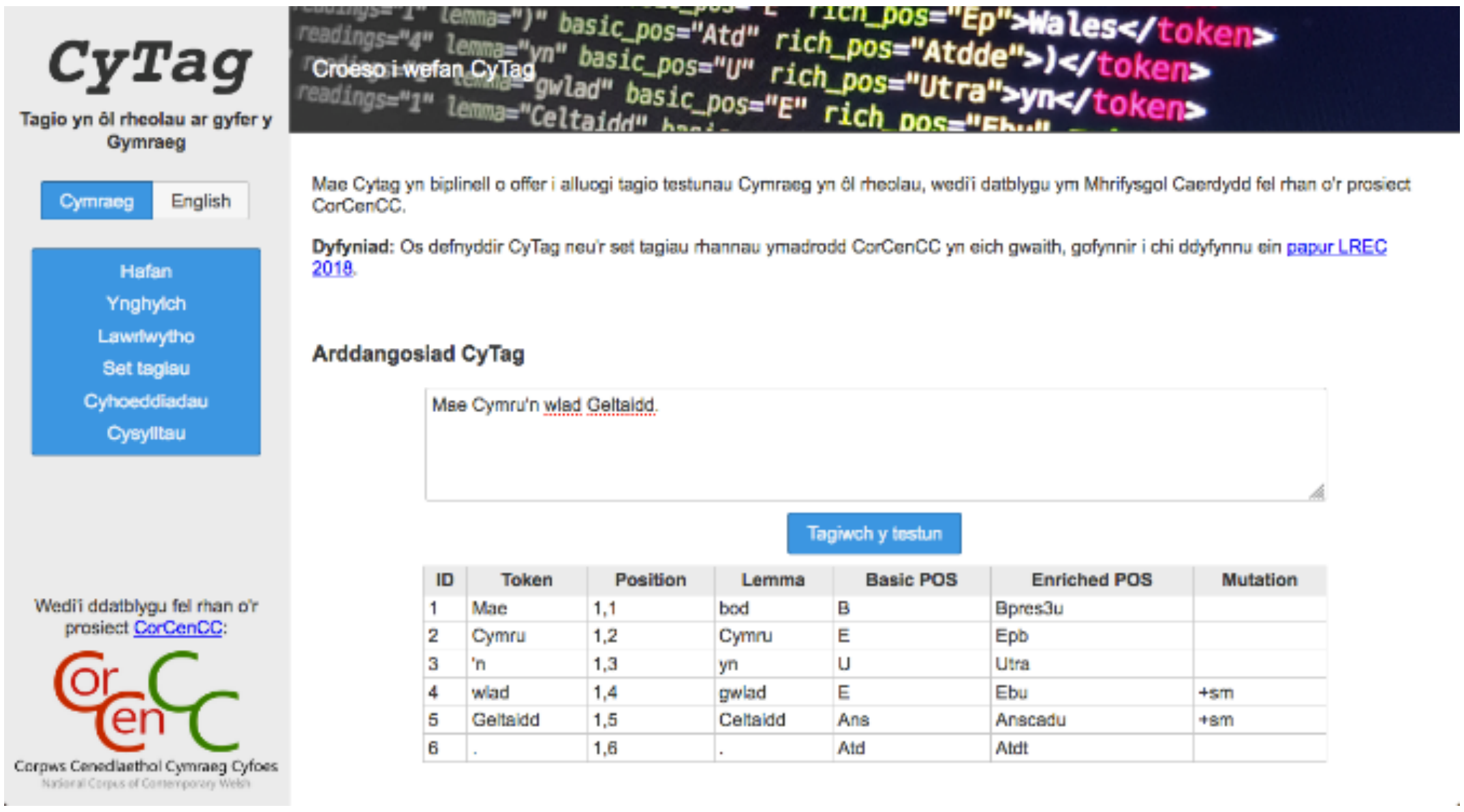

***Ffigur 1b.*** Sgrinlun o ryngwyneb arddangosol ar-lein CyTag yn Saesneg (ar gael yn https://cytag.corencc.org)

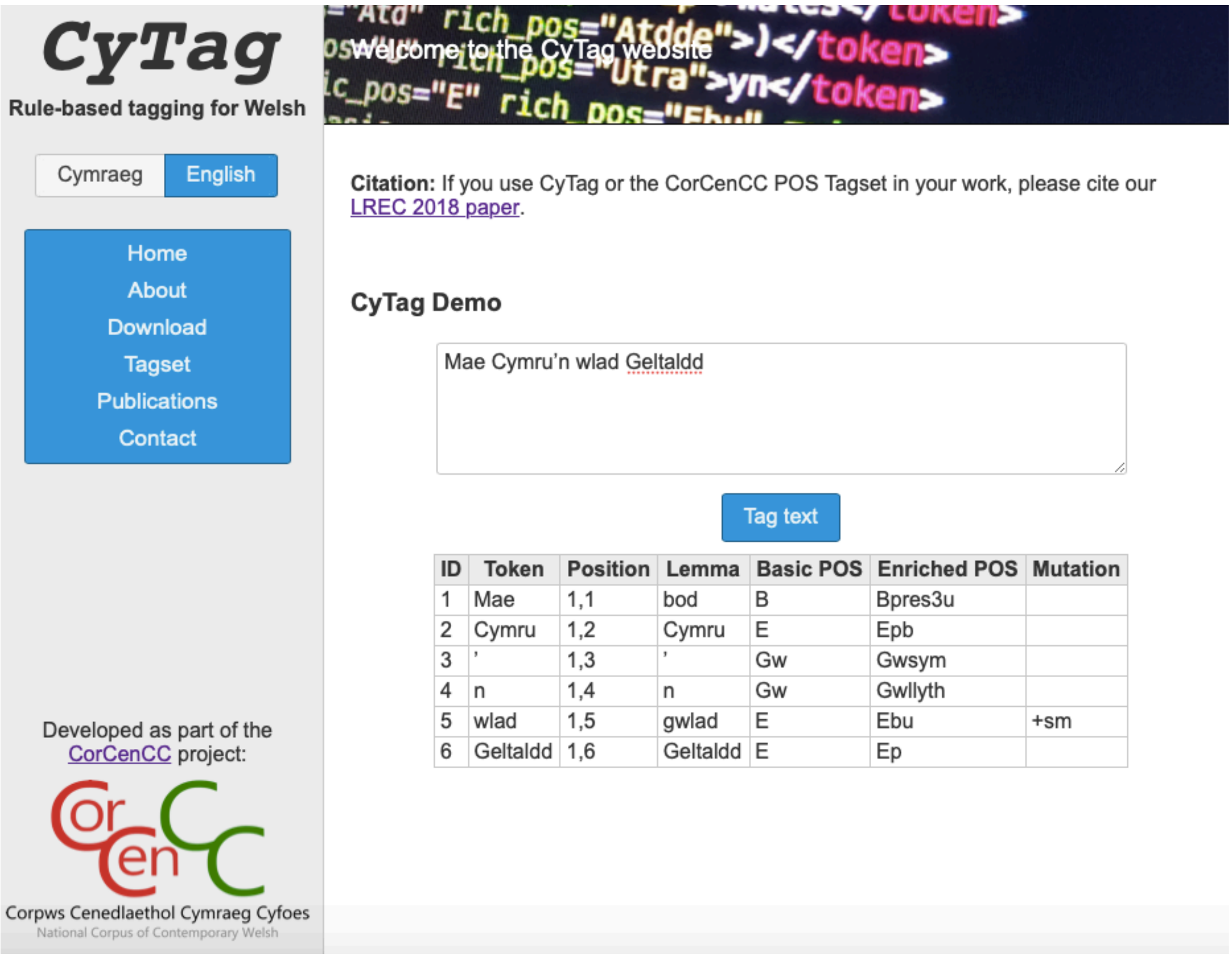

Mae fersiwn gyfredol CyTag yn cynnwys y gwelliannau canlynol a wnaed gan Tovey-Walsh:

- fersiwn ddiwygiedig o'r tocyneiddiwr
- diweddariad i'r modd yr ymdrinnir â'r treigladau sy'n gwella sut mae'n medru dal treiglad mewn geiriau byrrach ac enwau priod
- cynnwys rhai geiriau gyda phrif gymeriad â llythyren fach yn y lecsicon (e.e. 'cymraeg')
- cynnwys geiriau Saesneg amlder uchel fel bod modd i'r tagiwr eu hadnabod (fel geiriau nad ydynt yn rhai Cymraeg) sy'n lleihau nifer y geiriau sy'n cael eu tagio'n 'anhysbys'
- cynnwys banodolion a rhagenwau amhendant yn y set o dagiau (e.e. 'neb', 'pob') a'r modiwl tagio gan ychwanegu'r rheolau hyn i ramadeg cyfyngu CyTag

## 4.5. WP2: Cymwysiadau ac effaith

Mae CyTag ar gael fel tagiwr ynddo'i hun. Mae hyn yn golygu y gall unrhyw un sy'n gwybod sut i ddefnyddio tagiwr ei ddefnyddio i bennu tagiau i'w ddata ei hun. Mae CyTag wedi'i chymhwyso i CorCenCC yn ei gyfanrwydd, fel nad oes angen i ddefnyddwyr wybod sut i ddefnyddio tagiwr er mwyn cael gwybodaeth bwysig am yr iaith. Gallai defnyddwyr o'r fath gynnwys ymchwilwyr, athrawon iaith, datblygwyr technoleg a geiriadurwyr. Gall defnyddwyr eraill hefyd wella CyTag yn y dyfodol.

# 5. Pecyn Gwaith 3: Cwmpasu a datblygu tagiwr semantig ar gyfer y Gymraeg a'i ddefnyddio i bennu tagiau semantig i ddata WP1

## 5.1. WP3: Disgrifiad

Cafodd Pecyn Gwaith 3 (WP3) ei arwain gan Rayson, gyda Piao yn gweithredu fel cysylltai ymchwil hyd at fis Awst 2018 (pan ymgymerodd â swydd darlithio amser llawn), pan ymunodd Ezeani â'r tîm. Defnyddiodd tîm WP3 yr arbenigedd yn y Gymraeg a oedd ar gael ar draws gymuned gyfan prosiect CorCenCC yn ôl yr angen. Prif elfen gwaith ymchwil WP3 oedd cynllunio a datblygu'r system feddalwedd ar gyfer anodiadau semantig y Gymraeg a fyddai'n llywio'r adnoddau ieithyddol cysylltiedig.

## 5.2. WP3: Amcanion

Roedd gan WP3 bedwar nod ac amcanion fel a ganlyn:

- cynllunio set dagiau semantig newydd ar gyfer y Gymraeg
- datblygu'r system anodi awtomatig
- rhoi prawf ar ddulliau torfol ar gyfer pennu tagiau semantig
- pennu tagiau semantig i'r holl ddata

Roedd y gwaith ymchwil a gyflawnwyd yn WP3 yn adeiladu ar gwerth 30 blynedd o waith dadansoddi semantig awtomatig ym maes ieithyddiaeth corpws ac ieithyddiaeth gyfrifiadurol yng nghanolfan ymchwil Prifysgol Caerhirfryn. Roedd y gwaith helaeth a oedd eisoes wedi'i wneud ar ieithoedd eraill (Ffinneg, Rwsieg, Tsieineeg, Iseldireg, Eidaleg, Portiwgaleg, Sbaeneg, Maleieg) o werth arbennig ar gyfer CorCenCC.

Fel rhan o brosiect CorCenCC, roedd angen i ni ailwerthuso set dagiau System Dadansoddi Semantig UCREL (USAS) er mwyn darparu ar gyfer nodweddion arbennig y Gymraeg, a gofynion ymarferol y gwaith o ddatblygu'r pecyn cymorth pedagogaidd (WP4) a defnyddwyr terfynol y corpws (WP5). Gwnaethom hefyd anelu at ddatblygu algorithmau a dulliau newydd i ddynodi meysydd semantig a oedd yn briodol i'r cyd-destun i unedau geiriadurol y Gymraeg, boed y rheiny'n eiriau unigol neu'n ymadroddion â geiriau lluosog. Hefyd, i ategu at y dull torfol o gasglu data'r corpws yn WP1, gwnaethom anelu at beilota dulliau cyfrannu torfol ar gyfer dynodi meysydd semantig er mwyn ymestyn y geiriaduron semantig gwaelodol a'u galluogi i adlewyrchu'r hyn a ddehonglwyd gan siaradwyr Cymraeg.

## 5.3. WP3: Cyflawniadau

Y dasg gyntaf oedd creu set dagiau USAS ar gyfer y Gymraeg. Cyflawnwyd hyn drwy ddefnyddio dulliau a oedd wedi'u datblygu yn ystod prosiect blaenorol a ariannwyd gan y Cyngor Ymchwil i'r Celfyddydau a'r Dyniaethau (AHRC), sef SAMUELS (AH/L010062/1), a diweddaru'r fframwaith Java ar gyfer y Gymraeg. Roedd y gwaith dadansoddi semantig newydd yn y Gymraeg yn cynnwys y camau canlynol.

Ymgymerwyd â'r gwaith o fapio meysydd semantig draw oddi wrth y fframwaith amlieithog a oedd eisoes yn bodoli, fel y gellid adolygu ei addasrwydd ar gyfer y Gymraeg. Roedd yr ystyr posibl a ddynodwyd i dagiau wedi deillio'n awtomatig, yn y lle cyntaf felly, o'r broses o drosi geiriaduron Saesneg trwy eiriaduron dwyieithog a chorpysau bach a oedd yn cydredeg, ac yna cawsant eu gwirio gan siaradwyr Cymraeg tîm CorCenCC, a'u haddasu yn ôl yr angen. Cafodd y set dagiau semantig newydd ar gyfer y Gymraeg ei rhyddhau ym mis Ebrill 2016.

Roedd ein gwaith ymchwil ar gyfrannu torfol yn gofyn am greu dull ac arbrofion a olygodd gael cyfranogiad gan ddefnyddwyr Amazon Mechanical Turk (AMT) er mwyn gwneud ymchwil i'r graddau y gallai siaradwyr Cymraeg amhroffesiynol nad oeddent wedi'u hyfforddi greu cofnod geiriadurol semantig o safon eithaf uchel drwy ddynodi un neu fwy o feysydd semantig addas, yr oedd wedi'u pennu ymlaen llaw, i air neu ymadrodd o'r corpws. Cafodd y gwaith ymchwil hwn ei gyhoeddi yn Rayson a Piao, 2017 (gweler adran 10.2.).

Roedd angen tagiwr rhannau ymadrodd arnom er mwyn i ni allu gweithio ar WP3, ac er mwyn gwneud cynnydd â hynny cyn bod WP2 wedi'i gwblhau, gwnaethom ddefnyddio tagiwr rhannau ymadrodd dros dro a oedd eisoes yn bodoli, a oedd yn gydran o Becyn Cymorth Iaith Naturiol y Gymraeg (WNLT), fel rhan o'r tagiwr semantig newydd ar gyfer y Gymraeg a gafodd ei greu yn Java (CySemTagger). Yn ddiweddarach, pan oedd allbwn WP2 ar gael, gwnaethom fabwysiadu CyTag. I ddechrau, roedd gwefan rhyngwyneb rhaglennu cymwysiadau (API) (SOAP) ar gael ar gyfrif GitHub UCREL, a alluogodd defnyddwyr i gyrchu'r cymhwysiad. Yn ddiweddarach, roedd rhyngwyneb rhaglennu cymwysiadau newydd (REST) ar gael yn http://ucrel-api.lancaster.ac.uk/, o oedd unwaith eto â'r nod o gynyddu hygyrchedd i'r cymhwysiad hwn.

Cafodd fframwaith Java ei greu gan Piao ar gyfer pennu tagiau i eiriau unigol ac ymadroddion amleiriol, a thrwy ddefnyddio dulliau amrywiol cafodd yr adnoddau ieithyddol eu creu. Mae'r adnoddau terfynol yn cynnwys 143,287 o gofnodion o eiriau unigol a chasgliad o sampl o gofnodion amleiriol, yn ogystal â 329,800 o ffurfdroadau o gorpysau amrywiol. Cafodd fersiwn ar y we o ryngwyneb defnyddiwr graffigol (GUI) y tagiwr semantig Cymraeg ei chreu (gan ddefnyddio rhyngwyneb rhaglennu cymwysiadau SOAP), yn ogystal â fersiwn bwrdd gwaith ohono. Trwy gydweithrediad â'r tîm CorCenCC ehangach, cafodd corpws o safon aur ei wirio â llaw at ddibenion gwerthuso a gwella'r tagiwr (ceir manylion o'r corpws safon aur yn 4.4.).

Er mwyn cwblhau ein gwaith ymchwil ar gyfer y pecyn gwaith hwn, ymgymerodd Ezeani (cynorthwyydd ymchwil) ag arbrawf dysgu aml-dasg er mwyn ymchwilio i ba raddau y gellid defnyddio modelau ymgorffori geiriau o'r radd flaenaf sy'n seiliedig ar fectorau ar gyfer ieithoedd sydd ag adnoddau prin (Cymraeg yn ein hachos ni) gyda modelau rhwydwaith niwral ar gyfer pennu tagiau i rannau ymadrodd a phennu tagiau semantig. Dangosodd ein canlyniadau fod dull o bennu tagiau o'r fath yn cymharu'n dda iawn â'r tagwyr a oedd eisoes yn bodoli (gweler Rayson a Piao, 2017 – adran 10.2). Rydym wedi sicrhau bod ein tagwyr a'n hadnoddau ieithyddol i gyd ar gael ar ffurf ffynhonnell agored gyda thrwyddedau caniatadol.

## 5.4. WP3: Cyfraniadau allweddol

Un o brif gyfraniadau gwaith ymchwil WP3 yw'r offerynnau meddalwedd a'r adnoddau ieithyddol sydd ar gael yn rhad ac am ddim ac sy'n ymestyn y banc adnoddau ar gyfer dadansoddi, a chloddio testun, y Gymraeg. Mae'r cod Java ar gyfer CySemTagger wedi'i ryddhau ar GitHub ac mae wedi'i ymgorffori yn system anodi a dadansoddi corpws Wmatrix (Rayson ac eraill, 2004). Gwneir defnydd helaeth o'r system hon ym maes ieithyddiaeth corpws, ac mae'n golygu y gall ymchwilwyr ei defnyddio ar gyfer corpysau'r Gymraeg. Yn gyffredinol, mae gwaith WP3 wedi ymestyn cwmpas y gwaith ymchwil ym meysydd ieithyddiaeth corpysau ac ieithyddiaeth gyfrifiadurol mewn o leiaf dwy ffordd. Yn gyntaf, mae wedi arddangos dull ar gyfer ymestyn technegau dadansoddi semantig i heriau penodol y Gymraeg mewn modd effeithiol. Yn ail, mae wedi dangos y gellir defnyddio dulliau cyfrannu torfol i gyfrannu at y gwaith o ddatblygu adnoddau o'r fath.

### 5.5. WP3: Cymwysiadau ac effaith

Mae'r tagiwr semantig a ddatblygwyd yn WP3 wedi'i gymhwyso i CorCenCC yn ei gyfanrwydd, gan gynhyrchu cyfoeth o ddehongliadau o'r data a fydd o werth i'r bobl hynny sydd â diddordeb yn y ffordd y defnyddir y Gymraeg cyfoes i gyfleu ystyr ar draws genres, arddulliau a chyfryngau, a sut gellir awtomeiddio prosesau ieithyddol ar gyfer technolegau newydd. Yn ogystal, mae'r egwyddorion gwaelodol a sefydlwyd wrth greu'r CySemTagger yn darparu sylfaen ar gyfer estyniadau pellach yn y dyfodol fel y gellir dadansoddi ieithoedd eraill sydd ag adnoddau prin. Mewn modd tebyg, bydd ymchwilwyr yn gallu ymestyn yr arbrofion dysgu aml-dasg i ieithoedd eraill er mwyn ymchwilio i ba raddau y gall yr ieithoedd hynny fanteisio hefyd, er gwaethaf eu fframweithiau gramadegol gwahanol. Mae'r ffaith fod y CySemTagger wedi'i ymgorffori yn Wmatrix (Piao ac eraill, 2018) yn golygu y gall ymestyn ar draws y gymuned ymchwil ryngwladol, gan alluogi gwaith dadansoddi cynnwys awtomatig ar gorpysau'r Gymraeg y gellid eu casglu gan bobl eraill yn y dyfodol.

# 6. Pecyn Gwaith 4: Cwmpasu, dylunio ac adeiladu pecyn cymorth pedagogaidd ar-lein

### 6.1. WP4: Disgrifiad

Cafodd waith WP4 ei arwain gan Thomas a Fitzpatrick, a weithiodd â'r cynorthwywyr ymchwil Needs a J. Davies, ac arweinydd prosiect Knight, a chafwyd cyngor ymgynghorol gan Stonelake (cyd-ymchwilydd), Anthony (ymgynghorydd prosiect – dyluniwr a datblygwr AntConc), Cobb (ymgynghorydd prosiect – datblygwr Compleat Lexical Tutor) ac E. Davies (CBAC).

Mae dysgu/addysgu iaith yn un maes lle gall corpysau fod yn arbennig o addysgiadol. Wrth i athrawon a dysgwyr ddod yn fwy deallus wrth ddefnyddio technoleg, ac wrth i gorpysau barhau i ddatblygu o ran eu maint a'r modd o'u cymhwyso a'u hymarferoldeb, mae dysgu sydd wedi'i lywio gan gorpysau'n ennill ei blwyf yn gyflym o safbwynt lleoliadau dosbarth ac astudiaethau personol. Gellir defnyddio corpysau i amlygu'r geiriau, ymadroddion a phatrymau mwyaf cyffredin mewn iaith benodol. Gallant ddangos pa eiriau sy'n dueddol o fynd gyda'i gilydd, a pha rai sy'n digwydd ym mha fathau o destun (e.e. testunau ysgrifenedig ffurfiol, sgyrsiau llafar, negeseuon e-bost proffesiynol, neu negeseuon testun personol). Gall defnyddwyr corpysau chwilio am eiriau penodol a'u gweld mewn brawddegau enghreifftiol. Gall corpysau, felly, ddarparu ffynhonnell sydd â chyfoeth o iaith i'r dysgwr sy'n arddangos sut mae ei iaith darged yn cael ei defnyddio mewn gwirionedd, ac o fewn meysydd amrywiol.

Un o agweddau arloesol prosiect CorCenCC, a arweiniwyd gan dîm WP4, yw datblygiad cyfres o offerynnau pwrpasol – Y Tiwtiadur – a fwriedir iddo gael ei ddefnyddio yn ystod dosbarthiadau Cymraeg a'r tu allan iddynt, o ysgolion cynradd i addysg i oedolion. Gyda'i gilydd, mae'r offerynnau hyn yn helpu i arddangos sut y caiff y Gymraeg ei defnyddio mewn gwirionedd mewn pedwar ymarfer arwahanol sy'n seiliedig ar y corpws sy'n defnyddio data a gasglwyd yn WP1 a'r tagwyr / setiau tagiau a ddatblygwyd yn WP2 a WP3. Mae'r offerynnau'n cynnwys y canlynol:

- offeryn Cau Bylchau (Cloze) sy'n galluogi athrawon (neu ddysgwyr, mewn cyd-destunau astudiaethau personol) i ddileu geiriau o unrhyw ddarn o destun yn y corpws, fesul bwlch penodol, er mwyn annog neu asesu medrau deall a strategaethau rhagfynegi
    - Mae'r offeryn hwn yn galluogi defnyddwyr i greu tasg llenwi bylchau gan ddefnyddio darnau o destun yn CorCenCC. Mae'r opsiwn "Math o destun" yn ei gwneud yn bosibl i ddefnyddwyr ddethol genres o destun arbennig (e.e. genres 'blog' neu 'ffuglen llyfrau'). Mae'r gosodiad "Amlder bylchau" yn galluogi defnyddwyr i osod pa mor aml y maent am i fwlch ymddangos, gan ddibynnu ar ba mor anodd y mae angen i'r dasg fod (y gosodiad a argymhellir yw pob seithfed i nawfed gair). Gan ddefnyddio'r opsiwn "Hyd y testun", gall defnyddwyr ddewis gweld sampl ar hap o ddarnau o destun hyd at 100, 200, 300, 400, neu 500 o eiriau o hyd. Wrth glicio "Dechrau", mae panel newydd yn dangos y dasg llenwi bylchau ac mae'r geiriau sydd wedi'u tynnu allan o'r darn o destun yn ymddangos mewn panel ar wahân. Er mwyn cwblhau'r dasg, mae'n ofynnol i ddefnyddwyr ddewis geiriau o'r rhestr a'u teipio i mewn i'r bylchau priodol yn y darn o destun. Pan gaiff "Gwirio" ei glicio, caiff y geiriau sydd wedi'u gosod yn gywir eu hamlygu'n wyrdd, a'r geiriau na osodwyd yn gywir yn goch. Mae argaeledd y darnau o destun o'r corpws yn galluogi athrawon a dysgwyr i gwblhau'r gweithgaredd hwn lawer o weithiau, a cheir cyfleoedd dysgu newydd bob tro.
- offeryn Proffilydd Geiriau sy'n ei gwneud yn bosibl graddio darnau o destun yn ôl pa mor aml mae geiriau'n ymddangos
    - Mae'r offeryn hwn yn pennu proffil i ddarn o destun sydd wedi'i ddethol neu'i greu gan y defnyddiwr yn ôl pa mor aml mae geiriau'n ymddangos. Mae'n ofynnol i ddefnyddwyr gopïo a gludo darn o destun i'r maes "Mewnbynnu testun" neu i deipio testun i'r maes yn uniongyrchol. Wrth glicio ar "Dechrau" mae proffil yn cael ei greu, lle caiff pob gair ei ddynodi i gategori yn ôl pa mor aml y mae'n ymddangos. Mewn panel ar wahân, ceir esboniad o'r canlyniadau. Mae'r colofnau "Lefel" / "Band amlder" yn ymwneud â'r nifer o weithiau y mae gair yn ymddangos yn y 10 miliwn o eiriau a geir yn CorCenCC. Mae geiriau sy'n perthyn i'r band "K1" (y 1,000 uchaf) ymhlith 1,000 o eiriau mwyaf cyffredin y Gymraeg, yn ôl CorCenCC. Fel arfer, y mwyaf o eiriau sydd wedi'u cynnwys mewn darn o destun sy'n perthyn i'r bandiau amlder isaf (e.e. y rheiny o fewn 3,001 – 4,000 (K4), 4,001 – 5,000 (K5) a >5,001 (K6)), y mwyaf heriol y bydd i'r dysgwr ei ddeall. Gellir hefyd pennu proffil i ddarnau o destun a gynhyrchir gan ddysgwyr; mae dysgwyr fel arfer yn casglu mwy o eiriau sydd ag amlder uchel i ddechrau, ac yn meistroli bandiau amlder isel wrth i'w hyfedredd ddatblygu (gweler, er enghraifft, Nation, 2001). Yn y gosodiad diofyn, bydd yr offeryn yn amlygu geiriau sy'n perthyn i lefelau K1 i K6+. Gall defnyddwyr newid yr offeryn i amlygu geiriau nad ydynt yn perthyn i'r lefelau

hyn drwy glicio ar yr opsiwn "Amlygu geiriau nad ydynt yn perthyn i lefel". Gall geiriau sy'n perthyn i'r band 5,001+ gynnwys geiriau sydd wedi'u camsillafu a geiriau o ieithoedd eraill, yn ogystal â geirfa a ddefnyddir yn anaml yn y corpws, neu nad yw wedi'i chipio gan y corpws.

- offeryn Nodi Geiriau sy'n profi gallu dysgwyr i ddyfalu gair o fewn cyd-destun
  - Mae'r offeryn hwn yn arddangos darnau corpws lluosog (llinellau cydgordio) sydd i gyd yn cynnwys gair penodol. Mae'r gair wedi'i guddio, a'r dasg yw nodi'r gair sy'n cyfateb â'r bylchau i gyd. Ceir opsiynau ar gyfer dethol "Band amlder" y gair (K1, K2 a K3), y "Math o air" penodol (e.e. enw, berf) a'r nifer fwyaf o frawddegau i'w harddangos. Er mwyn cynhyrchu'r darnau hyn, mae defnyddwyr yn clicio ar "Dechrau". Yn yr offeryn hwn, yr ymatebion 'cywir' yw'r rheiny sydd wedi'u cynnwys yn CorCenCC; mewn rhai achosion, gallai ymateb gwahanol hefyd fod yn gredadwy o fewn yr iaith ar y cyfan.
- offeryn Creu Tasgau Geiriau sy'n hwyluso gwaith dwys ar eitem benodol o eirfa
  - Mae'r offeryn hwn yn cynhyrchu darnau corpws lluosog (llinellau cydgordio) o CorCenCC sydd i gyd yn cynnwys gair targed a bennir gan y defnyddiwr. Mae defnyddwyr yn teipio'u gair targed i mewn i'r blwch mewnbynnu "Gair". Yna, maent yn dethol sawl darn maent am eu cynhyrchu (uchafswm o 20) gan ddefnyddio'r opsiwn "Uchafswm llinellau". Os yw defnyddwyr am bennu rhan ymadrodd y gair targed (e.e. enw, berf), gallant wneud hynny drwy'r opsiwn "Rhan ymadrodd". Mae'r dasg yn cael ei chreu drwy glicio ar "Dechrau". Mae'r offeryn hwn yn gwneud dau fath o weithgaredd dysgu'n bosibl. Un ohonynt yw arsylwi ar y geiriau sydd o gwmpas y gair penodol. Mae hyn yn helpu'r broses o gasglu strwythurau gramadegol a phatrymau cydleoliad. Y llall yw ffurf fwy firieniol ar yr offeryn Nodi Geiriau uchod, lle gall athro bennu'r gair penodol i'w ddyfalu, yn hytrach na gair a gynhyrchir gan y feddalwedd o set yn unig. Er mwyn hwyluso ail ddefnydd yr offeryn hwn, mae'r gair wedi'i guddio yn y tabl canlyniadau. Mae clicio ar "Dangos" yn datgelu'r gair targed.

Mae'r offerynnau hyn yn ei gwneud yn bosibl i ddefnyddwyr weithio gyda'r iaith mewn sawl ffordd. Er enghraifft, gallant archwilio patrymau cydgordio ar draws sawl cystrawen (e.e. berf + arddodiad, ansoddair + arddodiad, cysylltair (os, fel) + amser dilynol), a dyfalu'r gair coll drwy archwilio a dadansoddi geiriau eraill a ddefnyddir ar y cyd ac yn y cyd-destun yn yr amgylchedd uniongyrchol. Gallant nodi bylchau yn eu geirfa, a rhoi blaenoriaeth i ba eiriau i'w dysgu nesaf. Ochr yn ochr â'r offerynnau ymholi cyffredinol y gellir eu defnyddio yn CorCenCC, mae Y Tiwtiadur yn enghraifft unigryw o ddysgu a yrrir gan ddata (Johns, 1991) ar ffurf addysgeg anwythol a ddefnyddir yn uniongyrchol ac sy'n seiliedig ar sy'n gallu helpu i ychwanegu at ddysgu'r Gymraeg ar hyd oes.

## 6.2. WP4: Amcanion

Roedd y gwaith a wnaed yn WP4 yn ymateb i dri amcan fel a ganlyn:

- dylunio a chynhyrchu pecyn cymorth pedagogaidd ar-lein sy'n seiliedig ar y Gymraeg (fel a ddisgrifir uchod) sy'n gweithio'n uniongyrchol â data'r corpws er mwyn cefnogi dysgu ac addysgu ieithyddol

- cynhyrchu rhestrau o eiriau addysgegol sy'n seiliedig ar eu hamlder

Un o nodweddion arloesol allweddol prosiect CorCenCC yw ei fod yn integreiddio corpws â phecyn cymorth addysgeg ar-lein. Mae Y Tiwtiadur yn gweithio'n uniongyrchol â data'r corpws er mwyn cefnogi dysgu ac addysgu ieithyddol drwy ddarparu cyfleoedd estynedig i archwilio darnau o Gymraeg go iawn. Lle ceir offerynnau corpws addysgegol cyn hyn sydd wedi bod yn ategiadau eilaidd i gorpysau a oedd eisoes yn bodoli, yn yr achos hwn, ac o'r cychwyn, roedd cynllun CorCenCC yn cynnwys rhyngwyneb addysgol er mwyn cefnogi dysgu ac addysgu'r Gymraeg. Ysbrydolwyd Y Tiwtiadur gan adnodd ar-lein y Compleat Lexical Tutor (Lextutor – https://www.lextutor.ca/ – Cobb, 2000) – un o'r pecynnau cymorth dysgu iaith ar-lein sydd wedi'i yrru gan ddata sydd â'r proffil uchaf ac sydd wedi'i gyrchu fwyaf (tua 15,000 o ddefnyddwyr y dydd). Mae'r bedair dasg a geir yn Y Tiwtiadur yn seiliedig ar rai o'r ymarferion mwyaf poblogaidd sydd i'w gweld yn Lextutor.

Un o'r nodweddion arloesol eraill yw'r ffaith fod y pecyn cymorth yn seiliedig ar ddysgu a yrrir gan ddata (Johns, 1991), lle 'mae dysgwyr yn archwilio'r dystiolaeth ac yn chwilio am batrymau yn y data y gallant gyffredinoli ohonynt' (Thompson, 2005: 10). Diben y pedwar ymarfer addysgegol yw galluogi ac annog dysgwyr (yn rhai iaith gyntaf ac ail iaith) i arsylwi ar batrymau iaith y Gymraeg, a dod i gasgliadau ohonynt er mwyn hwyluso dysgu. Mae hyn yn hanfodol bwysig i annibyniaeth dysgwyr (gweler Aston, 2001; Little, 2007). Yn hytrach na fo'r athro'n dweud wrth y myfyriwr sut mae'r iaith yn gweithio, caiff y myfyrwyr eu cefnogi wrth ei weithio allan drostynt eu hunain, gan ddefnyddio'r cysyniad o ddysgu anwythol (neu 'darganfodol'), a eiriolir yn glir gan y dull lluniadaethol o ddysgu. Ceir gwrthgyferbyniad rhwng y dull hwn a dysgu diddwythol (neu 'dan gyfarwyddyd tiwtoriaid'), fel pan ddarperir rheolau strwythurol i ddysgwyr. Mae integreiddio adnoddau dysgu i mewn i'r corpws, ac adeiladu'r corpws yn ôl anghenion dysgwyr, yn ei gwneud yn bosibl i ddysgwyr gyrchu data corpws perthnasol. Mae'r pedwar ymarfer addysgegol sydd wedi'u hymgorffori yn CorCenCC yn hwyluso'r broses o ganolbwyntio ar ffurf, y'i cydnabyddir fel elfen allweddol o ddysgu iaith yn effeithiol.

O gofio mai CorCenCC yw'r cyntaf o'i fath i dynnu ar amrywiaeth eang o genres ieithyddol, ffurfiau ieithyddol a ffyrdd ieithyddol o gynrychioli'r Gymraeg, roedd galluogi dysgwyr ac athrawon/tiwtoriaid i hidlo data'r corpws yn ôl ei fath a'i swm wrth gynhyrchu'r allbynnau a geir o ganlyniad yn drydedd nodwedd arloesol a ymgorfforwyd yn Y Tiwtiadur. Fel bod modd cymhwyso'r cyfleuster dysgu i bob oedran, un o agweddau allweddol y drydedd nodwedd arloesol hon yw'r gallu i hidlo darnau o destun sy'n debygol o gynnwys cynnwys amhriodol (fel rhegfeydd) fel nad ydynt yn ymddangos. Un o'r nodweddion y gellir ei gymhwyso i ddysgwyr o bob oedran yw'r gallu i wahaniaethu rhwng enghreifftiau o iaith o fathau gwahanol o ddata wrth roi ystyriaeth benodol i strwythurau cymhleth y mae modd iddynt amrywio rhwng siaradwyr (e.e. treiglo'r Gymraeg, neu rywfaint o ffurfiau enw lluosog, mewn darnau o destun sy'n amrywio o safbwynt eu ffurfioldeb). Yn y modd hwn, gall athrawon a dysgwyr reoli'r posibilrwydd am ddryswch mewn achosion o ddefnyddio mwy nag un ffurf.

## 6.3. WP4: Cyflawniadau

Roedd gwaith datblygu Y Tiwtiadur wedi'i lywio gan ddefnyddwyr o'r cychwyn. Ymgynghorwyd â thiwtoriaid/athrawon a dysgwyr, a oedd yn cynrychioli amrediad eang o lefelau cymhwysedd yn y Gymraeg ar draws y sectorau addysg ieithyddol gwahanol, ar amserau gwahanol er mwyn sicrhau y gellid dylunio, profi a gwella Y Tiwtiadur mewn modd ailadroddus. Dilynwyd nodau ac amcanion WP4 ar draws tri phrif gyfnod: (i) cyfnod ymgynghori, (ii) cyfnod datblygu cynnyrch, a (iii) cyfnod arddangos.

*(i) Y cyfnod ymgynghori*

Yn ystod y cyfnod ymgynghori, cafodd holiadur ei beilota gydag ychydig o ymarferwyr ym maes addysgu'r Gymraeg er mwyn archwilio pa adnoddau yr oedd dysgwyr ac athrawon yn eu defnyddio yn barod a pha rai y byddent am eu gweld yn cael eu datblygu yn ddelfrydol fel rhan o Y Tiwtiadur. Datblygwyd holiadur mwy manwl, a oedd wedi'i dargedu ar gyfer cynulleidfa fwy o faint, o ganlyniad i'w hadborth. Cafodd hwn ei ddosbarthu i athrawon a thiwtoriaid y Gymraeg yn ystod dwy gynhadledd genedlaethol, a chafodd ei rannu ar ffurf ar-lein yn ddiweddarach. Cafodd cyfanswm o 44 holiadur eu dychwelyd gan athrawon, hyfforddwyr, darlithwyr a thiwtoriaid y Gymraeg o amrediad eang o gyd-destunau – o'r bobl hynny sy'n addysgu mewn ysgolion cynradd (cyfrwng Cymraeg a chyfrwng Saesneg) i'r rheiny sy'n addysgu oedolion – a chynhaliwyd deg grŵp ffocws, gan gasglu safbwyntiau 55 o athrawon/tiwtoriaid ac 14 o ddysgwyr Cymraeg i Oedolion.

Yn ogystal â'r holiadur, gwnaethom gyfarfod wyneb yn wyneb ag athrawon a thiwtoriaid, er mwyn codi ymwybyddiaeth o'r corpws ac Y Tiwtiadur, ac i barhau i gasglu safbwyntiau y byddai'n helpu i lywio gwaith datblygu'r pecyn cymorth. Gwnaeth ymatebion yr holiadur, ynghyd â chyfarfodydd grŵp ffocws dilynol, ein helpu i nodi blaenoriaethau ar gyfer Y Tiwtiadur. Roedd y trafodaethau a gynhaliwyd gennym yn hynod ddefnyddiol, a gwnaeth y broses o gyfnewid syniadau a ddigwyddodd dros ystod y grwpiau ffocws, a thrwy'r adborth a gafwyd drwy'r holiaduron, wella'n dull o feddwl ynghylch sut i lywio'r gwaith wrth i ni symud ymlaen.

*(ii) Y cyfnod datblygu cynnyrch a (iii) y cyfnod arddangos*

Yn dilyn yr ymgynghoriad, cydweithredodd timau WP4 a WP5 i ddatblygu prototeipiau o'r offeryn. Yna, cafodd y gwaith hwn ei ddatblygu ymhellach gan J. Davies (a gafodd ei oruchwylio gan Teahan) mewn cydweithrediad ag Anthony. Cafwyd cyfle i arddangos y gwaith yr oedd wedi'i gwblhau ar Y Tiwtiadur hyd at y pwynt hwnnw yn ystod cynhadledd flynyddol y Ganolfan Dysgu Cymraeg Genedlaethol yn 2019, a chafwyd adborth cefnogol ac adeiladol gan y tiwtoriaid a fynychodd y gweithdai, gan gynnwys llawer o syniadau defnyddiol mewn perthynas â'i ddefnyddioldeb. Llywiodd yr adborth a gafwyd ganddynt weddill y gwaith datblygu. Roedd mewnwelediadau cadarnhaol yn cynnwys sut y gellid defnyddio Y Tiwtiadur gyda dysgwyr. Daeth tair prif thema i'r amlwg:

*1. Cefnogi dealltwriaeth o safbwynt treiglo ar lefel frawddegol yn erbyn lefel eiriadurol*

Un o nodweddion heriol y Gymraeg (a'r ieithoedd Celtaidd eraill) yw treiglo – proses forffoffonolegol lle mae newid ffonolegol yn cael ei achosi mewn set gaeedig o gytseiniaid cyntaf geiriau pan fo geiriau penodol yn ymddangos mewn cyd-destunau cystrawennol penodol (Ball and Müller, 1992; Thomas and Mayr, 2010). Er enghraifft, mae sain y llythyren gyntaf 'c' /k/ yn ***cath*** /kɑθ/ yn mynd trwy broses o feddaliad lle caiff /k/ ei meddalu i /g/ – /gɑθ/. Mae'r newid hwn yn digwydd ar ôl y fannod, *y*, ac ar ôl y rhif *dau* (gwrywaidd) / *dwy* (benywaidd), er enghraifft. Adlewyrchir y newid ffonolegol hwn yn y ffurf ysgrifenedig, lle ysgrifennir *cath* fel *gath*. Mewn rhai achosion, gellir trawsnewid cytsain gyntaf geiriau mewn tair ffordd, gan ddibynnu ar y cyd-destun (e.e. *cath* /kɑθ/ (ffurf waelodol), *gath* /gɑθ/ (treiglad meddal), *nghath* /ŋ̊ɑθ/ (treiglad trwynol) a *chath* /χɑθ/ (treiglad llaes). Gall y ffurfiau newidiol hyn beri problemau wrth ddarllen ac ysgrifennu, a gallant gael effaith ar ddatblygu/dysgu o safbwynt geirfa a llythrennedd, oherwydd eu bod yn ei gwneud yn anodd adnabod geiriau ac oherwydd bod y rheolau ar gyfer y treigladau yn gymhleth weithiau. Gall enghreifftiau a yrrir gan gorpysau dynnu sylw i ffurf a helpu dysgwyr i nodi lle, pryd a sut mae geiriau'n newid ar draws cyd-destunau a faint o amrywiaeth a geir o safbwynt mynegi ffurfiau 'targed'.

*2. Llenwi bylchau mewn cymorth geiriadurol*

Mewn perthynas uniongyrchol â threiglo, yn ogystal â ffurfiau newidiol eraill, mae llawer o ddysgwyr yn methu â dod o hyd i eiriau mewn geiriaduron papur a geiriaduron ar-lein oherwydd eu bod yn chwilio am y ffurf dreigledig (e.e. ***g**ath*) ac nid y ffurf waelodol (h.y. ***c**ath*), neu oherwydd eu bod wedi camsillafu neu gam-gynrychioli'r gair mewn testun. Bydd y corpws yn galluogi defnyddwyr i ddarganfod sut y mae ffurf sydd o ddiddordeb iddynt yn cael ei defnyddio fel arfer mewn mathau gwahanol o genres a chyfryngau.

*3. Cefnogi dysgwyr wrth iddynt werthuso'u gwaith ysgrifenedig*

Un o nodweddion allweddol Y Tiwtiadur yw'r opsiwn sydd ar gael i ddysgwyr a/neu athrawon/tiwtoriaid i ddefnyddio'r feddalwedd i godio darn o destun o'u dewis o ran pa mor anodd ydyw yn nhermau amlder y ffurfiau a ddefnyddir. Gall hyn fod yn ddefnyddiol wrth benderfynu pa mor addas yw darn o destun ar gyfer dysgwr neu grŵp o ddysgwyr. Yn ogystal, gall dysgwyr broffilio'u gwaith ysgrifenedig eu hunain, er mwyn gwerthuso pa mor eang yw eu gwybodaeth am eirfa a'u defnydd ohoni.

Yn ogystal â'r mewnwelediadau hyn o ran sut y gellid rhoi'r offerynnau hyn ar waith, cafwyd myfyriadau gwerthfawr gan randdeiliaid ar eu pryderon ynghylch defnyddio adnodd o'r fath yn y dosbarth. Codwyd pum prif broblem, fel a amglygir isod:

1. *Modelu iaith 'anghywir'*

Un o'r pryderon a gododd sawl tro ymhlith athrawon a thiwtoriaid o bob cyd-destun addysgol oedd y gallai data'r corpws, oni iddo gael ei 'gywiro', fodelu'r union ffurfiau ac ymadroddion y mae athrawon yn ceisio cael gwared ohonynt o ymdrechion eu dysgwyr. Gan nad yw corpws yn gwahaniaethu'n benodol rhwng yr hyn a ystyrir yn gywir neu'n anghywir ar lefel ragnodol (ar wahân i'r hyn a wneir yn ôl amlder neu ymddangosiad geiriau), cynghorwyd athrawon i ymgyfarwyddo â'r darnau o destun y byddent yn eu defnyddio yn ystod sesiynau ymlaen llaw, ac i nodi unrhyw beth y byddent am gynghori'r dysgwyr yn ei gylch. Gwnaeth y dull hwn gydnabod bod y berthynas rhwng dulliau disgrifiadol a rhagnodol o addysgu iaith yn gymhleth. Roedd yn cydnabod bod dyletswydd ar athrawon i hysbysu dysgwyr am y ffurfiau y bydd pobl eraill yn eu hystyried yn rhai 'anghywir' (gan fod ymwybyddiaeth o'r fath yn un o'r agweddau sydd ynghlwm wrth wybodaeth am yr iaith). Ar yr un pryd, byddai'r dull â llaw yn annog athrawon i gwestiynu eu credoau eu hunain ynghylch yr hyn sy'n 'dderbyniol' fel ffurf darged, neu beidio, o gofio'r defnydd a dystir.

2. *Iaith sarhaus*
Cododd llawer o athrawon (o'r sector addysg gynradd yn enwedig) bryderon am y posibilrwydd o gyrchu cynnwys amhriodol ar ddamwain, gan ganolbwyntio'n bennaf ar gynnwys a oedd yn ymwneud â geiriau amhriodol neu sarhaus (fel rhegfeydd). Er bod gwaith codio ar gyfer lefelau a mathau o fod yn sarhaus y tu hwnt i gwmpas y prosiect presennol (na fyddai, mewn rhai achosion, i'w cael ar lefel gair unigol bob amser), tagwyd y corpws am eiriau rhegi a chynnwys o natur sensitif benodol ar y lefel destun. Er bod y cyfrifiadau ystadegol y mae'r offerynnau addysgeg yn seiliedig arnynt yn cyfeirio at y 10 miliwn o eiriau sydd wedi'u cynnwys yn CorCenCC, mae pedwar ymarfer y pecyn cymorth pedagogaidd yn hidlo'r darnau testun hyn fel nad ydynt yn ymddangos.

3. *Cymhlethdod y rhyngwyneb*
Un o nodau clir CorCenCC oedd datblygu corpws a fyddai'n hwylus i ddefnyddwyr ac y byddai'n gallu cael ei drosglwyddo i faes addysg yn hawdd. Er mwyn i hynny weithio, roedd yn rhaid i'r rhaglen fod yn addas i'r diben ac yn hwylus ei ddefnyddio. Cododd yr ymdeimlad hwn ymhlith rhai o'r tiwtoriaid a'r athrawon y gwnaethom gwrdd â nhw. Helpodd y safbwyntiau hyn i sicrhau bod rhyngwyneb syml a chlir yn perthyn i'r pedwar ymarfer addysgegol a oedd yn reddfol o safbwynt y defnyddiwr. Ar yr un pryd, roedd y rhyngwyneb hwn yn adlewyrchu'r rhyngwyneb a ddefnyddir gan CorCenCC er mwyn sicrhau cydlyniant ar draws y ddau ac i ennyn hyder mewn athrawon o safbwynt dilyniant o ddefnyddio'r offerynnau addysgeg i ddefnyddio'r corpws yn ehangach pe byddent am wneud hynny.

*4. Hygyrchedd*
Er bod gwaith datblygu platfformau amgen ar gyfer cynnal y corpws y tu hwnt i'r hyn sydd dan sylw yn yr astudiaeth bresennol, roedd yn glir, mewn ysgolion yn enwedig, lle ceir mynediad cyfyngedig i gyfrifiaduron yn aml, y byddai ap y gellid ei lawrlwytho ar ffonau yn ddefnyddiol. Mae ysgolion yn defnyddio apiau sydd eisoes yn bodoli ar gyfer y Gymraeg yn barod, fel Duolingo, ac maent yn gweld bod y platfform hwnnw'n ddefnyddiol, felly gofynnodd athrawon a fyddai'n bosibl datblygu platfform tebyg ar gyfer y prosiect presennol. Ar y cyfan, roedd yn amlwg y bydd datblygu ap ar gyfer ei ddefnyddio yn y dosbarth yn addasiad gwerthfawr o Y Tiwtiadur yn y dyfodol. Gan nad oedd hwn yn un o nodau'r prosiect presennol, ac y byddai'n cymryd gwaith ymchwil ac adnoddau sylweddol i'w gwblhau, mae wedi'i nodi fel ystyriaeth bwysig ar gyfer gwaith dilynol.

*5. Nifer brawychus yr enghreifftiau yn yr allbwn*
Mae CorCenCC yn gorpws o dros 11 miliwn o eiriau y mae modd ei chwilio. Mae hyn yn golygu y bydd rhai chwiliadau'n arwain at nifer anferth o allbynnau. I'r bobl hynny nad oes ganddynt brofiad o ddefnyddio corpysau a'r mathau o allbynnau a gynhyrchir ganddynt, gall swm yr allbynnau a gynhyrchir fod yn llethol, a gallai atal dysgwyr ac athrawon/tiwtoriaid. Am y rheswm hwnnw, un o elfennau arloesol Y Tiwtiadur yw ei fod yn rhoi'r gallu i athrawon, tiwtoriaid a dysgwyr reoli swm yr allbwn a gynhyrchir gan ymholiadau (fel a eiriolir gan y dull dysgu a yrrir gan ddata). Yn ogystal â'r gallu i ddethol testunau neu gategorïau semantig wrth archwilio darnau o destun, mae'r offeryn yn rhoi'r gallu i athrawon/tiwtoriaid a dysgwyr gyfyngu ar hyd y darnau o destun wrth orchuddio geiriau yn yr ymarfer Llenwi Bylchau, ac yn gosod uchafswm o 20 o achosion ar yr allbwn a gynhyrchir ar gyfer yr ymarferion Nodi Geiriau a Geiriau yn eu Cyd-destun ac ati. Gyda'i gilydd, mae'r nodweddion hyn yn cynyddu lefel annibyniaeth y dysgwr a'r athro/tiwtor fel eu bod yn gallu sicrhau bod y pecyn cymorth yn gweithio iddynt.

## 6.4. WP4: Cyfraniadau allweddol

Drwy'r gwaith o ddatblygu rhyngwyneb addysgegol, a arweinir gan y cysyniad o ddysgu ac asesu a yrrir gan ddata, mae WP4 wedi cyfrannu (i) adnodd addysgegol newydd sydd (ii) wedi'i dynnu o gorpws ar-lein o Gymraeg cyfoes, o fath (iii) nad yw wedi bodoli erioed o'r blaen ar gyfer addysgu'r Gymraeg, a (iv) y gall fod yn fodel ar gyfer gwaith tebyg o safbwynt ieithoedd lleiafrifol eraill. Mae wedi gwneud cyfraniad gwerthfawr i faes dysgu ac addysgu ieithoedd a, gan ei fod o natur ffynhonnell agored, mae ar gael i gefnogi dysgwyr fel rhan o'u dysgu anwythol, ni waeth eu hoedran, eu lefel gallu a'u lleoliad daearyddol. Mae'r adnodd yn cynnig cyfle newydd ac unigryw ar gyfer ysgolion yng Nghymru i ymgymryd â'r cysyniad o ddysgu a yrrir gan ddata ac i ymgysylltu â'u dulliau addysgeg eu hunain a arweinir gan y corpws, a'u datblygu. Yn ogystal â hyn, unwaith y bydd athrawon a dysgwyr wedi'u cyflwyno i'r corpws drwy Y Tiwtiadur, gallent hefyd deimlo'n ddigon hyderus i archwilio'r corpws mewn ffyrdd eraill drwy ddefnyddio prif offerynnau ymholi CorCenCC.

Cafwyd galwadau amrywiol dros y blynyddoedd diwethaf am gorpws sy'n gallu llywio'r dull o ddarparu'r Gymraeg (NFER, 2008: 48; Llywodraeth Cymru 2013: 27, 71; Mac Giolla Chríost et al., 2012). Fel corpws cyfoes o'r Gymraeg sy'n cynnwys pecyn cymorth pedagogaidd integredig (Y Tiwtiadur), mae CorCenCC yn diwallu'r angen hwnnw, gan lywio gwaith ysgrifenedig y cwricwlwm, gwaith asesu iaith ac adnoddau dysgu iaith yn yr un modd ag y mae corpysau tebyg yn ei wneud ar gyfer y Saesneg (e.e. mae'r Cambridge English Corpus (CEC) yn llywio adnoddau addysgu Cambridge English Language; mae'r British National Corpus (BNC) yn llywio adnoddau Pearson Longman). Gallai datblygu set gyfatebol lawn ac annibynnol o adnoddau addysgu'r Gymraeg, sy'n seiliedig ar CorCenCC, fod yn drywydd posibl i'r gwaith hwn yn y dyfodol.

Yn unol â'r disgwyliadau a bennir yn y *Cwricwlwm i Gymru: 2022* newydd, bydd adnodd o'r fath yn helpu i ddatblygu galluoedd ymwybyddiaeth ieithyddol dysgwyr a gloywi eu sgiliau Cymraeg mewn ffordd naturiolaidd, a fydd, yn y pen draw, yn cael effaith ar agenda *Cymraeg 2050: Miliwn o Siaradwyr* Llywodraeth Cymru (Llywodraeth Cymru, 2017), sydd â'r uchelgais o sicrhau miliwn o siaradwyr Cymraeg erbyn 2050.

## 6.5. WP4: Cymwysiadau ac effaith

Mae defnyddio data o gorpysau i gefnogi dysgu ieithoedd mewn ysgolion yn arfer sy'n datblygu'n gyflym, ond nid yw'r dull o'i weithredu'n effeithiol wedi'i ddatblygu'n llawn a phrin yw'r gwaith ymchwil i'w effeithiolrwydd. Yn y Cwricwlwm i Gymru: 2022 newydd, bydd yn ofynnol i blant ddysgu am gysyniadau ieithoedd, dadansoddi mân-wahaniaethau ieithyddol, a deall sut maent yn wahanol mewn ieithoedd gwahanol. Mae data o gorpysau wedi'i alinio'n berffaith â'r gwaith o hyrwyddo sgiliau a gwybodaeth feta-ieithyddol, yn enwedig mewn cyd-destun dwyieithog o'r math a geir yng Nghymru. Cam pwysig nesaf fydd rhannu'r adnodd ar-lein, rhad ac am ddim hwn i ysgolion yng Nghymru. Bydd hyn yn golygu gweithio'n agos â CBAC, dylunwyr cwricwlwm a Llywodraeth Cymru wrth nodi'r arfer gorau ar gyfer defnyddio CorCenCC mewn cyd-destunau dysgu ac addysgu ledled Cymru, a modelu'r dull o'i weithredu mewn cyd-destunau iaith leiafrifol eraill mewn lleoedd eraill. Gallai'r gwaith hwn arwain at brosiectau ymchwil sydd wedi'u hariannu ar gyfer gwerthuso effeithiolrwydd cymwysiadau amrywiol CorCenCC ac Y Tiwtiadur gyda mathau gwahanol o ddysgwyr gyda'r bwriad o ymestyn ei ymarferoldeb lle bo hynny'n briodol. Er enghraifft, gellid addasu ymarferion ar gyfer ap ffôn neu blatfformau technolegol eraill, gan gynyddu defnydd yr adnodd a chynyddu ei effaith ar faes addysg.

# 7. Pecyn Gwaith 5: Adeiladu'r seilwaith i letya CorCenCC

## 7.1. WP5: Disgrifiad

Roedd WP5 yn ymwneud â'r elfennau technegol o adeiladu CorCenCC, gan greu offerynnau i gefnogi pob cam o'r broses o adeiladu'r corpws, o gasglu data (gan ganolbwyntio ar yr ap cyfrannu torfol), trwy'r cam o goladu (drwy ddefnyddio'r offerynnau rheoli data), i'r gwaith ymholi a dadansoddi (y rhyngwyneb ar y we).

Spasić a arweiniodd WP5, gan weithio gyda Knight, Rayson a Piao, a Neale a Muralidaran, sef dau o gynorthwywyr ymchwil y prosiect. Darparodd Anthony (ieithydd corpws, arbenigwr technoleg addysgol a chrëwr Antconc), Scannell (gwyddonydd cyfrifiadurol ag arbenigedd mewn prosesu iaith naturiol (NLP), cyfieithu peirianyddol ac ieithoedd lleiafrifol) a Donnelly (ieithydd cyfrifiadurol a weithiodd yn agos ar ddatblygu corpysau Cymraeg blaenorol) arbenigedd technegol ac ymgynghorol ychwanegol.

## 7.2. WP5: Amcanion

Bwriad WP5 oedd datblygu seilwaith cyfrifiadurol i gefnogi'r gwaith o gasglu a storio'r swm mawr hwn o destun a data dadansoddol mewn modd systematig, ynghyd â rhyngwyneb hwylus ei ddefnyddio er mwyn ei gwneud yn bosibl rhyngweithio â'r data hyn ar-lein. Un o elfennau pwysig y gwaith hwn oedd cynllunio ac adeiladu system ystorio a fyddai'n ei gwneud yn bosibl ychwanegu data newydd i'r corpws dros amser, fel y gallai defnyddwyr gefnogi'r gwaith o gynnal y corpws ac fel y byddai cyfraniadau i'r corpws yn fenter gymdeithasol. Cafodd cyfres o offerynnau dadansoddi'r corpws ei datblygu ar ben yr ystorfa er mwyn cefnogi swyddogaethau sydd fel arfer wedi'u hintegreiddio i mewn i gorpysau cyfoes, fel peiriannau cydgordio ac offerynnau cydleoli Gair Allweddol mewn Cyd-destun (KWIC), offerynnau chwilio a threfnu, rhestrau amlder geiriau, peiriannau dadansoddi geiriau allweddol a chyfleusterau profi ystadegol.

Mae Ffigur 2 yn dangos y llif gwaith ar gyfer casglu data. Amlygir y gwahaniaeth rhwng y tri phrif fath o iaith (llafar, ysgrifenedig ac electronig (e)), gan fod angen cwblhau dulliau prosesu gwahanol arnynt cyn y gallai'r data gael ei integreiddio i mewn i'r corpws.

Fel y mae Ffigur 2 yn dangos, cafodd yr holl wybodaeth berthnasol am gyfranogwyr a metadata disgrifiadol ei gofnodi ar yr un pryd ag y cafodd y data ei gasglu. Roedd cael caniatadau i rannu'r data mewn adnodd cyhoeddus ar-lein yn hanfodol i waith datblygu CorCenCC. Cafwyd y caniatadau hyn gan yr endidau cyfreithiol perthnasol (e.e. perchennog yr hawlfraint; y siaradwr ei hun) cyn y cafodd y data ei gasglu a'i storio'n lleol.

*Ffigur 2.* Llif gwaith ar gyfer casglu data yn CorCenCC

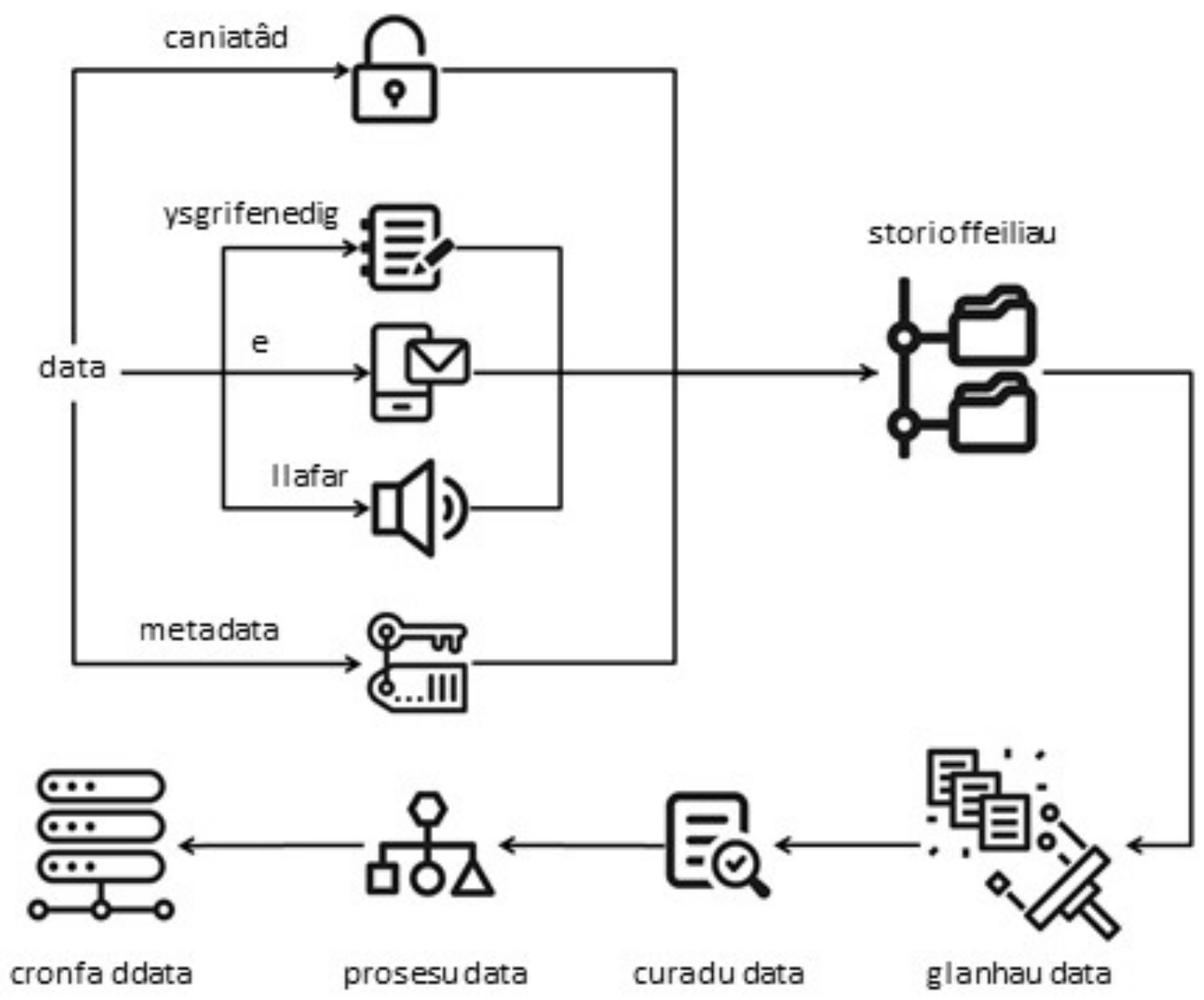

Cafodd y data crai ynghyd â'r caniatadau a metadata cyfatebol eu hadneuo mewn system storio ffeiliau leol. Yn dilyn hynny, cafodd fformatau data gwahanol eu safoni'n destun plaen. Mae'n bosib prosesu testun plaen yn awtomatig gan ddefnyddio offerynnau prosesu iaith naturiol (NLP), pe byddai angen ychwanegu haen arall o fetadata ieithyddol yn ddiweddarach; mae hyn yn helpu i ddiogelu'r corpws at y dyfodol, drwy ei gwneud yn bosibl ychwanegu gwybodaeth ychwanegol.

## 7.3. WP5: Cyflawniadau

Un o nodweddion arloesol allweddol prosiect CorCenCC oedd ailddiffinio'r ffyrdd o gynllunio ac adeiladu corpysau ieithyddol, gan alinio'r dulliau ag oes Gwe 2.0 mewn modd mwy cryno. I'r perwyl hwn, cymerwyd camau i adeiladu a gwerthuso system sy'n ei gwneud yn bosibl casglu data llafar gan siaradwyr drwy gyfrannu torfol. Mae cyfrannu torfol yn ffordd o gasglu adnoddau (enghreifftiau ieithyddol yn yr achos hwn) gan y cyhoedd, drwy ofyn i wirfoddolwyr gymryd rhan. Gwnaethom hwyluso cyfrannu torfol drwy ddefnyddio ap y mae modd ei gynnal ar unrhyw ddyfais sydd â chysylltiad â'r we; gweithiodd ar ddwy ffurf: fel cymhwysiad ffôn ac fel gwefan ryngweithiol. Er mwyn cynyddu'r sylfaen o ddefnyddwyr posibl, gweithredwyd fersiwn symudol yr ap ar blatfformau iOS (h.y. Apple) ac Android. Roedd y cymhwysiad yn sicrhau bod y profiad o gyfrannu at y corpws yn un personol iawn, gan roi perchenogaeth i ddefnyddwyr dros eu recordiadau eu hunain, a rheolaeth drostynt.

Cafodd yr holl ddata crai (yn llafar, yn ysgrifenedig ac yn electronig) ei storio mewn modd systematig o fewn strwythur ffolder a oedd wedi'i ddiffinio ymlaen llaw, ac a gyfatebodd â'r fframwaith samplu. O hynny ymlaen, aeth y data trwy'r prosesau glanhau a churadu perthnasol. I gefnogi mynediad cydweithredol ar gyfer defnyddwyr lluosog gan aelodau'r tîm (o ymchwilwyr i drawsgrifwyr) ar draws safleoedd gwahanol (ar draws ac o fewn y sefydliadau lluosog a oedd ynghlwm wrth y prosiect), datblygwyd offeryn rheoli data ar-lein ar ben y system storio ffeiliau. Darparodd ryngwyneb defnyddiwr graffigol (GUI) a hwylusodd y broses o lanlwytho data crai, mynegeio'r metadata cyfatebol a chofnodi trawsnewidiadau data dilynol, gan sicrhau y gallai pob ymchwilydd fonitro cynnydd holl elfennau'r broses o adeiladu'r corpws yn fanwl.

Unwaith yr oedd y darnau o destun wedi'u trosi i fformat testun plaen, cawsant eu nodi â haenau o fetadata sosioieithyddol (e.e. ffynhonnell, genre, tarddiad daearyddol) a fyddai'n cael eu defnyddio i ymholi'r data, a chawsant eu tagio'n awtomatig. Fel a ddisgrifiwyd yn Adran 4, datblygodd tîm WP2 CorCenCC CyTag (Neale et al., 2018) er mwyn cyflawni'r broses o bennu tagiau. Mae CyTag yn gyfres o offerynnau prosesu iaith naturiol ar lefel wyneb ar gyfer y Gymraeg sy'n seiliedig ar gysyniad gramadeg cyfyngiadol (Karlsson, 1990, Karlsson et al., 1995). Mae'n cefnogi'r syniad o segmentu testun, gan gynnwys hollti brawddegau a thocyneiddio, yn ogystal â phennu tagiau i rannau ymadrodd a'u lemateiddio. Mae'n cynnig datrysiad pwrpasol ar gyfer rhagbrosesu'r Gymraeg mewn modd ieithyddol sylfaenol, gan gynnwys set dagiau sydd â chyfoeth digonol o ddata ynddi i gipio mympwyon yr iaith – a geir ar ei ffurfiau llafar yn enwedig. I hwyluso'r broses o gyflawni dadansoddiad semantig ar ddata am y Gymraeg ar raddfa fawr, cafodd yr holl ddata a ragbroseswyd ei ddynodi ymhellach yn ôl categorïau semantig gan ddefnyddio'r CySemTagger a ddatblygwyd yn WP3 (gweler Adran 5).

Cafodd y corpws ei storio a'i reoli mewn cronfa ddata berthynol lle'r oedd yn bosibl cyrchu data mewn modd diogel gan ddefnyddwyr lluosog ac ar yr un pryd â'i gilydd. Er mwyn rhannu'r data ar-lein, gwnaethom ddatblygu rhyngwyneb ar y we ar gyfer y gronfa ddata. Y prif reswm dros ddatblygu rhyngwyneb pwrpasol yn hytrach nag ailddefnyddio datrysiad a oedd eisoes yn bodoli, fel CQPweb (Hardie, 2012) oedd y gofyniad i deilwra ei ymarferoldeb i fetadata penodol CorCenCC a'i ddarpar ddefnyddwyr. Er mwyn casglu gwybodaeth am ofynion defnyddwyr, gwnaethom ddefnyddio'r cyfryngau cymdeithasol i gynnal arolwg o ddefnyddwyr corpysau ar y pryd. Cafwyd ymatebion gan gyfanswm o 62 o unigolion, a gwnaeth eu mewnbwn nodi'r gofynion allweddol o safbwynt ymarferoldeb.

Un o'r ystyriaethau pwysig ar gyfer datblygu seilwaith y corpws oedd gwirio'i fod yn addas i'r pwrpas. Gwerthusodd grŵp o ieithyddion corpws ddefnyddioldeb ac ymarferoldeb y rhyngwyneb ar y we. Roedd y broses werthuso hon yn cynnwys cyfuniad o holiaduron ac ymarferion siarad yn uchel. Yn gyffredinol, roedd y cyfranogwyr yn meddwl bod y system yn ddefnyddiol yn nhermau diwallu eu hanghenion gwybodaeth o fewn cwmpas eu gweithgareddau proffesiynol. Roedd yr ymarferoldeb yn hawdd ei ddeall heb orfod cyfeirio at y sgrin gymorth. Cytunodd pob cyfranogwr y byddent yn debygol o fabwysiadu'r system a'i hargymell i ieithyddion eraill.

## 7.4. WP5: Cyfraniadau allweddol

Mae creu seilwaith corpws newydd yn golygu gwaith sylweddol. Mae'r mwyafrif o ymchwilwyr corpws yn defnyddio meddalwedd sydd eisoes yn bodoli i ddadansoddi'r darnau o destun y maent yn eu casglu. Fodd bynnag, yn yr achos hwn, nid oedd y fath hon o feddalwedd eisoes yn bodoli, ac felly roedd angen ei hadeiladu cyn y gallai'r darnau o destun roeddem wedi'u casglu gael eu mynegeio'n briodol er mwyn eu mewnbynnu i'r corpws. Felly, roeddem yn mynd i'r afael â sawl her sylweddol ar y cyd a oedd yn bwysig ar gyfer datblygu gwaith ymchwil corpysau ieithyddol .

Yn ail, roedd yn rhaid i ni ddylunio llawer o'r seilwaith cyfrifiadurol gwaelodol ar gyfer pennu tagiau i'r iaith, a'i dadansoddi, gan ddechrau o'r dechrau, gan gofio bod llawer o nodweddion yn perthyn i'r Gymraeg, gan gynnwys gwahaniaethau gramadegol, nad yw'n hawdd eu trosglwyddo o ieithoedd sydd ag adnodd corpws sylweddol sydd eisoes yn bodoli (yn enwedig Saesneg), ac amrywiaeth ranbarthol a chyweiriol sylweddol sy'n deillio o hanes cymdeithasol arbennig yr iaith. Mae pileri allweddol y seilwaith yn cynnwys fframwaith sy'n cefnogi'r broses o gasglu metadata, ap symudol arloesol sydd wedi'i ddylunio ar gyfer casglu data llafar (gan ddefnyddio dull cyfrannu torfol), cronfa ddata o'r ochr gefn sy'n storio data curedig, a rhyngwyneb ar y we sy'n galluogi defnyddwyr i ymholi'r data ar-lein. Drwy ddefnyddio tagiau Cymraeg, rydym wedi sicrhau nad yw'r corpws yn cael ei ganfod yn offeryn allanol (Saesneg) sydd wedi'i arosod ar y Gymraeg, ac na fydd yn bosib ei ganfod felly, ond yn hytrach ei fod yn perthyn i Gymru a'r Gymraeg. Yn y modd hwn, anogir defnyddwyr i lwyr gefnogi'r iaith, nid yn unig fel ffynhonnell wybodaeth ond hefyd fel y cyfrwng y gellir ei hastudio drwyddo. Ar yr un pryd, bydd y ffaith fod rhyngwyneb Saesneg ychwanegol ar gael yn sicrhau pwynt mynediad ar gyfer y nifer fawr o bobl hynny sydd â diddordeb yn y Gymraeg sy'n fwy na'u hyfedredd ynddi, gan gynnwys y miloedd o ddysgwyr Gymraeg.

Yn drydydd, rydym wedi creu offerynnau sydd ar gael yn rhad ac am ddim er mwyn i bobl eraill eu haddasu wrth greu eu corpysau eu hunain. Rydym yn ymrwymedig yn enwedig i gefnogi'r gwaith o adeiladu corpysau ar gyfer ieithoedd lleiafrifol eraill, ac mae ein model a yrrir gan ddefnyddwyr yn llywio prosiectau o'r fath yn uniongyrchol trwy ddarparu templed ar gyfer datblygu corpysau mewn unrhyw iaith arall.

Yn bedwerydd, drwy gasglu tîm rhyngwladol o arbenigwyr ynghyd, rydym wedi gallu rhoi'r datblygiadau arloesol technolegol diweddaraf ar waith, a datblygu ein syniadau newydd ein hunain, gan fwrw golau ar drywydd ar gyfer gwaith yn y dyfodol yn oes Gwe 2.0. Mae'r ap cyfrannu torfol yn un o'r cyntaf o'i fath i gael ei ddefnyddio ar gyfer adeiladu corpws cytbwys o ddata iaith naturiol trwy ategu dulliau mwy traddodiadol o gasglu data, ac sydd wedi mynd i'r afael yn llwyddiannus â phroblem sylweddol a chyson o safbwynt casglu data llafar o ansawdd uchel â chaniatâd. Yn ogystal, rydym wedi dangos ei bod yn bosibl darganfod yr adnoddau dynol angenrheidiol ar gyfer trawsgrifio data o'r fath ac ar gyfer cwblhau'r gwaith angenrheidiol o bennu tagiau â llaw, hyd yn oed ar gyfer iaith sydd â charfan gymharol fach o siaradwyr rhugl.

### 7.5. WP5: Cymwysiadau ac effaith

Er y datblygwyd y seilwaith cyfrifiadurol ar gyfer casglu gwybodaeth am y Gymraeg, mae'n bosib i'w gynllun gael ei ailddefnyddio i gefnogi gwaith datblygu corpysau yng nghyd-destunau ieithoedd lleiafrifol neu brif ieithoedd eraill, ac mae hynny'n ehangu defnyddioldeb ac effaith y gwaith hwn.

## 8. Crynodeb o gymwysiadau posibl a'u heffaith

Fel y disgrifiwyd yn gynharach yn y ddogfen hon, mae gan bob rhan o'r prosiect (fel y nodweddir gan y pecynnau gwaith) gymwysiadau gwerthfawr sy'n cynnig manteision cymdeithasol, economaidd a/neu academaidd. Ar lefel gymdeithasol, mae'r corpws yn cynnig y cyfle i ddeall y Gymraeg fel iaith fyw sy'n cael ei defnyddio. Mewn termau economaidd, mae'r corpws yn cynnig lle i ddatblygu adnoddau newydd gwerthfawr ar gyfer dysgwyr a defnyddwyr Cymraeg, gan gynnwys y posibilrwydd o greu geiriadur sy'n seiliedig ar y corpws ac amrediad o offerynnau technolegol sydd wedi'u llywio gan ddata a allai gynnwys apiau dysgu iaith, cynhyrchu testun rhagfynegol, offerynnau prosesu geiriau, cyfieithu peirianyddol, ac offerynnau adnabod llais a chwilio'r we. Yn anuniongyrchol, bydd cefnogaeth y corpws ar gyfer sicrhau'r canlyniadau cymdeithasol ac economaidd hyn yn hyrwyddo cydnabyddiaeth y Gymraeg fel elfen sylweddol o dirwedd ieithyddol y DU a'r byd. Mae'r cymwysiadau academaidd posibl yn eang ac yn amrywiol, fel a ganlyn:

- Ieithyddiaeth corpws: Bydd cynllun arloesol CorCenCC yn llywio ac yn arwain y gwaith o ddatblygu corpysau yn y dyfodol a'u defnydd, mewn unrhyw iaith, drwy ddarparu adnoddau ffynhonnell agored a'r protocolau ar gyfer dulliau torfol o gasglu a dadansoddi data ac ar gyfer integreiddio swyddogaethau gwaith ymchwil a dysgu ac addysgu.
- Caffael iaith a dwyieithrwydd: Mae'r cyfleusterau dysgu a yrrir gan ddata sydd wedi'u hymgorffori yn cynnig cyfle unigryw ar gyfer ymchwilio i ymddygiad dysgwyr annibynnol a dysgu cyfunol. Mae CorCenCC yn ymgorffori cyfres o offerynnau sy'n seiliedig ar amlder ar gyfer gwneud gwaith ymchwil i batrymau o gaffael iaith ac i broffilio testun a dysgwyr.
- Sosioieithyddiaeth, tafodieitheg a dadansoddiadau morffogystrawennol: Mae CorCenCC yn ymestyn cwmpas y data sgyrsiol sydd ar gael o'r corpws Siarad (www.bangortalk.org.uk) er mwyn cynnwys siaradwyr o amrediad ehangach o ranbarthau daearyddol. Bydd hyn yn rhoi mwy o wybodaeth inni o safbwynt faint o gyswllt ieithyddol a geir rhwng y Gymraeg a'r Saesneg. Bydd yn hwyluso'r gwaith o archwilio amrywiaeth sosioieithyddol ym mhatrymau defnydd yr iaith o safbwynt geiriadurol, gramadegol, semantig ac ymarferol, ac o safbwynt priodweddau defnyddio'r iaith sy'n seiliedig ar acenion, o fewn ac ar draws rhanbarthau, ac yn ôl proffiliau defnyddwyr y Gymraeg, a thrwy hynny ceir dealltwriaeth ddyfnach o ddynameg gymdeithasol y Gymraeg fel iaith fyw ac sy'n adfywio.
- Cynllunio ieithyddol: Mae CorCenCC yn darparu data llinell sylfaen y gellir ei ddefnyddio i ymchwilio i ddefnydd yr iaith yng nghyd-destun polisïau sy'n ymwneud â defnyddio'r Gymraeg ym meysydd gweinyddiaeth gyhoeddus, masnach ac addysg.
- Geiriadura: Bydd CorCenCC yn uniongyrchol berthnasol i waith y tîm sydd wrthi'n diweddaru Geiriadur Prifysgol Cymru (sy'n bartner prosiect), drwy ddarparu tystiolaeth sylfaenol o ddefnydd geiriadurol o ffynonellau y gellir eu priodoli ac y gellir eu cynnwys yn niwygiadau'r geiriadur yn y dyfodol.
- Technoleg gyfrifiadurol a thechnoleg cyfieithu: Fel corpws â thagiau a bennir, bydd yn bosibl defnyddio CorCenCC ar gyfer datblygu systemau cyfieithu peirianyddol drwy gyfrannu at beiriant cyfieithu peirianyddol ystadegol, ac iddo hwyluso'r broses o haniaethu rheolau gramadegol o'r corpws. Bydd meysydd eraill o brosesu iaith naturiol (e.e. cywiro sillafu, rhagfynegi geiriau, meddalwedd gynorthwyol ar gyfer y Gymraeg) yn manteisio o'r modelau ieithyddol mwy manwl gywir, ac sydd ar gael yn ehangach, a gaiff eu cynhyrchu o ganlyniad i ddadansoddi data o CorCenCC.
- Meysydd ymchwil eraill: Mae CorCenCC yn cynnig cyfleoedd newydd i academyddion sy'n astudio meysydd arddulleg a llenyddiaeth, y cyfryngau, seicoieithyddiaeth, ymarferoleg, busnes, iechyd a meddygaeth, a seicoleg ymestyn eu gwaith ymchwil i gyd-destun y Gymraeg.

Mae CorCenCC yn adnodd sydd ar gael yn rhad ac am ddim o dan drwydded agored a fydd, o'i gyfuno â'i gynllun, a'r dull o'i adeiladu a yrrir gan ddefnyddwyr, yn cynyddu ei effaith gymdeithasol bosibl, gan lywio gwaith a gweithgareddau defnyddwyr presennol y Gymraeg, a rhai'r dyfodol, mewn sawl maes hanfodol bwysig. Mae meysydd ymarferwyr posib nad ydynt yn academaidd a meysydd proffesiynol posib yn cynnwys y canlynol:

- Dysgu ac addysgu ail iaith: Fel a drafodwyd yn Adran 6, mae adroddiadau ar addysgu Cymraeg i Oedolion (Mac Giolla Chríost ac eraill, 2012; Llywodraeth Cymru, 2013) wedi tynnu sylw at yr angen am gorpws o Gymraeg cyfoes fel ffordd o wella effeithiolrwydd mentrau dysgu'r Gymraeg. Mae CorCenCC yn gweithio i ddiwallu'r angen hwn. Drwy lywio gwaith ysgrifenedig y cwricwlwm, gwaith asesu'r iaith ac adnoddau dysgu'r iaith fel y mae corpysau tebyg yn ei wneud yn effeithiol yn y Saesneg (e.e. CEC a BNC), bydd CorCenCC yn hwyluso dysgu a yrrir gan ddata, gan wella effeithiolrwydd addysgu Cymraeg fel ail iaith (sy'n orfodol ymhob ysgol yng Nghymru hyd at ddiwedd Cyfnod Allweddol 4). Rhagwelir y bydd yr effeithiau yn y tymor canolig yn cynnwys gwelliant yn effeithiolrwydd dysgu'r iaith; gwell ymwybyddiaeth gan athrawon a dysgwyr, o safbwynt mân-wahaniaethau, o'r amrywiaeth gynhenid a naturiol a geir yn y Gymraeg fesul rhanbarth, genre, a math o siaradwr; cynnydd yn hyder siaradwyr Cymraeg ynghylch dilysrwydd eu patrymau defnydd eu hunain; a gwell ymwybyddiaeth o'r Gymraeg, a balchder ynddi, fel ffordd fyw a datblygol o fynegi Cymreictod.
- Llywodraeth Cymru a Senedd Cymru (polisi iaith): Mae CorCenCC yn hwyluso gwireddiad y pwyntiau gweithredu yn strategaeth Comisiynydd y Gymraeg sy'n ymwneud â chynnwys a chymwysiadau digidol, cyfieithu, terminoleg, cynllunio ieithyddol a gwaith ymchwil. Mae'r rhain yn adlewyrchu blaenoriaethau Llywodraeth Cymru (2014; 2017). O ganlyniad, mae CorCenCC yn debygol o gael effaith ar ddatblygu integreiddiad y Gymraeg ym mywyd beunyddiol ymhellach, fel iaith a ddefnyddir ym meysydd llywodraethu a masnach, ac wrth ryngweithio'n gymdeithasol.
- Y diwydiant cyfieithu yng Nghymru: Mae allbynnau CorCenCC yn cyfateb â gwaith datblygu tymor canolig meddalwedd Microsoft Translate. Mae gwaith ymchwil rhagarweiniol (Screen, 2014) yn dangos y gall cyfieithu peirianyddol sy'n seiliedig ar enghreifftiau yn unig wella cynhyrchiant cyfieithwyr dynol hyd at 55%. Drwy gyfrannu at system cyfieithu peirianyddol hybrid yn y pen draw, gallai CorCenCC wella effeithlonrwydd cyfieithu ymhellach.
- Y cyfryngau yng Nghymru: Mae CorCenCC yn cynnig ffordd i gwmnïau'r cyfryngau werthuso natur ieithyddol eu hallbwn drwy fesur pa mor anodd yw eu deunydd i'w ddeall (e.e. cyfrifo cyfran yr eirfa sydd ag amlder isel) a thrwy ddarganfod Cymraeg pwy a gynrychiolir ac a dangynrychiolir. Drwy'r modd hwn, gallai CorCenCC gael effaith faterol ar hygyrchedd rhaglenni Cymraeg ac allbynnau'r cyfryngau, a pha mor ddeniadol ydynt, i'w cynulleidfaoedd targed, gan arwain at gynnydd mewn ffigurau gwylio/ymgysylltu o ganlyniad, yn ogystal â chefnogi cydraddoldeb cymdeithasol.

- Cyhoeddwyr a geiriadurwyr Cymraeg: Mae CorCenCC yn cynnig modd o dargedu cynnwys at gynulleidfaoedd sydd â gallu darllen gwahanol a gwella'r offerynnau ieithyddol sydd ar gael i awduron ar gyfer creu llyfrau darllen sydd wedi'u graddio. Bydd yn galluogi gwaith comisiynu geiriaduron Cymraeg modern sy'n seiliedig ar ddefnydd iaith gwirioneddol, gan gau'r bwlch a gydnabyddir, ac sy'n aml yn peri problemau, rhwng yr hyn sydd ei angen ar ddefnyddwyr y Gymraeg a'r hyn y gallant ddod o hyd iddo mewn ffynonellau cyfeiriadurol. Bydd hyn, yn ei dro, yn meithrin hyder siaradwyr Cymraeg yn eu galluoedd ieithyddol, gan eu gwneud yn fwy bodlon i ddefnyddio'r Gymraeg mewn amrediad ehangach o gyd-destunau.
- Cwmnïau technoleg ieithyddol: Mae corpws hyfforddi o'r radd flaenaf yn ofyniad allweddol ar gyfer cwmnïau sy'n defnyddio data'r cyfryngau cymdeithasol ar y we ac ar-lein, a dyma'r hyn y mae CorCenCC yn ei ddarparu. Bydd set ddata CorCenCC felly'n ei gwneud yn bosibl datblygu amrediad o adnoddau technoleg Cymraeg nad ydynt eto'n bodoli ar gyfer yr iaith.
- Y cyhoedd: Drwy ei gynllun a yrrir gan ddefnyddwyr, mae cynrychiolwyr y bobl sy'n debygol o ddefnyddio CorCenCC yn y dyfodol wedi bod ynghlwm wrth y gwaith o adeiladu a chynllunio'r corpws yn uniongyrchol, ac mae hyn wedi sicrhau ei fod yn hwylus i ddefnyddwyr, a'i fod yn hygyrch ac yn briodol i'w hanghenion. Bwriad y dull hwn yw adeiladu ar ddiddordeb sydd eisoes yn bodoli yn y Gymraeg a'i threftadaeth, a meithrin 'perchenogaeth' gymunedol o'r corpws. Mae'r effaith hirdymor bosibl yn cynnwys newid pendant mewn canfyddiadau o'r Gymraeg, ac ymagweddau tuag ati, yng Nghymru a'r tu hwnt.

## 9. Prosiectau cysylltiedig a chyllid pellach

Er y derbyniodd CorCenCC gyllid hael gan y Cyngor Ymchwil Economaidd a Chymdeithasol (ESRC) a'r Cyngor Ymchwil i'r Celfyddydau a'r Dyniaethau (AHRC), cafodd amrediad o is-brosiectau a phrosiectau ymchwil cysylltiedig eraill eu hariannu gan ffynonellau eraill. Ceir manylion y rhain isod:

| Dyddiad | Cyllidwr | Swm | Disgrifiad [ynghyd â'r prif ymchwilydd] |
|---|---|---|---|
| Ion 2017 | Prifysgol Caerdydd | £56,000 | Derbyniwyd cyllid gan Goleg y Celfyddydau, y Dyniaethau a'r Gwyddorau Cymdeithasol (AHSS) er mwyn cynnal ysgoloriaeth doethuriaeth tair blynedd ar gyfer Vigneshwaran Muralidaran ac astudiaeth yn dwyn y teitl '*Using insights from construction grammar for usage-based parsing*' [Knight a Spasić]. |
| Chwef 2017 | British Council | £2,000 | Cyllid i gefnogi lansiad cyhoeddus prosiect CorCenCC yn Adeilad y Pierhead, Caerdydd [Knight]. |
| Chwef 2017 | Prifysgol Abertawe | £1,000 | Derbyniwyd cyllid gan Sefydliad Ymchwil y Celfyddydau a'r Dyniaethau (RIAH) Prifysgol Abertawe i gefnogi lansiad prosiect CorCenCC [Fitzpatrick]. |
| Chwef 2017 | Prifysgol Caerdydd | £1,500 | Derbyniwyd cymorth gan Gronfa Ymchwil ac Arloesedd Ysgolion ar gyfer lansiad prosiect CorCenCC [Knight]. |
| Hyd 2017 | Llywodraeth Cymru | £24,992 | Comisiwn cystadleuol gan Lywodraeth Cymru i ddarparu asesiad cyflym o'r dystiolaeth o ymagweddau a dulliau effeithiol ar gyfer addysgu ail iaith. Am fwy o wybodaeth, gweler: https://tinyurl.com/ybtdsvfy [Fitzpatrick]. |
| Ion 2018 | Cynllun Grant Cymraeg 2050 2017-2018 GC2050/17-18/20: | £19,964 | Cyllid ar gyfer adeiladu WordNet cyfrifiadurol ar gyfer y Gymraeg. Mae WordNet Cymru yn gronfa ddata eiriadurol lle caiff geiriau eu grwpio'n setiau o gyfystyron (synsetiau), sydd yno'n cael eu trefnu'n rhwydwaith o gysylltiadau semanteg-eiriadurol. I weld gwefan WordNet Cymru, ewch i: http://corcencc.org/wncy/ [Spasić]. |
| Ion 2018 | Cyd-bwyllgor Addysg Cymru (CBAC) | £1,968 | Grant ymchwil (gan gynnwys rhaglen fewnfurol). Grant ymchwil i gwblhau gwaith ar lunio geirfa graidd B1 ar gyfer Cymraeg i Oedolion (lefel Canolradd). Am fwy o wybodaeth, ewch i: http://cronfa.swan.ac.uk/Record/cronfa48953 [Morris]. |
| Maw 2018 | Prifysgol Abertawe | £1,200 | Lleoliad SPIN (interniaeth a delir gan Brifysgol Abertawe) ar gyfer gwaith casglu data, trawsgrifio a chyfweld ag athrawon/tiwtoriaid 2017-18. Ysgoloriaeth at ddibenion meithrin gallu [Morris]. |
| Ebr 2018 | Prifysgol Abertawe | £57,121 | Cyllid gan Goleg y Celfyddydau a'r Dyniaethau (COAH) i gynnal ysgoloriaeth doethuriaeth tair blynedd ar gyfer Bethan Tovey-Walsh ac astudiaeth yn dwyn y teitl '*Purism and populism: The contested roles of code-switching and borrowing in minority language evolution*'. Talwyd ffioedd a chynhaliaeth [Morris a Fitzpatrick]. |
| Gorff 2018 | Prifysgol Caerdydd | £2,100 | Cyllid mewnol CUROP (Cyfle Ymchwil Prifysgol Caerdydd) ar gyfer prosiect yn dwyn y teitl: '*Corpws Cenedlaethol Cymraeg Cyfoes: National Corpus of Contemporary Welsh – a focus on spoken data*'. Ysgoloriaeth at ddibenion meithrin gallu [Morris]. |
| Gorff 2018 | Prifysgol Caerdydd | £2,100 | Cyllid mewnol CUROP (Cyfle Ymchwil Prifysgol Caerdydd) ar gyfer prosiect yn dwyn y teitl: '*Corpws Cenedlaethol Cymraeg Cyfoes: National Corpus of Contemporary Welsh – semantic tagging and data annotation*'. Ysgoloriaeth at ddibenion meithrin gallu [Morris]. |
| Hyd 2018 | Ysgoloriaeth gydweithredol ESRC DTP, Prifysgol Abertawe | £81,253 | Y Gymraeg ac Ieithyddiaeth Gymhwysol Ysgoloriaeth Doethuriaeth Partneriaeth Hyfforddiant Doethurol (DTP) Cyngor Ymchwil Economaidd a Chymdeithasol (ESRC) Cymru, yn dwyn y teitl '*Strategic bilingualism: identifying optimal context for Welsh as a second language in the curriculum*' [Morris]. |
| Ion 2019 | Llywodraeth Cymru | £20,000 | Cyllid i gefnogi'r gwaith o ddatblygu boniwr Cymraeg [Spasić]. |
| Awst 2019 | Llywodraeth Cymru | £90,000 | Prosiect yn dwyn y teitl: '*Welsh language processing infrastructure: Welsh word embeddings*'. Mae ymgorffori geiriau'n math o gynrychioli geiriau lle caiff geiriau neu ymadroddion sydd ag ystyr tebyg eu mapio i fectorau o rifau real. Roedd y prosiect yn canolbwyntio ar ymgorffori geiriau ar gyfer y Gymraeg (ar greu geiriadur ac ymgorffori geiriau a thermau Cymraeg yn bennaf) ac mae'n cyfrannu at nod y Cynllun Gweithredu Technoleg Gymraeg i 'hybu adnoddau codio a thechnoleg Gymraeg ar gyfer athrawon a phlant ac eraill'. |
| Mai 2020 | Llywodraeth Cymru | £90,000 | Prosiect yn dwyn y teitl: 'Learning English-Welsh bilingual embeddings and applications in text categorisation'. Nod y prosiect hwn yw ymestyn canlyniadau'r prosiect ymgorffori geiriau blaenorol drwy greu cynrychioliadau trawsieithol o eiriau mewn man ymgorffori ar y cyd ar gyfer Cymraeg a Saesneg [Knight]. |
| | | **£451,198** | |

# 10. Crynodeb o allbynnau'r prosiect

## 10.1. Offerynnau meddalwedd

| Enw | Manylion | Dolen |
|---|---|---|
| Ap cyfrannu torfol CorCenCC: | Gyda'r nod o alluogi siaradwyr Cymraeg i recordio sgyrsiau rhyngddynt eu hunain a phobl eraill ar draws amrediad o gyd-destunau ac i'w lanlwytho, gan gynnwys caniatâd gan gyfranogwyr sy'n cydymffurfio'n foesegol, ar gyfer eu cynnwys yn y corpws terfynol. Mae'r dull torfol o gasglu data corpws yn gyfeiriad datblygu cymharol newydd sy'n ategu dulliau mwy traddodiadol o gasglu data ieithyddol, ac mae'n cyfateb yn ddelfrydol â'r ysbryd cymunedol cadarnhaol sy'n bodoli ymhlith siaradwyr a defnyddwyr y Gymraeg. | http://www.corcencc.cymru/ap/<br><br>http://app.corcencc.org<br><br>**Cyfeiriad:** Knight, D., Loizides, F., Neale, S., Anthony, L. a Spasić, I. (2020). Developing computational infrastructure for the CorCenCC corpus – the National Corpus of Contemporary Welsh. *Language Resources and Evaluation (LREV).* |
| CyTag – Tagiwr rhannau ymadrodd Cymraeg | Mae CyTag yn dagiwr Cymraeg arloesol (sy'n cynnwys set dagiau bwrpasol) a gafodd ei gynllunio a'i adeiladu ar gyfer y prosiect. Caiff ei ddefnyddio ar y cyd â'r tagiwr | http://cytag.corcencc.org<br><br>**Cyfeiriad:** Neale, S., Donnelly, K., Watkins, G. a Knight, D. (2018). Leveraging Lexical Resources and Constraint Grammar for Rule-Based Part-of-Speech Tagging in Welsh. |
| | semantig i bennu tagiau i holl eitemau geiriadurol y corpws. | Poster a gyflwynwyd yn ystod *Cynhadledd Gwerthuso Adnoddau Iaith (LREC) 2018,* Mai 2018, Miyazaki, Japan. |
| Fersiwn 1 tagiwr semantig y Gymraeg CySemTag | Mae tagiwr semantig y Gymraeg yn cymhwyso anodiadau corpws i ddata am y Gymraeg mewn modd awtomataidd. | http://ucrel.lancs.ac.uk/usas/<br><br>**Cyfeiriad:** Piao, S., Rayson, P., Knight, D. a Watkins, G. 2018). Towards a Welsh Semantic Annotation System. *Proceedings of the LREC (Language Resources Evaluation 2018 Conference,* Mai 2018, Miyazaki, Japan.<br><br>Piao, S., Rayson, P., Knight, D., Watkins, G. a Donnelly, K. (2017). Towards a Welsh Semantic Tagger: Creating Lexicons for A Resource Poor Language. *Proceedings of the Corpus Linguistics 2017 Conference,* Gorffennaf 2017, Prifysgol Birmingham, Birmingham, DU. |
| Seilwaith ac offerynnau ymholi CorCenCC | Mae offerynnau ymholi CorCenCC yn cynnwys y swyddogaethau canlynol:<br>▪ Ymholiad syml<br>▪ Ymholiad cymhleth<br>▪ Cynhyrchu rhestrau amlder<br>▪ Dadansoddi cydleoliad<br>▪ Dadansoddi n-gramau<br>▪ Cydgordio<br>▪ Dadansoddi geiriau allweddol | I weld yr offerynnau a chanllaw i ddefnyddwyr, ewch i: www.corcencc.cymru/archwilio<br><br>**Cyfeiriad:** Knight, D., Loizides, F., Neale, S., Anthony, L. a Spasić, I. (2020). Developing computational infrastructure for the CorCenCC corpus – the National Corpus of Contemporary Welsh. *Language Resources and Evaluation (LREV).* |
| Y Tiwtiadur | Pecyn cymorth pedagogaidd CorCenCC sydd wedi'i integreiddio yn y prif offerynnau ymholi. Mae hyn yn cynnwys yr offerynnau dysgu ac addysgu canlynol:<br>▪ Llenwi bylchau<br>▪ Proffilydd geirfa<br>▪ Nodi geiriau<br>▪ Peiriant creu bylchau mewn brawddegau | I weld yr offerynnau a chanllaw i ddefnyddwyr, ewch i: www.corcencc.cymru/archwilio<br><br>**Cyfeiriad:** Davies, J., Thomas, E-M., Fitzpatrick, T., Needs, J., Anthony, L., Cobb, T. a Knight, D. (2020). *Y Tiwtiadur.* [Adnodd Digidol]. Ar gael yn: www.corcencc.cymru/Y-tiwtiadur |

10.2. Cyhoeddiadau (gyda'r dyddiadau yn y drefn wrthol, gydag enwau aelodau tîm y prosiect wedi'u nodi'n fras):

12. **Knight, D., Morris, S., Arman, L., Needs, J.** a **Rees, M.** (2021a, yn cael ei baratoi). *Blueprints for minoritised language corpus design: a focus on CorCenCC.* Llundain: Palgrave.
13. **Knight, D., Morris, S.** a **Fitzpatrick, T.** (2021b, yn cael ei baratoi). *Corpus Design and Construction in Minoritised Language Contexts:* **A focus on CorCenCC (Corpws**

**Cenedlaethol Cymraeg Cyfoes – National Corpus of Contemporary Welsh).** Llundain: Palgrave.

14. **Knight, D.,** Loizides, F., **Neale, S., Anthony, L.** a **Spasić, I.** (2020). Developing computational infrastructure for the CorCenCC corpus – the National Corpus of Contemporary Welsh. **Language Resources and Evaluation (LREV).**
15. Corcoran, P., Palmer, G., **Arman, L., Knight, D.** a **Spasić, I.** (2020, derbyniwyd). Word Embeddings in Welsh. *Journal of Information Science*.
16. **Muralidaran, V., Knight, D.** a **Spasić, I**. (2020, derbyniwyd). A systematic review of unsupervised approaches to usage-based grammar induction. *Natural Language Engineering.*
17. **Spasić, I.**, Owen, D., **Knight, D.** ac Arteniou, A. (2019). Data-driven terminology alignment in parallel corpora. *Proceedings of the* **Celtic Language Technology Workshop 2019,** Dulyn, Iwerddon.
18. **Piao, S., Rayson, P., Knight, D.** a **Watkins, G**. 2018). Towards a Welsh Semantic Annotation System. *Proceedings of the LREC (Language Resources Evaluation) 2018 Conference,* Mai 2018, Miyazaki, Japan.
19. **Neale, S., Donnelly, K., Watkins, G.** a **Knight, D**. 2018). Leveraging Lexical Resources and Constraint Grammar for Rule-Based Part-of-Speech Tagging in Welsh. Poster a gyflwynwyd yn ystod *Cynhadledd Gwerthuso Adnoddau Iaith (LREC) 2018,* Mai 2018, Miyazaki, Japan.
20. **Rayson, P.** 2018). Increasing Interoperability for Embedding Corpus Annotation Pipelines in Wmatrix and other corpus retrieval tools. Trafodion y gweithdy Heriau wrth Reoli Corpysau Mawr yn ystod *Cynhadledd Gwerthuso Adnoddau Iaith (LREC) 2018,* Mai 2018, Miyazaki, Japan.
21. **Rayson, P**. a **Piao, S**. (2017). Creating and Validating Multilingual Semantic Representations for Six Languages: Expert versus Non-Expert Crowds. Trafodion y Gweithdy Cyntaf ar 'Sense, Concept and Entity Representations and their Applications' a gynhaliwyd yn ystod cynhadledd yr *European Chapter of the Association for Computational Linguistics 2017* (EACL), Ebrill, Valencia.
22. **Piao, S., Rayson, P.,** Archer, D., Bianchi, F., Dayrell, C., El-Haj, M., Jiménez, R-M., **Knight, D.,** Křen, M., Löfberg, L., Nawab, R. M. A., Shafi, J., Teh, P-L., a Mudraya, O. (2016). Lexical Coverage Evaluation of Large-scale Multilingual Semantic Lexicons for Twelve Languages. *Proceedings of the LREC (Language Resources Evaluation) 2016 Conference,* Mai 2016, Miyazaki, Slofenia.

### 10.3. Prif gyflwyniadau a gwahoddiadau i siarad

Mae gwaith ymchwil prosiect CorCenCC wedi'i gyflwyno yn ystod 17 o brif gyflwyniadau a gwahoddiadau i siarad, ac mae wedi'i ledaenu drwy 37 o bapurau cynhadledd eraill mewn 11 o wledydd o gwmpas y byd. Gellir cael manylion am y digwyddiadau siarad hyn ar brif wefan CorCenCC (gweler: www.corcencc.cymru/allbynnau).

# Cyfeiriadau

Aston, G. (2001) *Learning with Corpora*, Athelstan, Open Library.

Aston, G. a Burnard, L. (1997) *The BNC Handbook: Exploring the British National Corpus with SARA,* Caeredin: Gwasg Prifysgol Caeredin.

Ball, M. a Müller, N. (1992) *Mutation in Welsh,* Clevedon: Multilingual Matters.

Brabham, D. C. (2008) 'Crowdsourcing as a model for problem solving: An introduction and cases', *Convergence* 14**:** 75-90.

Carter, R. a McCarthy, M. (2004) 'Talking, creating: Interactional language, creativity, and context', *Applied Linguistics* 25**:** 62-88.

Cobb, T. (2000). *The compleat lexical tutor* [Ar-lein], ar gael: http://www.lextutor.ca/ [Cyrchwyd 07/07/20].

Collins. (2020). *Collins Corpus online* [Ar-lein], ar gael: https://collins.co.uk/pages/elt-cobuild-reference-the-collins-corpus [Cyrchwyd 07/07/20].

Cooper, Jones, D. a Prys (2019) 'Crowdsourcing the Paldaruo Speech Corpus of Welsh for Speech Technology', *Information* 10**:** 247.

Cambridge University Press. (2020) *Cambridge English Corpus online* [Ar-lein], ar gael: https://www.cambridge.org/us/cambridgeenglish/better-learning-insights/corpus [Cyrchwyd 07/07/20].

Deuchar, D., Webb-Davies, P. a Donnelly, K. (2018) *Building and Using the Siarad Corpus,* Amsterdam: John Benjamins.

Deuchar, M., Davies, P., Herring, J. R., Parafita Couto, M. a Carter, D. (2014) 'Building bilingual corpora: Welsh-English, Spanish-English and Spanish-Welsh', yn Thomas, E. M. a Mennen, I. (gol.) *Advances in the Study of Bilingualism.* Bristol: Multilingual Matters.

Donnelly, K. (2013a) *Eurfa v3.0 - Free (GPL) Dictionary (incorporating Konjugator and Rhymer)* [Ar-lein], ar gael: http://eurfa.org.uk [Cyrchwyd 07/07/20].

Donnelly, K. (2013b) *Kynulliad3: a corpus of 350,000 aligned Welsh and English sentences from the Third Assembly (2007-2011) of the National Assembly for Wales* [Ar-lein], ar gael: http://cymraeg.org.uk/kynulliad3/ [Cyrchwyd 07/07/20].

Donnelly, K. a Deuchar, M. (2011) 'The Bangor Autoglosser: A multilingual tagger for conversational text', yn Proceedings of *the Fourth International Conference on Internet Technologies and Applications (ITA11),* Wrecsam, Cymru. tt. 17-25.

Expert Advisory Group on Language Engineering Standards. (1996) *EAGLES guidelines* [Ar-lein], ar gael: http://www.ilc.cnr.it/EAGLES/browse.html [Cyrchwyd 07/07/20].

Estellés-Arolas, E. a González-Ladrón-De-Guevara, F. (2012) 'Towards an integrated crowdsourcing definition', *Journal of Information Science* 38**:** 189-200.

Evas, J. a Williams, C. H. (1998) 'Community language regeneration: realising potential', yn Nhrafodion yr *International Conference on Community Language Planning,* Caerdydd, Bwrdd yr Iaith Gymraeg. tt. 1-13.

Hardie, A. (2012) 'CQPweb – combining power, flexibility and usability in a corpus analysis tool', *International Journal of Corpus Linguistics* 17**:** 380-409.

Hawtin, A. (2018) *The Written British National Corpus 2014: Design, compilation and analysis,* Traethawd PhD heb ei gyhoeddi: Prifysgol Caerhirfryn.

Johns, T. (1991) 'Should you be persuaded: Two samples of data-driven learning materials', *English Language Research Journal* 4: 1-16.

Karlsson, F. (1990) 'Constraint grammar as a framework for parsing running text', yn Nhrafodion y *13th International Conference on Computational Linguistics (COLING),* Helsinki, Y Ffindir. tt. 168-173.

Karlsson, F., Voutilainen, A., Heikkilä, J. a Anttila, A. (1995) *Constraint grammar: A language-independent framework for parsing unrestricted text,* Berlin/Efrog Newydd: Mouton de Gruyter.

Knight, D., Adolphs, S. a Carter, R. (2013) 'Formality in digital discourse: a study of hedging in CANELC', in Romero-Trillo, J. (gol.) *Yearbook of corpus linguistics and pragmatics,* Yr Iseldiroedd: Springer. tt. 131-152.

Leńko-Szymańska, A. a Boulton, A. (2015) *Multiple Affordances of Language Corpora in Data-driven Learning,* Amsterdam: John Benjamins.

Little, D. (2007) 'Language learner autonomy: Some fundamental considerations revisited', *Innovations in Language Learning and Teaching* 1: 14-29.

Llywodraeth Cymru (2013) *Codi golygon: adolygiad o Gymraeg i Oedolion. Adroddiad ac argymhellion [Raising our sights: review of Welsh for Adults. Report and recommendations]*, Bedwas: Llywodraeth Cymru.

Llywodraeth Cymru (2014) *Iaith fyw: iaith byw - Bwrw mlaen*, Caerdydd: Llywodraeth Cymru.

Llywodraeth Cymru (2017) *Cymraeg 2050: Miliwn o siaradwyr Cynllun Gweithredu 2019-20*, Caerdydd: Llywodraeth Cymru.

McEnery, T., Love, R., & Brezina, V. (2017) 'Compiling and analysing the Spoken British National Corpus 2014', *International Journal of Corpus Linguistics 22*(3): 311-318.

Mac Giolla Chríost, D., Carlin, P., Davies, S., Fitzpatrick, T., Jones, A. P., Heath-Davies, R., Marshall, J., Morris, S., Price, A., Vanderplank, R., Walter, C. a Wray, A. (2012). *Adnoddau, dulliau ac ymagweddau dysgu ac addysgu ym maes Cymraeg i Oedolion: astudiaeth ymchwil gynhwysfawr ac adolygiad beirniadol o'r ffordd ymlaen [Welsh for Adults teaching and learning approaches, methodologies and resources: a comprehensive research study and critical review of the way forward]*, Bedwas: Llywodraeth Cymru.

Nation, I.S.P. (2001) *Learning Vocabulary in Another Language*, Caergrawnt: Gwasg Prifysgol Caergrawnt.

Neale, S., Donnelly, K., Watkins, G. a Knight, D. (2018) 'Leveraging lexical resources and constraint grammar for rule-based part-of-speech tagging in Welsh' yn Nhrafodion yr *Eleventh International Conference on Language Resources and Evaluation (LREC),* Miyazaki, Siapan. tt. 3946-3954.

NFER (2008) *Ymchwil i'r Cwrs Dwys ar gyfer Cymraeg i Oedolion*, Abertawe: Sefydliad Cenedlaethol er Ymchwil i Addysg.

ONS (2011) *DC2612WA – Ability to speak Welsh by occupation* [Ar-lein], Office for National Statistics, Durham: Nomis. Ar gael: www.nomisweb.co.uk/census/2011/dc2612wa [Cyrchwyd 07/07/20].

Piao, S., Rayson, P., Knight, D. a Watkins, G. (2018) 'Towards a Welsh semantic annotation system' yn Nhrafodion yr *Eleventh International Conference on Language Resources and Evaluation (LREC),* Miyazaki, Siapan. tt. 980-985.

Rayson, P., Archer, D., Piao, S. a McEnery, T. (2004) 'The UCREL semantic analysis system', yn Nhrafodion y *Workshop on Beyond Named Entity Recognition Semantic Labelling for NLP tasks at the 4th International Conference on Language Resources and Evaluation (LREC),* Lisbon, Portiwgal. tt. 1-6.

Scannell, K. (2007) 'The Crúbadán Project: Corpus building for under-resourced languages' yn *Building and Exploring Web Corpora: Proceedings of the 3rd Web as Corpus Workshop*. tt. 1-10.

Scannell, K. (2012) *Kevin Scannell's website* [Ar-lein], ar gael: http://borel.slu.edu/ [Cyrchwyd 07/07/20].

Sinclair, J. (2005) 'Corpus and text - basic principles', yn Wynne, M. (gol.) *Developing Linguistic Corpora: a Guide to Good Practice,* Rhydychen: Oxbow Books. tt. 1-16.

Thomas, E. M. a Mayr, R. (2010) 'Children’s acquisition of Welsh in a bilingual setting: a psycholinguistic perspective, yn Morris, D. (gol.) *Welsh in the 21st Century,* Caerdydd: Gwasg Prifysgol Caerdydd.

Thompson, P. (2005) 'Spoken Language Corpora' yn Wynne, M. (gol.) *Developing Linguistic Corpora: a Guide to Good Practice,* Rhydychen: Oxbow Books. tt. 59-70.

Wilson, A. (2002) *The Language Engineering Resources for the Indigenous Minority Languages of the British Isles and Ireland Project* [Ar-lein], ar gael: https://www.lancaster.ac.uk/fass/projects/biml/default.htm [Cyrchwyd 07/07/20].